\documentclass{article}

\PassOptionsToPackage{numbers, compress}{natbib}

\usepackage[final]{neurips_2024}

\usepackage[utf8]{inputenc} 
\usepackage[T1]{fontenc}    
\usepackage{url}            
\usepackage{booktabs}       
\usepackage{amsfonts}       
\usepackage{nicefrac}       
\usepackage{microtype}      
\usepackage{xcolor}         

\usepackage{booktabs}       
\usepackage{amsfonts}       
\usepackage{nicefrac}       
\usepackage{microtype}      
\usepackage{mathtools}
\usepackage{multirow}
\usepackage{bm}
\usepackage{epsfig}
\usepackage{graphicx}
\usepackage{caption}
\usepackage{subcaption}
\usepackage{amsthm}
\usepackage{amsmath}
\usepackage{amssymb}
\usepackage{enumitem}
\usepackage{makecell}
\usepackage{wrapfig}
\usepackage{indentfirst}
\usepackage{verbatim}
\usepackage{color}
\usepackage{setspace}
\usepackage{array}
\usepackage{booktabs}
\usepackage{stackengine}
\usepackage{algorithm}
\usepackage{graphicx}
\usepackage{xcolor}

\usepackage{soul}
\usepackage{twemojis}
\usepackage{rotating}
\usepackage{caption}
\renewcommand{\captionlabelfont}{\normalsize} 

\usepackage[pagebackref=true,breaklinks=true,colorlinks=true,bookmarks=false]{hyperref}
\definecolor{deepred}{HTML}{940000}
\hypersetup{linkcolor=deepred}
\hypersetup{urlcolor  = [rgb]{0.4,0.15,0.95}}
\hypersetup{citecolor=[rgb]{0.4,0.15,0.95}}

\usepackage{colortbl}
\definecolor{Gray}{gray}{0.94}
\definecolor{Gray}{gray}{0.94}

\newlength\savewidth

\usepackage{pifont}
\usepackage{tocloft}
\usepackage[toc,page,header]{appendix}
\usepackage{adjustbox}
\usepackage{minitoc}

\renewcommand \thepart{}
\renewcommand \partname{}

\usepackage{xcolor}
\usepackage{multicol,multirow}
\usepackage{overpic}
\usepackage{float}
\usepackage{wrapfig}
\usepackage{amsmath}
\usepackage{graphicx}
\usepackage{amssymb}
\usepackage{booktabs}
\usepackage{afterpage}
\usepackage{algorithm, algpseudocode}
\usepackage{tabularx}
	
\usepackage{color, colortbl}
\usepackage{soul}

\newcommand{\methodName}{3Diffusion}

\newcommand{\mat}[1]{\mathbf{#1}}

\newcommand{\gsplat}[0]{\mathcal{G}}

\colorlet{shadecolor20}{gray!20}
\colorlet{shadecolor10}{gray!10}

\usepackage{amsmath,amsfonts,bm}

\def\eqref#1{equation~\ref{#1}}

\def\1{\bm{1}}

\definecolor{forward}{RGB}{165,93,115}

\def\renderer{\textcolor{forward}{\texttt{renderer}}}

\def\rvx{{\mathbf{x}}}

\DeclareMathAlphabet{\mathsfit}{\encodingdefault}{\sfdefault}{m}{sl}
\SetMathAlphabet{\mathsfit}{bold}{\encodingdefault}{\sfdefault}{bx}{n}

\usepackage[capitalize]{cleveref}
\crefname{section}{Sec.}{Secs.}
\Crefname{section}{Section}{Sections}
\Crefname{table}{Table}{Tables}
\crefname{table}{Tab.}{Tabs.}
\Crefname{paragraph}{Paragraph}{Paragraphs}
\crefname{paragraph}{Para.}{Paras.}

\title{Human-3Diffusion: Realistic Avatar Creation \\ via Explicit 3D Consistent Diffusion Models}

\author{
Yuxuan Xue$^{1, 2}$\quad Xianghui Xie$^{1, 2,3}$\quad Riccardo Marin$^{1,2}$\quad Gerard Pons-Moll$^{1, 2, 3}$\vspace{8pt}\\
$^{1}$University of T\"ubingen\qquad $^{2}$ T\"ubingen AI Center\qquad \\ 
$^{3}$Max Planck Institute for Informatics, Saarland Informatics Campus
\\
\url{https://yuxuan-xue.com/human-3diffusion/}
\vspace{8pt}
}

\begin{document}

\doparttoc
\faketableofcontents

\maketitle

\begin{figure*}[h!]
\centering
\vspace{-0.5in}
\begin{tabular}{@{}c@{\,}}
\includegraphics[width=1.0\textwidth]{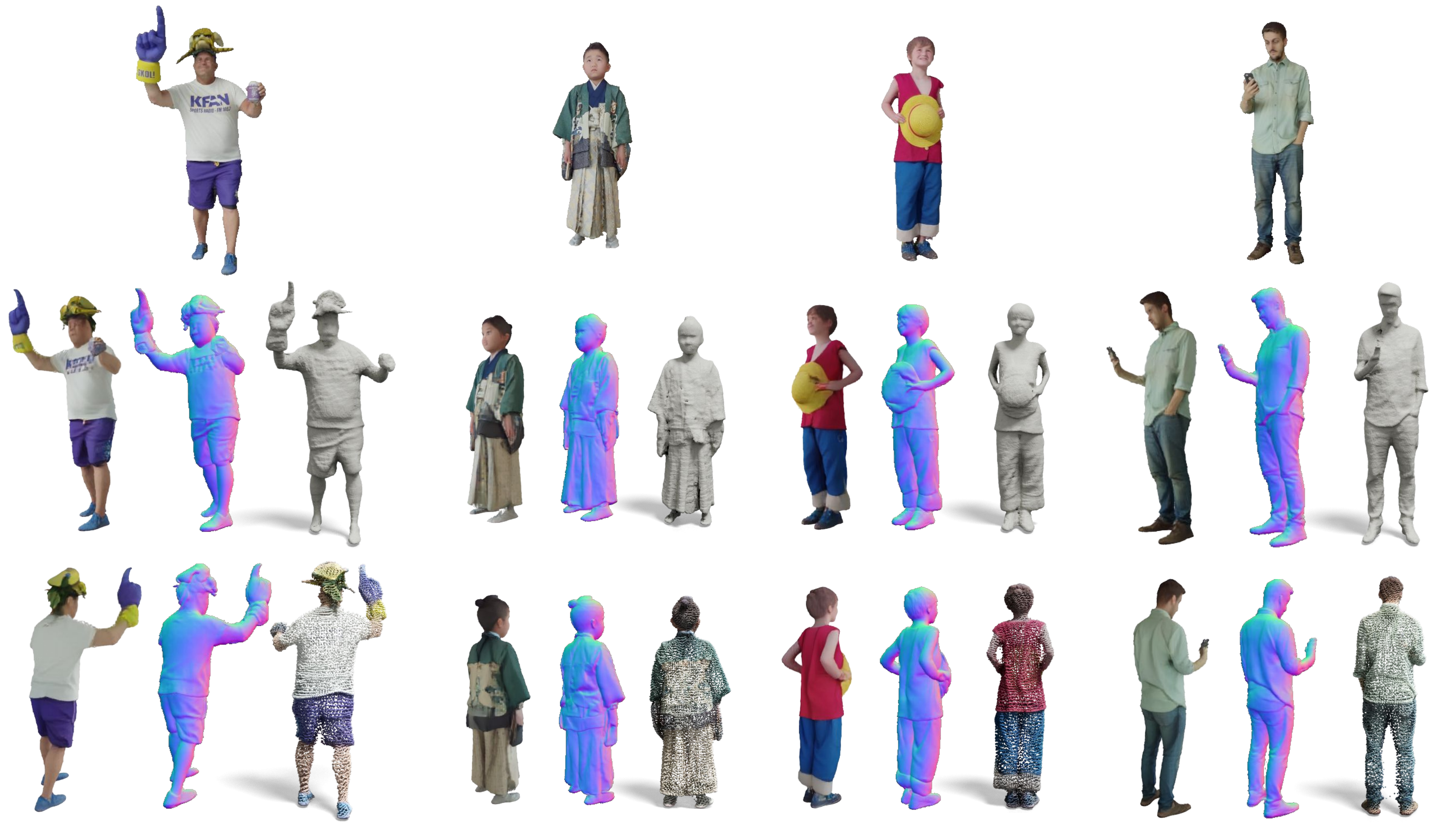}\\
\end{tabular}
\caption{Given a single image of a person (top), our method \textbf{Human-\methodName{}} creates 3D Gaussian Splats of realistic avatars with cloth and interacting objects with high-fidelity geometry and texture.}
\label{fig:teaser}
\end{figure*}

\begin{abstract}
Creating realistic avatars from a single RGB image is an attractive yet challenging problem. 
To deal with challenging loose clothing or occlusion by interaction objects, we leverage powerful shape prior from 2D diffusion models pretrained on large datasets.
Although 2D diffusion models demonstrate strong generalization capability, they cannot provide multi-view shape priors with guaranteed 3D consistency.
We propose \textbf{Human-\methodName{}}: Realistic Avatar Creation via Explicit \textbf{3D} Consistent \textbf{Diffusion}. 
Our key insight is that 2D multi-view diffusion and 3D reconstruction models provide complementary information for each other. By coupling them in a tight manner, we can fully leverage the potential of both models.  
We introduce a novel image-conditioned generative 3D Gaussian Splats reconstruction model that leverages the prior from 2D multi-view diffusion models, and provides an explicit 3D representation, which further guides the 2D reverse sampling process to have better 3D consistency.
Experiments show that our proposed framework outperforms state-of-the-art methods and enables the creation of realistic avatars from a single RGB image, achieving high-fidelity in both geometry and appearance. 
Extensive ablations also validate the efficacy of our design, (1) multi-view 2D priors conditioning in generative 3D reconstruction and (2) consistency refinement of sampling trajectory via the explicit 3D representation. 
Our code and models will be released \href{https://yuxuan-xue.com/human-3diffusion}{here}.

\end{abstract}

\section{Introduction}
\label{sec:introduction}
Realistic human avatar creation is crucial for various applications such as AR/VR, as well as the movie and gaming industry. Methods for creating a 3D avatar from a single RGB image are especially important to scale up avatar creation and make it more consumer-friendly compared to traditional studio-based capture methods. This task is, however, very challenging due to the vast diversity of human bodies and poses, further complicated by the wide variety of clothing and accessories. These challenges are exacerbated by the lack of large-scale 3D human data and ambiguities inherent in a monocular 2D view setting.

Recent image-to-3D approaches can be categorized into reconstruction-based and multi-view diffusion-based methods. 
Reconstruction-based approaches directly predict a 3D representation that can be rendered from any viewpoint. Due to the explicit 3D representation, these methods produce an arbitrary number of consistent viewpoint renderings. 
They obtain the 3D reconstruction either based on common template ~\cite{ho2023sith, xiu2023icon, xiu2023econ, zhang2023sifu} which utilize the SMPL~\cite{loper2015smpl} body model as the shape prior, or a flexible implicit function to represent loose clothing~\cite{alldieck_photorealistic_2022, saito2019pifu, saito2020pifuhd}. 
These methods, either template-based~\cite{ho2023sith, xiu2023icon, xiu2023econ, zhang2023sifu} or template-free ~\cite{alldieck_photorealistic_2022, hong2023lrm, saito2019pifu, saito2020pifuhd, tochilkin2024triposr, zou2023triplanegaussian}, are typically deterministic which produce blurry textures and geometry in the occluded regions. 
More importantly, they are trained on small-scale datasets due to the limited amount of high-quality 3D data, which further restricts their ability to generalize to diverse shapes and textures.

Multi-view diffusion methods~\cite{Liu2023zero123, shi2023zero123++, Wang2023ImageDream} distill the inherent 3D structure present in 2D diffusion models~\cite{Rombach2022StableDiffusion}.
Typically, they fine-tune a large-scale 2D foundation model~\cite{Ho2020DDPM, song_score-based_2021} on a large 3D dataset of objects~\cite{Deitke2023Objaverse, wu2023omniobject3d, yu2023mvimagenet}, to produce a \emph{fixed} number of viewpoints. However, since these models diffuse images purely in 2D without explicit 3D constraints or representation, the resulting multi-views often lack 3D consistency~\cite{qian2023magic123, liu2023one2345}, which restricts downstream applications~\cite{Tang2024LGM}.

To address these challenges, we propose \textbf{Human-\methodName{}}: realistic avatar creation via \textbf{3D} consistent \textbf{Diffusion} models. 
We design our method based on two key insights: 
1) 2D multi-view diffusion models provide large-scale shape priors that can help 3D reconstruction; 
2) A reconstructed 3D representation ensures 3D consistency across multi-views in 2D diffusion. 
Specifically, we propose a novel diffusion method, which bridges 3D Gaussian Splatting (3D-GS)~\cite{Kerbl20233dgs} generation with a 2D multi-view diffusion model. 
At every iteration, multi-view images are denoised and reconstructed to 3D-GS to be re-rendered to continue the diffusion process. 
This 3D lifting during iterative sampling ensures the 3D consistency of the 2D diffusion model while leveraging a large-scale foundation model trained on billions of images. 
Our framework elegantly combines reconstruction methods with multi-view diffusion models. 
In summary, our contributions are: 
\begin{itemize}
    \item We propose a novel image-conditioned 3D-GS generation model for 3D reconstruction that bridges large-scale priors from 2D multi-view diffusion models and the efficient and explicit 3D-GS representation.
    \item A sophisticated diffusion process that incorporates reconstructed 3D-GS to improve the 3D consistency of 2D diffusion models by refining the reverse sampling trajectory.   
    \item Our proposed formulation enables us to jointly train 2D diffusion and our 3D model on $\sim6000$ high-quality human scans and our method shows superior performance and generalization capability than prior works. 
    Our code and pretrained models will be publicly released on our \href{https://yuxuan-xue.com/human-3diffusion}{project page}.

\end{itemize}


\section{Related Work}
\label{sec:related_work}
\paragraph{Image to 3D.} Creating realistic human avatar from consumer grade sensors~\cite{kabadayi24ganavatar, kim2024paintit,tiwari2021neuralgif, xue2022e-nr, xue2023nsf, xue2024e-nr-ijcv} is essential for downstream tasks such as human behaviour understanding~\cite{bhatnagar22behave, petriv2023objectpopup, xie2022chore, xie2023template_free, xie2023vistrack} and gaming application~\cite{li2023diffavatar, liu2023gshell,guzov2021hps, zhang2022couch, zhang2024force}. Researchers have explored avatar creation from monocular RGB~\cite{jiang2022instantavatar, weng_humannerf_2022_cvpr}, Depth~\cite{dong2022pina, xue2023nsf} video or single image~\cite{saito2019pifu, saito2020pifuhd, sengupta_diffhuman_2024, xiu2023icon, xiu2023econ}. Avatar from single image is particularly interesting and existing methods can be roughly categorized as template-based~\cite{ho2023sith, xiu2023icon, xiu2023econ, zhang2023sifu} and template-free~\cite{saito2019pifu, saito2020pifuhd, sengupta_diffhuman_2024, yang_d-if_2023}. Despite the impressive performance, template-based approaches rely on the naked body model~\cite{loper2015smpl, pavlakos2019smplx} and fail to reason extremely loose clothing, while template-free methods produce blurry back side textures. Instead of SMPL shape prior, our method is template-free and leverages strong 2D image priors to create high-quality avatars.  
\\
Orthogonal to humans, object reconstruction methods typically adopt template-free paradigms and early works~\cite{Alwala_CVPR22_pretrain_ss, thai3dv2020SDFNet, shapehd, Xian2022gin, genre} focus mainly on geometry. With the advance of 2D diffusion models~\cite{Rombach2022StableDiffusion} and efficient 3D representation~\cite{chan2022eg3d}, recent works can reconstruct 3D objects with detailed textures~\cite{hong2023lrm,liu2023one2345, long2023wonder3d, shi2023zero123++,  Tang2024LGM, tochilkin2024triposr, xu2024instantmesh, Xu2023DMV3D, zou2023triplanegaussian}. One popular paradigm is first using strong 2D models~\cite{Liu2023zero123, shi2023mvdream, Wang2023ImageDream} to produce multi-view images and then train another model to reconstruct 3D from multi-view images~\cite{liu2023one2345, liu2023one2345++, long2023wonder3d, Tang2024LGM, xu2024grm}. In practice, their performance is limited by the accuracy of the multi-view images generated by 2D diffusion modes.  
Our method tightly couples 2D and 3D models and yields better performance by guiding 2D sampling with 3D reconstruction. 

\paragraph{Shape Prior from 2D Diffusion Model.} Being trained on billions of images~\cite{Schumann2022Laion5B}, 2D image diffusion models~\cite{Rombach2022StableDiffusion} have been shown to have 3D awareness and some works tried to use score distillation sampling~\cite{poole2022dreamfusion, wang2023prolificdreamer} to distil 3D knowledge of 2D models~\cite{liao2023tada, melaskyriazi2023realfusion, zhou2023sparsefusion}. Other works propose to further enhance the 3D reasoning ability by fine-tuning the model on large-scale datasets~\cite{Deitke2023Objaverse, wu2023omniobject3d, yu2023mvimagenet} to generate multi-view images~\cite{kant2024spad, kong2024eschernet, Liu2023zero123, liu2023syncdreamer, shi2023zero123++, shi2023mvdream, tang2023mvdiffusion++,  voletiSV3DNovelMultiview2024, Wang2023ImageDream}. Dense self-attention~\cite{vaswani2017attention, Wang2023ImageDream}, depth-aware attention~\cite{huMVDFusionSingleview3D2024} or epipolar attention~\cite{huangEpiDiffEnhancingMultiView2024, suhail2022epipolarattention} are introduced to enhance the 3D consistency of multi-views. However, these methods do not have explicit 3D while our method incorporates explicit 3D consistency into the reverse sampling process and obtains better results.


\section{Preliminaries}

\paragraph{Denoising Diffusion Probabilistic Models.} DDPM~\cite{Ho2020DDPM} is a generative model which learns a data distribution by iteratively adding (forward process) and removing (reverse process) the noise.
Formally, the forward process iteratively adds noise to a sample $\rvx_0$ drawn from a distribution $p_\text{data}(\mat{x})$:
\begin{equation}
    \rvx_t \sim\mathcal{N}(\rvx_t; \sqrt{\alpha_t} \rvx_{t-1},  (1-\alpha_t)\mathbf{I}) := \sqrt{\bar{\alpha}_t}\rvx_0 + \sqrt{1-\bar{\alpha}_t}\mat{\epsilon}, \text{ where }\mat{\epsilon}\sim\mathcal{N}(0, \mat{I}),
    \label{eq:ddpm_forward}
\end{equation}
where $\alpha_t, \bar{\alpha}_t$ schedules the amount of noise added at each step $t$~\cite{Ho2020DDPM}. To 
sample data from the learned distribution,
the reverse process starts from $\rvx_T\sim\mathcal{N}(0, \mat{I})$ and iteratively denoises it until $t=0$: 
\begin{equation}
    \rvx_{t-1}\sim\mathcal{N}(\rvx_{t-1};\mat{\mu}_\theta(\rvx_t, t), \tilde{\beta}_{t-1}\mat{I}), \text{ where }\tilde{\beta}_{t-1}=\frac{1-\bar{\alpha}_{t-1}}{1-\bar{\alpha}_{t}} (1-\alpha_t)
    \label{eq:ddpm_reverse}
\end{equation}
A network parametrized by $\theta$ is trained to estimate the posterior mean $\mat{\mu}_\theta$ at each step $t$. One can also model conditional distribution with DDPM by adding the condition to the network input~\cite{dhariwal2021diffusionClassGuidance, ho2021classifierfreeCFG}. 

\paragraph{2D Multi-View Diffusion Models.}
\label{subsec:MVD} 
Many recent works~\cite{Liu2023zero123, liu2023syncdreamer, long2023wonder3d, shi2023zero123++, tang2023mvdiffusion++, Wang2023ImageDream} propose to leverage strong 2D image diffusion prior~\cite{Rombach2022StableDiffusion} pre-trained on billions images~\cite{Schumann2022Laion5B} to generate multi-view images from a single image. 
Among them, ImageDream~\cite{Wang2023ImageDream} demonstrated a superior generalization capability to unseen objects~\cite{Tang2024LGM}.
Given a single condition image $\mathbf{x}^\text{c}$ and an optional text description $y$, ImageDream generate 4 orthogonal target views $\mathbf{x}^\text{tgt}$ with a model $\mat{\epsilon}_\theta$, which is trained to estimate the noise added at each step $t$. 
With the estimated noise $\mat{\epsilon}_\theta$, one can compute the "clear" target views $\Tilde{\mathbf{x}}^\text{tgt}_0$ with close-form solution in~\cref{eq:ddpm_forward}:  
\begin{align}
    \tilde{\mathbf{x}}_{0}^{\text{tgt}} &= \frac{1}{\sqrt{\bar{\alpha}_t}}( \mathbf{x}^{\text{tgt}}_t-\sqrt{1-\bar{\alpha}_t} \boldsymbol{\epsilon}_{\theta}(\mathbf{x}^{\text{tgt}}_t, \mathbf{x}^{\text{c}}, y, t)).
    \label{eq:one-step-mvd} 
\end{align}
This \emph{one-step} estimation of $\tilde{\mathbf{x}}_{0}^{\text{tgt}}$ can be noisy, especially when $t$ is large and $\mathbf{x}_t^{\text{tgt}}$ is extremely noisy. 
Thus, the iterative sampling of $\mathbf{x}_{t}^{\text{tgt}}$ is required until $t=0$. To sample next step $\mathbf{x}_{t-1}^{\text{tgt}}$, standard DDPM~\cite{Ho2020DDPM} computes the posterior mean $\mu_\theta$ from current $\mathbf{x}_{t}^{\text{tgt}}$ and estimated $\tilde{\mathbf{x}}_{0}^{\text{tgt}}$ at step $t$ with: 
\begin{equation}
    \mu_\theta(\rvx_t^\text{tgt}, t) :=\mu_{t-1}(\rvx_t^\text{tgt}, \tilde{\rvx}_0^\text{tgt}) = \frac{\sqrt{\alpha_{t}}\left(1-\bar{\alpha}_{t-1}\right)}{1-\bar{\alpha}_{t}} \rvx^{\text{tgt}}_{t} + \frac{\sqrt{\bar{\alpha}_{t-1}} \beta_{t}}{1-\bar{\alpha}_{t}}  \tilde{\rvx}_{0}^{\text{tgt}}, \text{ where } \beta_t=1-\alpha_t.
    \label{eq:xt_minus_one_from_xt_x0_old}
\end{equation}
Afterwards, $\mathbf{x}_{t-1}^{\text{tgt}}$ can be sampled from Gaussian distribution with mean $\mu_{t-1}$ and variance $\tilde{\beta}_{t-1}\mat{I}$ (\cref{eq:ddpm_reverse}) and used as the input for the next iteration. The reverse sampling is repeated until $t=0$ where 4 clear target views are generated.

Although multi-view diffusion models~\cite{liu2023syncdreamer, shi2023zero123++, Wang2023ImageDream} generate multiple views together, the 3D consistency across these views is not guaranteed due to the lack of an explicit 3D representation. 
Thus, we propose a novel 3D consistent diffusion model, which ensures the multi-view consistency at each step of the reverse process by diffusing 2D images using reconstructed 3D Gaussian Splats~\cite{Kerbl20233dgs}.

\section{\methodName}
\label{sec:method}
\begin{figure}[t!]
  \centering
  \includegraphics[width=\linewidth]{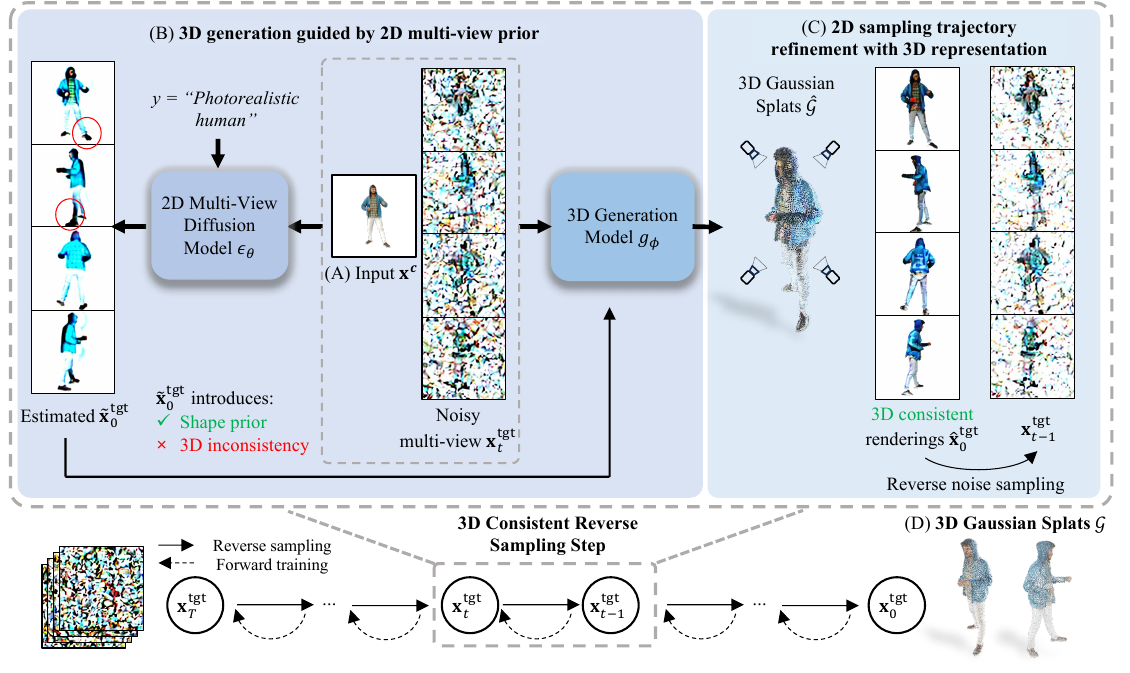}
  \caption{\textbf{Method Overview.} Given a single RGB image (A), we sample a realistic 3D avatar represented as 3D Gaussian Splats (D). At each reverse step, our 3D generation model $g_\phi$ leverages 2D multi-view diffusion prior from $\epsilon_\theta$ which provides a strong shape prior but is not 3D consistent (B, cf. \cref{subsec:MVR}). We then refine the 2D reverse sampling trajectory with generated 3D renderings that are guaranteed to be 3D consistent (C, cf. \cref{subsec:joint-sample}). Our tight coupling ensures 3D consistency at each sampling step and obtains a high-quality 3D avatar (D).    
  }
  \label{fig:pipeline}
\end{figure}

\paragraph{Overview.} Given a single RGB image, we aim to create a realistic 3D avatar consistent with the input. We adopt an image-conditioned 3D generation paradigm due to inherent ambiguities in the monocular view. We introduce a novel 3D Gaussian Splatting (3D-GS~\cite{Kerbl20233dgs}) generative model that combines shape priors from 2D multi-view diffusion models with the explicit 3D-GS representation. This allows us to jointly train our 3D generative model and a 2D multi-view diffusion model end-to-end and improves the 3D consistency of 2D multi-view generation at inference time. 

In this section, we first introduce our novel generative 3D-GS reconstruction model in~\cref{subsec:MVR}. We then describe how we leverage the 3D reconstruction to generate 3D consistent multi-view results by refining the reverse sampling trajectory (\cref{subsec:joint-sample}). An overview of our method can be found in \cref{fig:pipeline}.

\subsection{Generative 3D-GS Reconstruction with Diffusion Priors}
\label{subsec:MVR}
Given a context image $\mat{x}^c$, we use a conditional diffusion model to learn and sample from a plausible 3D distribution. Previous works demonstrated that 3D generation can be done implicitly via diffusing rendered images of a differentiable 3D represetation~\cite{anciukevicius2023renderdiffusion, karnewar2023holodiffusion, Tewari2023DiffusionWithForward} such as NeRF~\cite{Mildenhall2020NeRF, Yu2021pixelnerf}.
In this work, we introduce a novel generative model for 3D Gaussian Splatings~\cite{Kerbl20233dgs}, which diffuses rendered images of 3D-GS and enables sampling of 3D-GS at inference time. Single image to 3D generation is however very challenging, we hence propose to leverage 2D multi-view diffusion models in a tightly coupled manner which allows us to train it end-to-end with our novel 3D generative model. 

\paragraph{Generative 3D-GS Reconstruction.} In this work, we propose a 3D-GS generative model $g_\phi$, which is conditioned on input context image $\mat{x}^c$ to perform reconstruction of 3D Gaussian Splats $\gsplat$. 
Diffusing directly in the space of $\gsplat$ parameters requires pre-computing Gaussian Splats from scans, which is exorbitant. Instead, we diffuse the multi-view renderings of $\gsplat$ using a differentiable rendering function $\renderer$. \\ 
We denote $\mat{x}^\text{tgt}_0$ as the ground truth images at target views to be diffused and $\mat{x}^\text{novel}_0$ as the additional novel views for supervision. 
At training time, we uniformly sample a timestep $t\sim\mathcal{U}(0, T)$ and add noise to $\mat{x}^\text{tgt}_0$ using~\cref{eq:ddpm_forward} to obtain noisy target views $\mat{x}^\text{tgt}_t$. 
Our generative model $g_\phi$ takes $\mat{x}^\text{tgt}_t$, diffusion timestep $t$, and the conditional image $\mat{x}^c$ as input, and estimates 3D Gaussians $\hat{\gsplat}$:
\begin{equation}
    \hat{\gsplat} = g_\phi(\mat{x}^\text{tgt}_t, t, \mat{x}^c), \text{ where } \mat{x}^\text{tgt}_t=\sqrt{\Bar{\alpha}_t}\mat{x}_0^\text{tgt} + \sqrt{1- \Bar{\alpha}_t}\epsilon, \text{ and } \epsilon\sim\mathcal{N}(0, \mat{I})
    \label{eq:diffusion_wo_x0}
\end{equation}  
We adopt an asymmetric U-Net Transformer proposed by~\cite{Tang2024LGM} for $g_\phi$ to directly predict 3D-GS parameters from per-pixel features of the last U-Net layer. 
To supervise the generative model $g_\phi$, we use a differentiable rendering function $\renderer: \{\gsplat, \pi^\text{p}\}\mapsto \mat{x}^\text{p}$ to render images at target views $\pi^\text{tgt}$ and additional novel views $\pi^\text{novel}$. Denoting $\mat{x}_0:=\{\rvx^{\text{tgt}}_{0}, \rvx^{\text{novel}}_{0}\}$ as ground truth and $\hat{\mat{x}}_0:=\{\hat\rvx^{\text{tgt}}_{0}, \hat\rvx^{\text{novel}}_{0}\}$ as rendered images, we compute the loss on images and generated 3D-GS: 
\begin{equation}
    \begin{split}
        \mathcal{L}_{gs} = \lambda_1 \cdot \mathcal{L}_\text{MSE}\big(\mat{x}_0, \hat{\mat{x}}_0\big) 
    + \lambda_2 \cdot \mathcal{L}_\text{Percep}\big(\mat{x}_0, \hat{\mat{x}}_0\big) + \lambda_3 \cdot \mathcal{L}_\text{reg}(g_\phi(\mat{x}^\text{tgt}_t, t, \mat{x}^c)) ,\\
    \text{ where } \hat{\mat{x}}_0 := \{\hat{\mat{x}}^\text{tgt}_0, \hat{\mat{x}}^\text{novel}_0\} = \renderer(g_\phi(\mat{x}^\text{tgt}_t, t, \mat{x}^c), \{\pi^\text{tgt}, \pi^\text{novel}\}),
    \end{split}
    \label{eq:loss_3D_diffusion}
\end{equation}
here $\mathcal{L}_\text{MSE}$ denotes the Mean Square Error (MSE) and $\mathcal{L}_\text{Percep}$ is the perceptual loss based on VGG-19~\cite{Simonyan2015vgg}. We also apply $\mathcal{L}_\text{reg}$, a geometry regularizer~\cite{Huang20242dgs, Yu2024gof} to stabilize the generation of $\hat{\gsplat}$.

With this, we can train a generative model that diffuses 3D-GS \emph{implicitly} by diffusing 2D images $\mat{x}^\text{tgt}_t$. At inference time, we can generate 3D-GS given the input image by denoising 2D multi-views sampled from Gaussian distribution. We initialize $\mat{x}^\text{tgt}_T$ from $\mathcal{N}(0, \mathbf{I})$, and iteratively denoise the rendered images of predicted $\hat{\gsplat}$ from our model $g_\phi$. 
At each reverse step, our model $g_\phi$ estimates a clean state $\hat{\gsplat}$ and render target images $\hat{\mat{x}}^\text{tgt}_0$. 
We then calculate target images $\mat{x}^\text{tgt}_{t-1}$ for the next step via \cref{eq:xt_minus_one_from_xt_x0_old} and repeat the process until $t=0$. For more details, please refer to~\cref{sec_supp:3d_state_diffusion}

\begin{figure}[!h]
    \begin{minipage}[t]{0.25\textwidth} 
    \strut\vspace*{0em}\newline
        \centering
         \begin{tabular}{ l c}
            \hline
            Method & PSNR $\uparrow$ \\
             \hline
            w/o $ \tilde{\mathbf{x}}^{\text{tgt}}_{0} $ & $14.45$ \\
            \hline
             \emph{Ours} & $\mathbf{16.12}$  \\
             \hline
        \end{tabular}
    \end{minipage}
    \hfill
  \begin{minipage}[t]{0.72\textwidth}
  \strut\vspace*{-\baselineskip}\newline 
    \centering
      \includegraphics[width=\linewidth]{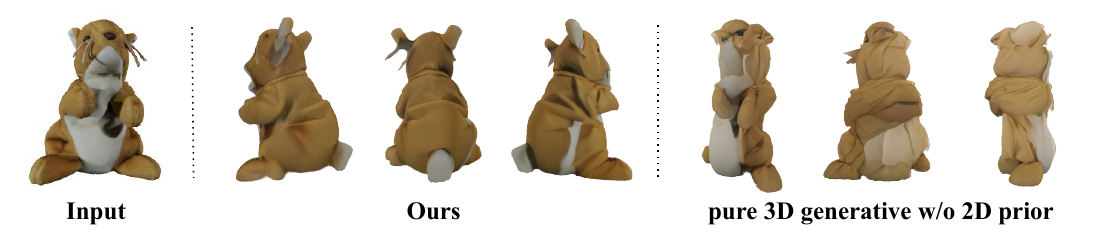}

  \end{minipage}
    \caption{2D multi-view priors $\tilde{\rvx}_0^\text{tgt}$ enhances generalization to general objects in GSO~\cite{downs2022gso} dataset.}
    \label{fig:ablate_multiview_cond_obj_gso}
\end{figure}

Our generative 3D-GS reconstruction model archives superior performance on in-distribution human reconstruction yet generalizes poorly to unseen categories such as general objects (\cref{subsec:ablation} ~\cref{fig:ablate_multiview_cond_obj_gso}). Our key insight for better generalization is leveraging strong priors from pretrained 2D multi-view diffusion models for 3D-GS generation. 

\paragraph{3D-GS Generation with 2D Multi-view Diffusion.} 
Pretrained 2D multi-view diffusion models (MVD)~\cite{liu2023syncdreamer, shi2023mvdream, Wang2023ImageDream} have seen billions of real images~\cite{Schumann2022Laion5B} and millions of 3D data~\cite{Deitke2023Objaverse}, which provide strong prior information and can generalize to unseen objects~\cite{Tang2024LGM, xu2024instantmesh}. Here, we propose a simple yet elegant idea for incorporating this multi-view prior into our generative 3D-GS model $g_\phi$. We can also leverage generated 3D-GS to guide 2D MVD sampling process which we discuss in~\cref{subsec:joint-sample}.\\
Our key observation is that both 2D MVD and our proposed 3D-GS generative model are diffusion-based and share the same sampling state $\mathbf{x}_{t}^{\text{tgt}}$ at timestep $t$. Thus, they are \emph{synchronized}. This enables us to couple and facilitate information exchange between 2D MVD $\boldsymbol{\epsilon}_{\theta}$ and 3D-GS generative model $g_{\phi}$ at the same diffusion timestep $t$.  
To inject the 2D diffusion priors into 3D generation, we first compute \emph{one-step} estimation of $\tilde{\mathbf{x}}^{\text{tgt}}_0$ (\cref{eq:one-step-mvd}) using 2D MVD $\epsilon_\theta$, and condition our 3D-GS generative mode $g_{\phi}$ additionally on it. Formally, our 3D-GS generative model enhanced with 2D multi-view diffusion priors is written as:
\begin{align}
    \hat{\gsplat} = g_\phi(\mat{x}^\text{tgt}_t, t, \mat{x}^c, \Tilde{\mat{x}}^\text{tgt}_0), \text{ where } \tilde{\mathbf{x}}_{0}^{\text{tgt}} = \frac{1}{\sqrt{\bar{\alpha}_t}}( \mathbf{x}^{\text{tgt}}_t-\sqrt{1-\bar{\alpha}_t} \boldsymbol{\epsilon}_{\theta}(\mathbf{x}^{\text{tgt}}_t, \mathbf{x}^{\text{c}}, y, t))
    \label{eq:diffusion_with_x0_clean}
\end{align}
The visualization of $\tilde{\mathbf{x}}^{\text{tgt}}_0$ along the whole sampling trajectory in~\cref{fig:ddim_intermediate_visualization} shows that the pretrained 2D diffusion model $\boldsymbol{\epsilon}_{\theta}$ can already provide useful multi-view shape prior even in large timestep $t=1000$. This is further validated in our experiments where the additional 2D diffusion prior $\tilde{\mathbf{x}}^{\text{tgt}}_0$ leads to better avatar reconstruction (\cref{tab:ablate_multiview_cond_human}) as well as more robust generalization to general objects (\cref{fig:ablate_multiview_cond_obj_gso}). By utilizing the timewise iterative manner of 2D and 3D diffusion models, we can not only leverage 2D priors for 3D-GS generation but also train both models jointly end to end, which we discuss next. 

\paragraph{Joint Training with 2D Model.} We adopt pretrained ImageDream~\cite{Wang2023ImageDream} as our 2D multi-view diffusion model $\epsilon_\theta$ and jointly train it with our 3D-GS generative model $g_\phi$. We observe that our joint training is important for coherent 3D generation, as opposed to prior works that frozen pretrained 2D multi-view models~\cite{Tang2024LGM, tochilkin2024triposr}. 
We summarize our training algorithm in ~\cref{alg:training}. 
We combine the loss of 2D diffusion and our 3D-GS generation loss $\mathcal{L}_{gs}$(~\cref{eq:loss_3D_diffusion}): 
\begin{equation}
    \begin{split}
    \mathcal{L}_{total} = \mathcal{L}_\text{MSE}(\boldsymbol{\epsilon}, \boldsymbol{\epsilon}_{\theta}) + \mathcal{L}_{gs}
    \end{split}
    \label{eq:loss_all}
\end{equation}
Once trained, one can sample a plausible 3D-GS avatar $\gsplat$ conditioned on the input image from the learned 3D distributions. 
However, we observe that the multi-view diffusion model $\boldsymbol{\epsilon}_{\theta}$ can still output inconsistent multi-views along the sampling trajectory (see~\cref{fig:pipeline}). On the other hand, our 3D generator produces explicit 3D-GS which can be rendered as 3D consistent multi-views. Our second key idea is to use the 3D consistent renderings to guide 2D sampling process for more 3D consistent mulit-view generation. We discuss this in~\cref{subsec:joint-sample}.

\algrenewcommand\algorithmicrequire{\textbf{Input:}} 
\algrenewcommand\algorithmicensure{\textbf{Output:}}
\subsection{Guide 2D Multi-view Sampling with Reconstructed 3D-GS}
\label{subsec:joint-sample} 
\algrenewcommand\algorithmicindent{0.5em}%
\begin{figure}[t]
\begin{minipage}[t]{0.49\textwidth}
\begin{algorithm}[H]
  \caption{Training} \label{alg:training}
  \small
  \begin{algorithmic}[1]
  \Require Dataset of posed multi-view images $\rvx^{\text{tgt}}_{0}$, $\pi^{\text{tgt}}$, $\rvx^{\text{novel}}_{0}$, $\pi^{\text{novel}}$, a context image $\rvx^{\text{c}}$, text description $y$
  \Ensure Optimized 2D multi-view diffusion model
  $\epsilon_{\theta}$ and 3D-GS generative model $g_{\phi}$
    \Repeat
      \State $\{\rvx^{\text{tgt}}_{0}, \rvx^{\text{novel}}_{0}, \rvx^{\text{c}}, y\} \sim q(\{\rvx^{\text{tgt}}_{0}, \rvx^{\text{novel}}_{0}, \rvx^{\text{c}}, y\} )$
      \State $t \sim \mathrm{Uniform}(\{1, \dotsc, T\})$; $\boldsymbol{\epsilon} \sim \mathcal{N}(\mathbf{0},\mathbf{I})$
      \State $\mathbf{x}^{\text{tgt}}_{t} = \sqrt{\bar\alpha_t} \rvx_{0}^{\text{tgt}} + \sqrt{1-\bar{\alpha}_t}\boldsymbol{\epsilon}$
      \State $\tilde{\mathbf{x}}^{\text{tgt}}_0 = \frac{1}{\sqrt{\bar{\alpha}_t}}( \mathbf{x}^{\text{tgt}}_t-\sqrt{1-\bar{\alpha}_t} \boldsymbol{\epsilon}_{\theta}(\mathbf{x}^{\text{tgt}}_t, \mathbf{x}^{\text{c}}, y, t))$ 
      \State $\hat{\gsplat} = g_{\phi}\left(\mathbf{x}_t^{\text{tgt}}, t, \mathbf{x}^{\text{c}}, \tilde{\mathbf{x}}_{0}^{\text{tgt}}\right) $ 
      \textcolor{gray}{// Enhance conditional 3D generation with 2D diffusion prior $\tilde{\mathbf{x}}_{0}^{\text{tgt}}$ from $\mat{\epsilon}_\theta$}
      \State $\{ \hat{\mathbf{x}}_{0}^{\text{tgt}}, \hat{\mathbf{x}}_{0}^{\text{novel}}\} = \renderer\left(\hat{\gsplat}, \{\pi^{\text{tgt}}, \pi^{\text{novel}}\} \right)$
      \State Compute loss $\mathcal{L}_{total}$ (~\cref{eq:loss_all}) 
      \State Gradient step to update $\mat{\epsilon}_\theta, g_\phi$

    \Until{converged}
  \end{algorithmic}
  \label{algm:train}
\end{algorithm}
\end{minipage}
\hfill
\begin{minipage}[t]{0.49\textwidth}
\begin{algorithm}[H]
  \caption{3D Consistent Sampling} \label{alg:sampling}
  \small
  \begin{algorithmic}[1]
  \Require A context image $\rvx^c$ and text $y$; Converged 2D diffusion model $\epsilon_{\theta}$ and 3D generative model $g_{\phi}$
  \Ensure A 3D Gaussian Avatar $\gsplat$ of the 2D image $\rvx^c$
  
    \vspace{.08in}
    \State $\rvx_T^{\text{tgt}} \sim \mathcal{N}(\mathbf{0}, \mathbf{I})$
    \For{$t=T, \dotsc, 1$}
      \State $\tilde{\mathbf{x}}_{0}^{\text{tgt}} = \frac{1}{\sqrt{\bar{\alpha}_t}}( \mathbf{x}^{\text{tgt}}_t-\sqrt{1-\bar{\alpha}_t} \boldsymbol{\epsilon}_{\theta}(\mathbf{x}^{\text{tgt}}_t, \mathbf{x}^{\text{c}}, y, t))$ 
      \State  $\hat{\gsplat} = g_{\phi}\left(\mathbf{x}_t^{\text{tgt}}, t, \mathbf{x}^{\text{c}}, \tilde{\mathbf{x}}_{0}^{\text{tgt}}\right) $
      \State $\hat{\mathbf{x}}_{0}^{\text{tgt}} = \renderer\left(\hat{\gsplat}, \pi^{\text{tgt}}\right)$
      \State $\mu_{t-1}(\rvx_t^\text{tgt}, \hat{\rvx}_0^\text{tgt}) = \frac{\sqrt{\alpha_{t}}\left(1-\bar{\alpha}_\text{t-1}\right)}{1-\bar{\alpha}_{t}} \rvx^{\text{tgt}}_{t} + \frac{\sqrt{\bar{\alpha}_\text{t-1}} \beta_{t}}{1-\bar{\alpha}_{t}} \hat{\rvx}_{0}^{\text{tgt}}$ \textcolor{gray}{// Guide 2D sampling with 3D consistent renderings}
      \State $\rvx^{\text{tgt}}_{t-1} \sim \mathcal{N}\left(\mathbf{x}^{\text{tgt}}_{t-1}; \tilde{\bm{\mu}}_{t}\left(\rvx^{\text{tgt}}_t, \hat\rvx^{\text{tgt}}_{0} \right), \tilde{\beta}_{t-1}\mathbf{I}) \right)$
    \EndFor
    \vspace{.1in}
    \State \textbf{return} $\gsplat =  g_{\phi}\left(\mathbf{x}_{0}^{\text{tgt}}, \tilde{\mathbf{x}}_{0}^{\text{tgt}}, \mathbf{x}^{\text{c}}, t=0\right) $
     \vspace{.015in}
  \end{algorithmic}
\end{algorithm}
\label{algm:sample}
\end{minipage}
\vspace{-1em}
\end{figure}

With the shared and \emph{synchronized} sampling state $\mat{x}_t^\text{tgt}$ of 2D multi-view diffusion model $\boldsymbol{\epsilon}_{\theta}$ and 3D-GS reconstruction model $g_\phi$, we couple both models at arbitrary $t$ during training. Similarly, they are also connected by both using estimated clean multi-views $\mat{x}_0^\text{tgt}$ at sampling time. To leverage the full potential of both models, we carefully design a joint sampling process that utilizes the reconstructed 3D-GS $\hat\gsplat$ at each timestep $t$ to guide 2D multi-view sampling, which is summarized in~\cref{alg:sampling}. \\
We observe that the key difference between the clean multi-views estimated $\mat{x}_0^\text{tgt}$ from 2D diffusion model and our 3D-GS generation lies in 3D consistency: 2D MVD computes multi-view $\tilde{\mat{x}}_0^\text{tgt}$ from 2D network prediction which can be 3D inconsistent while our $\hat{\mat{x}}_0^\text{tgt}$ are rendered from explicit 3D-GS representation which are guaranteed to be 3D consistent. Our idea is to guide the 2D multi-view reverse sampling process with our 3D consistent renderings $\hat{\mat{x}}_0^\text{tgt}$ such that the 2D sampling trajectory is more 3D consistent. Specifically, we leverage 3D consistent multi-view renderings $\hat{\mat{x}}_0^\text{tgt}$ to refine the posterior mean $\mu_\theta(\rvx_t^\text{tgt}, t)$ at each reverse step: 
\begin{align}
\begin{split}
    \text{Original: } \mu_\theta(\rvx_t^\text{tgt}, t) :=\mu_{t-1}(\rvx_t^\text{tgt}, \tilde{\rvx}_0^\text{tgt}) \quad \rightarrow \quad \text{Ours: } \mu_\theta(\rvx_t^\text{tgt}, t) :=\mu_{t-1}(\rvx_t^\text{tgt}, \hat{\rvx}_0^\text{tgt}),   \\
    \text{ where } \hat{\rvx}_{0}^{\text{tgt}}=\renderer(\hat{\gsplat}, \pi^\text{tgt}), \text{ and } \mu_{t-1}(\rvx_t^\text{tgt}, \hat{\rvx}_0^\text{tgt}) = \frac{\sqrt{\alpha_{t}}\left(1-\bar{\alpha}_{t-1}\right)}{1-\bar{\alpha}_{t}} \rvx^{\text{tgt}}_{t} + \frac{\sqrt{\bar{\alpha}_{t-1}} \beta_{t}}{1-\bar{\alpha}_{t}}  \hat{\rvx}_{0}^{\text{tgt}} 
\end{split}
    \label{eq:xt_minus_one_from_xt_x0_ours}
\end{align}

With this refinement, we guarantee the 3D consistency at each reverse step $t$ and avoid 3D inconsistency accumulation in original multi-view sampling~\cite{Wang2023ImageDream}. 
In~\cref{fig:ddim_intermediate_visualization}, we visualize the evolution of originally generated multi-views $\tilde{\mathbf{x}}^{\text{tgt}}_0$ and multi-views rendering $\hat{\mathbf{x}}^{\text{tgt}}_0$ from generated 3D-GS $\hat\gsplat$ along the whole reverse sampling process. It intuitively shows how effective the sampling trajectory refinement is. We perform extensive ablation in \cref{subsec:ablation} showing the importance of the consistent refinement for sampling trajectory.

\section{Experiments}
\label{sec:experiments}
In this section, we first compare against baseline methods for human reconstruction in~\cref{subsec:compare-baselines} and then ablate our design choices in~\cref{subsec:ablation}. 
\subsection{Experimental Setup}
\label{subsec:exp-setup}
\textbf{Datasets.} We train our model on a combined 3D human dataset~\cite{axyz, treedy, twindom, renderpeople, han20232k2k, ho2023customhuman, su2022thuman3, tao2021thuman2} compromising  $\sim6000$ high quality scans. We evaluate qualitatively and quantitatively on CAPE~\cite{ma2019cape, ponsmoll2017clothcap, zhang2017buff}, Sizer~\cite{antic2024closed, tiwari2020sizer} and IIIT~\cite{jinka2023iiit} dataset. Please refer to~\cref{suppsec:traninigdataset} and~\cref{suppsec:evaluationdataset} for more details. 

\textbf{Implementation Details.}
We trained our model on 8 NVIDIA A100 GPUs over approximately 5 days. Each GPU was configured with a batch size 2 and gradient accumulations of 16 steps to achieve an effective batch size of 256. For more training details regarding hyperparameters, diffusion schedulers, etc., please refer to~\cref{suppsec:training_details} for more details. Our model creates 3D Avatar from single images in 22.6 seconds on a A100 GPU and only consumes 11.7 GB VRAM, which allows the efficient large-scale avatar generation.

\textbf{Evaluation Metrics.} We evaluate the geometry quality using Chamfer Distance ($\text{CD}$ in ${cm}$), Point-to-Surface Distance ($\text{P2S}$ in ${cm}$), F-score~\cite{Tatarchenko2019fscore} (w/ threshold of $0.01m$), and Normal Consistency (NC) between the extracted mesh (\cref{suppsec:mesh_via_tsdf}) and the groundtruth scan. 
Appearance quality is assessed by rendering the reconstructed avatar from 32 novel views with uniform azimuth and 0 elevation angle.
The metrics for appearance reported include multi-scale Structure Similarity (SSIM)~\cite{wang2003ssim}, Learned Perceptual Image Patch Similarity (LPIPS)~\cite{zhang2018lpips}, and Peak Signal to Noise Ratio (PSNR) between rendered and ground-truth views. 
Moreover, we report the Fr\'echet inception distance (FID)~\cite{heusel2017fid} which reflects the quality and realism of the unseen regions.

\subsection{Realistic Avatar from Image}
\label{subsec:compare-baselines}
\begin{figure}[t!]
  \centering
  \includegraphics[width=\textwidth]{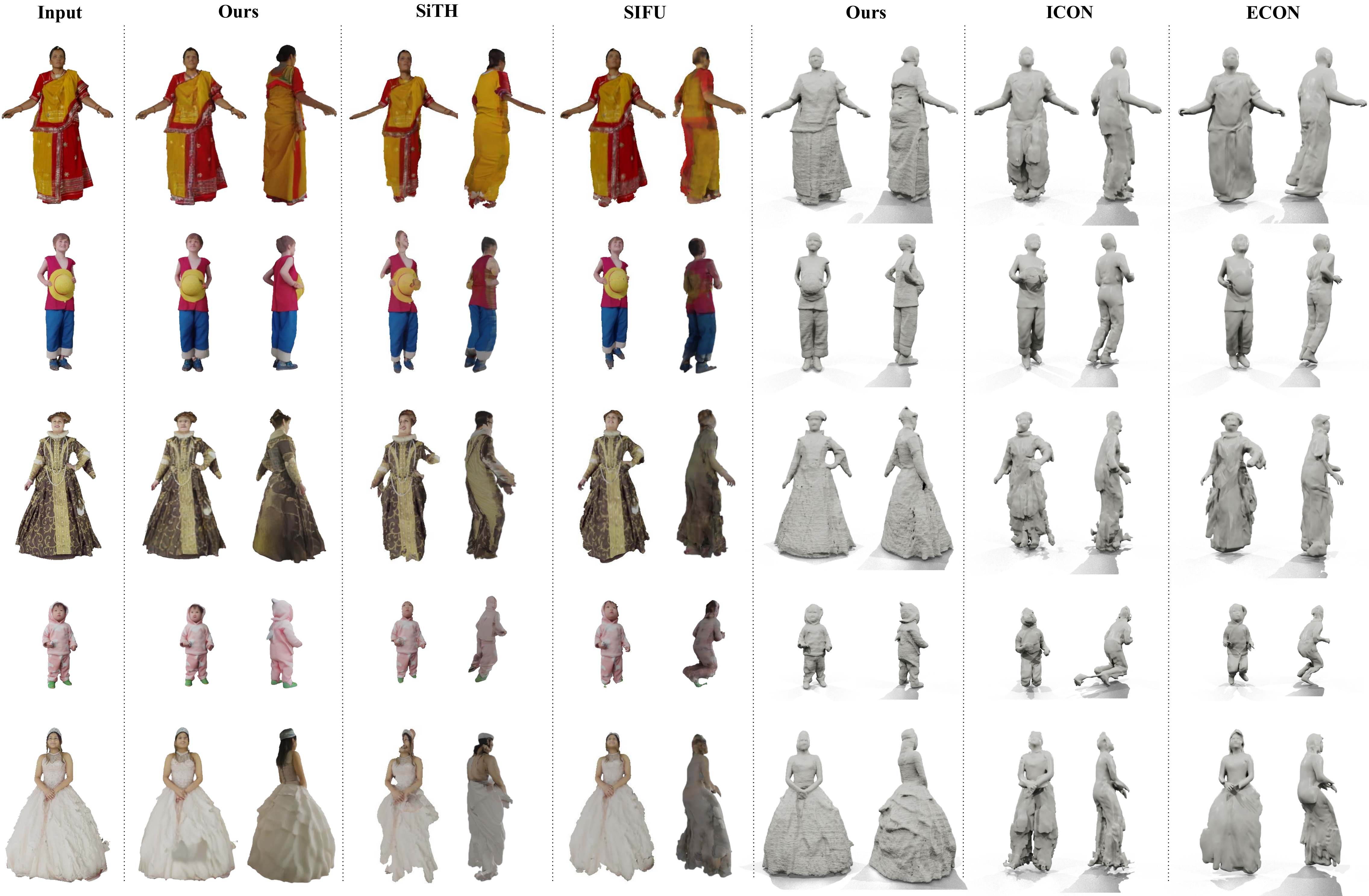}\hfill
  \caption{\textbf{Qualitative comparison with baselines}. Recent avatar reconstruction works ICON~\cite{xiu2023icon}, ECON~\cite{xiu2023econ}, SiTH~\cite{ho2023sith} and SIFU~\cite{zhang2023sifu}) cannot reconstruct loose clothing coherently. Additionally, SiTH and SIFU generate blurry texture in unseen regions due to their deterministic formulation of regressing 3D avatar directly from single RGB imagse. In contract, our method is able to reconstruct avatars with realistic textures and plausible 3D geometry in both seen and unseen region. 
  }
  \label{fig:compare_baselines}
  
\end{figure}
We compare our approach against prior methods for image-to-avatar reconstruction, including template-based~\cite{feng2022fof, ho2023sith, xiu2023icon, xiu2023econ, saito2019pifu, zhang2023sifu}, template-free~\cite{saito2019pifu} human reconstruction methods, as well as general image-to-3D methods~\cite{Tang2024LGM, tochilkin2024triposr, xu2024instantmesh}. To further assess performance, we also fine-tuned the state-of-the-art object reconstruction method LGM~\cite{Tang2024LGM} and its deployed multi-view diffusion model~\cite{Wang2023ImageDream} on our training data, denoted as $\text{LGM}_{\text{human}}$. Quantitative evaluations reported in~\cref{tab:compare_baselines_all} demonstrate that our proposed method excels in reconstructing realistic avatars with accurate geometry (CD, NC, F-score) and realistic texture (SSIM, LPIPS, PSNR, FID) from a single RGB image. 

We present qualitative comparison examples in~\cref{fig:compare_baselines} and~\cref{secsupp:comparison_img23d}, highlighting the strengths and weaknesses of competing methods. Template-based methods such as SiTH~\cite{ho2023sith} and SIFU~\cite{zhang2023sifu} struggle to accurately reconstruct the geometry of loose clothing (as shown in row 4) due to their reliance on the naked SMPL body shape. In contrast, template-free methods like PIFu~\cite{saito2019pifu} and TripoSR~\cite{tochilkin2024triposr} offer greater flexibility and better performance on loose clothing. However, they are not generative models and their deterministic formulations lead to blurry textures in unseen regions, as they tend to produce average textures rather than distinct details. 
Similar to our approach, LGM~\cite{Tang2024LGM} and InstantMesh~\cite{xu2024instantmesh} utilize 2D diffusion models to generate multi-view images and perform sparse-view 3D reconstruction. Nonetheless, their separation of 2D and 3D models cannot correct the 3D inconsistencies that may arise from the 2D models. 
Even further fine-tuning of LGM on human scans (\cref{fig:ablate_consistent_diff}) does not adequately address these challenges due to the complex and sensitive nature of human geometry and textures.
In contrast, our conditional generative formulation and inherent 3D consistency by tightly coupling 2D-3D models allow us to obtain accurate reconstruction in front view and realistic generation in unseen regions. 
 We also show the generative power of our method in~\cref{suppsec:generative_power}: by sampling with different seed, we obtain diverse yet plausible reconstruction.
 
\begin{figure}[t!]
  \centering
  \includegraphics[width=\textwidth]{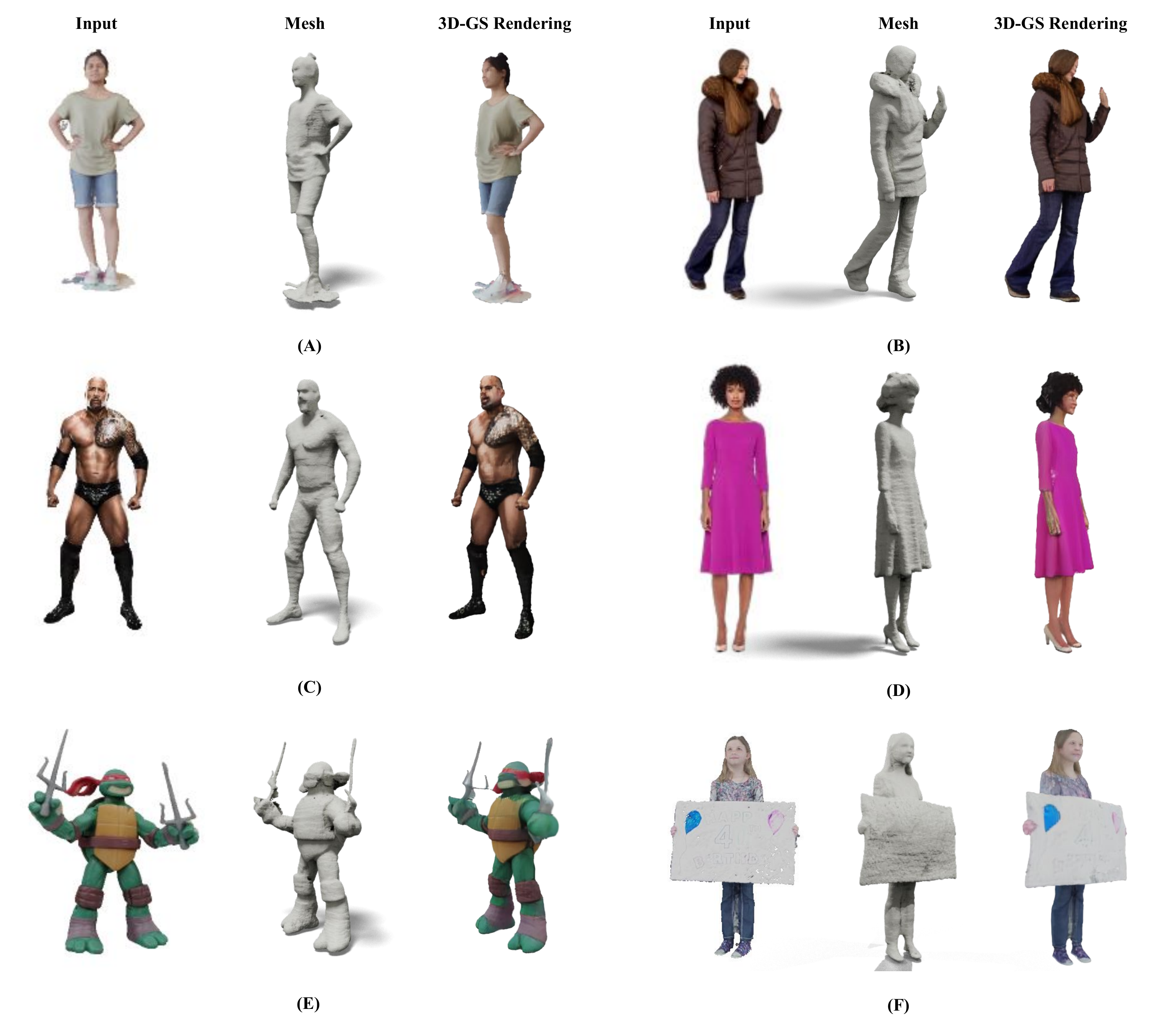}\hfill
  \caption{Visualization of reconstructed mesh and synthesized novel view of generated 3D-GS on subjects from Sizer~\cite{tiwari2020sizer}, RenderPeople~\cite{renderpeople}, Twindom~\cite{twindom}, UBC Fashion~\cite{zablotskaia2019ubcfashion}, GSO~\cite{downs2022gso} and  online image. More results are presented in~\cref{suppsec:moreresults} and our \href{https://yuxuan-xue.com/human-3diffusion}{project page}.
  }
  \label{fig:more_visualization}
\end{figure}

Please also refer to~\cref{fig:more_visualization},~\cref{suppsec:moreresults} and our \href{https://yuxuan-xue.com/human-3diffusion}{project page} for additional reconstruction results on challenging subjects not previously observed, encompassing a diverse range of appearances such as loose skirts and custom suits, as well as accessories like bags and gloves.

\begin{table*}[t!]
  \centering
\begin{tabular}{ l c c c c c c c c }
 \hline
Method & {$\text{CD}$ $\downarrow$} & {$\text{P2S}$ $\downarrow$} & {F-score $\uparrow$}   & $\text{NC} \uparrow$ & {SSIM $\uparrow$} & { LPIPS $\downarrow$}  &{PSNR $\uparrow$} &{FID $\downarrow$} \\
     
 \hline
 PIFu~\cite{saito2019pifu}   &  $2.75$  & $2.68$ & $0.359$  & $0.778$ & $ 0.909 $ & $0.077$ & $21.06$  &$ 29.57 $ \\
SiTH~\cite{ho2023sith}   &  $ 4.00 $ & $4.00$ & $ 0.257 $ &  $0.749$  & $ 0.907 $ & $ 0.073 $ & $ 20.00 $  &$ 22.33 $ \\
SIFU~\cite{zhang2023sifu}   &  $ 3.50 $ & $3.50$ & $ 0.273 $  &   $0.760$ & $ 0.0.900 $ & $ 0.081 $ & $20.73$   &$ 40.75 $ \\
$\text{LGM}_{\text{human}}$   & $3.44$ & $3.44$ & $0.272$ & $0.560$  & $0.893$ & $0.088$ & $20.56$  &$ 14.22 $ \\
\hline
FoF~\cite{feng2022fof} & $5.43$ & $5.34$ & $0.183$ & $0.683$ & - & - & - & - \\
 ICON~\cite{xiu2023icon} & $3.88$ & $3.94$ & $0.244$ & $0.749$ & - & - & - & - \\
ECON~\cite{xiu2023econ} & $2.83$ & $3.61$ & $0.291$ & $0.767$ & - & - & - & - \\
\hline
 \emph{Ours} &  $\textbf{1.41}$  & $\textbf{1.37}$ & $\textbf{0.557}$  &  $\textbf{0.791}$ & $\textbf{0.916}$ & $\textbf{0.058}$ & $\textbf{21.61}$  &$ \textbf{8.45} $\\
 \hline
\end{tabular}
  \caption{\textbf{Quantitative evaluation} on CAPE~\cite{ma2019cape}, SIZER~\cite{tiwari2020sizer}, and IIIT~\cite{jinka2023iiit} dataset. Our method can perform better reconstruction in terms of more accurate geometry (CD, P2S, F-score, NC) and realistic textures (SSIM, LPIPS, PSNR, FID).}
\label{tab:compare_baselines_all}
\end{table*}

\subsection{Ablation Studies}
\label{subsec:ablation}
\begin{figure}[t!]
  \centering
  \includegraphics[width=\textwidth]{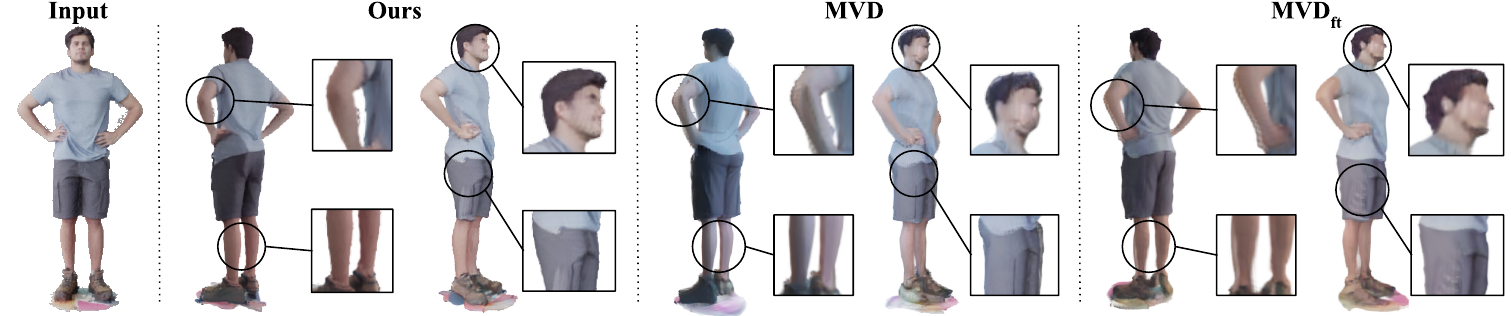}
  \caption{\textbf{3D reconstruction conditioned on different multi-view priors.} Without our 3D-consistent sampling, the 2D diffusion model cannot generate 3D consistent multi-views (MVD, $\text{MVD}_\text{ft}$), leading to artifacts like floating 3D Gaussians splats.}
  \label{fig:ablate_consistent_diff}
\end{figure}

\renewcommand{\captionlabelfont}{\small}
\setlength{\intextsep}{0.1cm} 
\begin{wraptable}{r}{0.45\textwidth} 
\centering
\small
\begin{tabular}{ l c c c }
\hline
{Method} & { LPIPS $\downarrow$} & { SSIM $\uparrow$}  &{PSNR $\uparrow$}  \\     
 \hline
MVD  & $0.078$ & $0.911$ & $22.32$   \\
$\text{MVD}_\text{ft}$ & $0.061$ & $0.926 $ &  $24.14$\\
\hline
 \emph{Ours} &  $\textbf{0.048}$ & $\textbf{0.934}$ & $\textbf{24.69}$ \\
 \hline
\end{tabular}
\caption{\small\textbf{Evaluating trajectory refinement} for 2D multi-view diffusion. Our proposed refinement improves multi-view image quality.}
\label{tab:ablate_traj_refine_for_2d}
\end{wraptable}
\setlength{\intextsep}{0cm} 
\renewcommand{\captionlabelfont}{\normalsize}

\textbf{Importance of Trajectory Refinement.} 
One of our key ideas is leveraging our explicit 3D model to refine the 2D multi-view reverse sampling trajectory, ensuring 3D consistency in Multi-View Diffusion (MVD) generation (see~\cref{subsec:joint-sample} and ~\cref{eq:xt_minus_one_from_xt_x0_ours}).
To evaluate this, we compare the multi-view images generated by pretrained MVD, fine-tuned MVD on our data ($\text{MVD}_\text{ft}$) and MVD with our 3D consistent sampling (ours), as shown in \cref{tab:ablate_traj_refine_for_2d}. The results demonstrate that our proposed method effectively enhances the quality of generated multi-view images by leveraging the explicit 3D model to refine sampling trajectory. 
Additionally. we analyze the 3D reconstruction results with the multi-view images generated by these models in \cref{fig:ablate_consistent_diff}. MVD and $\text{MVD}_{\text{ft}}$ produce inconsistent multi-view images, which typically lead to floating Gaussian and hence blurry boundaries. In contrast, our method can generate more consistent multi-views, result in better 3D Gaussians Splats and sharper renderings. 

\renewcommand{\captionlabelfont}{\footnotesize}
\setlength{\intextsep}{0.1cm} 
\begin{wraptable}{r}{0.8\textwidth} 
\centering
\footnotesize
\begin{tabular}{ p{2.3cm} | >{\centering\arraybackslash}p{1.1cm} >{\centering\arraybackslash}p{1.15cm} >{\centering\arraybackslash}p{0.7cm} >{\centering\arraybackslash}p{0.95cm} >{\centering\arraybackslash}p{0.95cm} >{\centering\arraybackslash}p{0.95cm}}
\hline
{Method} & {$\text{CD}_\text{(cm)}$$\downarrow$} & {F-score$\uparrow$}   & $\text{NC}\uparrow$ & { LPIPS$\downarrow$} & {SSIM$\uparrow$}  &{PSNR$\uparrow$}  \\     
\hline
Our w/o Traj. Ref. & $1.57$ & $0.498$ & $0.794$ & $0.064$ & $0.908 $ &  $21.09$\\
\hline
 \emph{Ours} &  $\textbf{1.35}$ &  $\textbf{0.550}$ &  $\textbf{0.798}$ & $\textbf{0.060}$ & $\textbf{0.918}$ & $\textbf{21.49}$ \\
 \hline
\end{tabular}
\caption{\footnotesize\textbf{Evaluating trajectory refinement} for final 3D reconstruction. Our sampling trajectory refinement ensures multi-view consistency and hence yields better 3D results.}
\label{tab:ablate_traj_refine_for_3d}
\end{wraptable}
\setlength{\intextsep}{0cm} 
\renewcommand{\captionlabelfont}{\normalsize}
We further quantitatively evaluate the impact of our proposed sampling trajectory refinement on final 3D reconstruction in~\cref{tab:ablate_traj_refine_for_3d}. We compare the reconstruction results of methods with and without our trajectory refinement while using the same 2D MVD and 3D reconstruction models. It can be clearly seen that our trajectory refinement improves the quality of 3D reconstruction.

\paragraph{Importance of 2D Multi-view Prior.} Another key idea of our work is the use of multi-view priors $\tilde{\rvx}_0^\text{tgt}$ from 2D diffusion model pretrained on massive data~\cite{Deitke2023Objaverse, Rombach2022StableDiffusion, Schumann2022Laion5B} to enhance our 3D generative model. This additional prior information is pivotal for ensuring accurate reconstruction of both in-distribution human dataset and generalizing to out-of-distribution objects. 
\renewcommand{\captionlabelfont}{\small}
\setlength{\intextsep}{0.0cm} 
\begin{wraptable}{r}{0.3 \textwidth}
    \centering
    \small
    \begin{tabular}{ l c}
    \hline
    Method & {PSNR $\uparrow$}  \\
    \hline
    Ours w/o $ \tilde{\mathbf{x}}^{\text{tgt}}_{0} $ & $20.98$ \\
    \hline
     \emph{Our full model} &  $\mathbf{21.49}$  \\
     \hline
    \end{tabular}
    \caption{\small \textbf{2D multi-view priors $ \tilde{\mathbf{x}}^{\text{tgt}}_{0} $} improve human reconstruction quality. 
    }
\label{tab:ablate_multiview_cond_human}
\end{wraptable}
\setlength{\intextsep}{0cm} 
\renewcommand{\captionlabelfont}{\normalsize}
We evaluate the performance of our 3D model $g_{\phi}$ by comparing generation results with and without the 2D diffusion prior $\tilde{\mathbf{x}}_{0}^{\text{tgt}}$ (refer to \cref{eq:diffusion_with_x0_clean} and \cref{eq:diffusion_wo_x0}). Notably, without the 2D multi-view conditioning, the alignment of the generated 3D model in the front view is not guaranteed due to the relative camera pose settings in our 3D generative model $g_{\phi}$. Therefore, we evaluate the overall quality solely through the Fréchet Inception Distance (FID).

For avatars reconstruction, our powerful 3D reconstruction model can already achieve state-of-the-art performance. Moreover, our full model with multi-view prior $\tilde{\rvx}_0^\text{tgt}$ generates avatars with higher quality as demonstrated in \cref{tab:ablate_multiview_cond_human}. We further evaluate it on the GSO~\cite{downs2022gso} dataset which consists of unseen general objects to our model. 
The improvements are even more pronounced in this setting, highlighting the challenges of generating coherent 3D structures from a single 2D image, particularly with unseen objects. For additional examples, please see \cref{fig:ablate_multiview_cond_obj_supp} in Supp..

\section{Limitations and Future Work}
\label{sec:limitation}
Currently, our method is constrained by the $256\times256$ resolution of the multi-view diffusion model, which restricts the sharpness of texture details (see \cref{sec:limitations}). 
Upgrading to more powerful high-resolution ($512\times512$) multi-view diffusion models~\cite{gao2024cat3d, tang2023mvdiffusion++} could potentially resolve these issues. Moreover, our approach may struggle in reconstructing subjects with challenging poses, as we further discussed in~\cref{sec:limitations}. Synthesizing training data with challenging poses~\cite{Black_bedlam_2023, xie2023template_free} could be a potential solution.

Our method is a general framework for image-to-3D reconstruction, 
which is applicable to various objects and compositional shapes like human-object interactions. We leave these to future works. 

\section{Conclusion}
\label{sec:conclusions}
In this paper, we introduce \textbf{Human-3Diffusion}, a 3D consistent diffusion model for creating realistic avatars from single RGB images. Our key ideas are two folds: 1) Leveraging strong multi-view priors from pretrained 2D diffusion models to generate 3D Gaussian Splats, and 2) Using the reconstructed explicit 3D Gaussian Splats to refine the sampling trajectory of the 2D diffusion model which enhances 3D consistency. 
We carefully designed a diffusion process that synergistically combines the strengths of both 2D and 3D models. Our experiments show that our approach outperforms all previous reconstruction works in both appearance and geometry. We also extensively ablate our method which proves the effectiveness of our proposed ideas. Our code and pretrained models will be released on our \href{https://yuxuan-xue.com/human-3diffusion}{Project Page} to foster future research.

\paragraph{Acknowledgements}
We appreciate G.Tiwari, Y.He, Y. Xiu, Z.Liu, Z.Qiu, S.Li and others for their feedback to improve the work. 
This work is made possible by funding from the Carl Zeiss Foundation. 
This work is also funded by the Deutsche Forschungsgemeinschaft (DFG, German Research Foundation) - 409792180 (EmmyNoether Programme, project: Real Virtual Humans) and the German Federal Ministry of Education and Research (BMBF): Tübingen AI Center, FKZ: 01IS18039A. 
The authors thank the International Max Planck Research School for Intelligent Systems (IMPRS-IS) for supporting Y.Xue.
R. Marin has been supported by innovation program under the Marie Skłodowska-Curie grant agreement No 101109330.
G. Pons-Moll is a member of the Machine Learning Cluster of Excellence, EXC number 2064/1 – Project number 390727645.

\newpage
\bibliographystyle{plainnat}
\bibliography{literatures}

\begin{thebibliography}{110}
\providecommand{\natexlab}[1]{#1}
\providecommand{\url}[1]{\texttt{#1}}
\expandafter\ifx\csname urlstyle\endcsname\relax
  \providecommand{\doi}[1]{doi: #1}\else
  \providecommand{\doi}{doi: \begingroup \urlstyle{rm}\Url}\fi

\bibitem[axy(2023)]{axyz}
Axyz, Nov 2023.
\newblock URL \url{https://secure.axyz-design.com}.

\bibitem[ren(2023)]{renderpeople}
Renderpeople, Nov 2023.
\newblock URL \url{https://renderpeople.com/}.

\bibitem[tre(2023)]{treedy}
Treedy, Nov 2023.
\newblock URL \url{https://treedys.com/}.

\bibitem[twi(2023)]{twindom}
Twindom, Nov 2023.
\newblock URL \url{https://web.twindom.com/}.

\bibitem[Alldieck et~al.(2022)Alldieck, Zanfir, and Sminchisescu]{alldieck_photorealistic_2022}
Thiemo Alldieck, Mihai Zanfir, and Cristian Sminchisescu.
\newblock Photorealistic {Monocular} {3D} {Reconstruction} of {Humans} {Wearing} {Clothing}.
\newblock In \emph{2022 {IEEE}/{CVF} {Conference} on {Computer} {Vision} and {Pattern} {Recognition} ({CVPR})}, pages 1496--1505, New Orleans, LA, USA, June 2022. IEEE.
\newblock ISBN 978-1-66546-946-3.
\newblock \doi{10.1109/CVPR52688.2022.00156}.
\newblock URL \url{https://ieeexplore.ieee.org/document/9878998/}.

\bibitem[Alwala et~al.(2022)Alwala, Gupta, and Tulsiani]{Alwala_CVPR22_pretrain_ss}
Kalyan~Vasudev Alwala, Abhinav Gupta, and Shubham Tulsiani.
\newblock Pre-train, self-train, distill: A simple recipe for supersizing 3d reconstruction.
\newblock In \emph{Proceedings of the IEEE/CVF Conference on Computer Vision and Pattern Recognition (CVPR)}, pages 3773--3782, June 2022.

\bibitem[Anciukevicius et~al.(2023)Anciukevicius, Xu, Fisher, Henderson, Bilen, Mitra, and Guerrero]{anciukevicius2023renderdiffusion}
Titas Anciukevicius, Zexiang Xu, Matthew Fisher, Paul Henderson, Hakan Bilen, Niloy~J. Mitra, and Paul Guerrero.
\newblock Renderdiffusion: Image diffusion for 3d reconstruction, inpainting and generation.
\newblock In \emph{{IEEE/CVF} Conference on Computer Vision and Pattern Recognition, {CVPR} 2023, Vancouver, BC, Canada, June 17-24, 2023}, pages 12608--12618. {IEEE}, 2023.
\newblock \doi{10.1109/CVPR52729.2023.01213}.
\newblock URL \url{https://doi.org/10.1109/CVPR52729.2023.01213}.

\bibitem[Antic et~al.(2024)Antic, Tiwari, Ozcomlekci, Marin, and Pons{-}Moll]{antic2024closed}
Dimitrije Antic, Garvita Tiwari, Batuhan Ozcomlekci, Riccardo Marin, and Gerard Pons{-}Moll.
\newblock Close: {A} 3d clothing segmentation dataset and model.
\newblock In \emph{International Conference on 3D Vision, 3DV 2024, Davos, Switzerland, March 18-21, 2024}, pages 591--601. {IEEE}, 2024.
\newblock \doi{10.1109/3DV62453.2024.00020}.
\newblock URL \url{https://doi.org/10.1109/3DV62453.2024.00020}.

\bibitem[Bhatnagar et~al.(2022)Bhatnagar, Xie, Petrov, Sminchisescu, Theobalt, and Pons-Moll]{bhatnagar22behave}
Bharat~Lal Bhatnagar, Xianghui Xie, Ilya Petrov, Cristian Sminchisescu, Christian Theobalt, and Gerard Pons-Moll.
\newblock Behave: Dataset and method for tracking human object interactions.
\newblock In \emph{{IEEE} Conference on Computer Vision and Pattern Recognition (CVPR)}. {IEEE}, jun 2022.

\bibitem[Black et~al.(2023)Black, Patel, Tesch, and Yang]{Black_bedlam_2023}
Michael~J. Black, Priyanka Patel, Joachim Tesch, and Jinlong Yang.
\newblock {BEDLAM}: A synthetic dataset of bodies exhibiting detailed lifelike animated motion.
\newblock In \emph{Proceedings IEEE/CVF Conf.~on Computer Vision and Pattern Recognition (CVPR)}, pages 8726--8737, June 2023.

\bibitem[Cao et~al.(2022)Cao, Santo, Shi, Okura, and Matsushita]{bini2022cao}
Xu~Cao, Hiroaki Santo, Boxin Shi, Fumio Okura, and Yasuyuki Matsushita.
\newblock Bilateral normal integration.
\newblock 2022.

\bibitem[Chan et~al.(2022)Chan, Lin, Chan, Nagano, Pan, Mello, Gallo, Guibas, Tremblay, Khamis, Karras, and Wetzstein]{chan2022eg3d}
Eric~R. Chan, Connor~Z. Lin, Matthew~A. Chan, Koki Nagano, Boxiao Pan, Shalini~De Mello, Orazio Gallo, Leonidas~J. Guibas, Jonathan Tremblay, Sameh Khamis, Tero Karras, and Gordon Wetzstein.
\newblock Efficient geometry-aware 3d generative adversarial networks.
\newblock In \emph{{IEEE/CVF} Conference on Computer Vision and Pattern Recognition, {CVPR} 2022, New Orleans, LA, USA, June 18-24, 2022}, pages 16102--16112. {IEEE}, 2022.
\newblock \doi{10.1109/CVPR52688.2022.01565}.
\newblock URL \url{https://doi.org/10.1109/CVPR52688.2022.01565}.

\bibitem[Deitke et~al.(2023)Deitke, Schwenk, Salvador, Weihs, Michel, VanderBilt, Schmidt, Ehsani, Kembhavi, and Farhadi]{Deitke2023Objaverse}
Matt Deitke, Dustin Schwenk, Jordi Salvador, Luca Weihs, Oscar Michel, Eli VanderBilt, Ludwig Schmidt, Kiana Ehsani, Aniruddha Kembhavi, and Ali Farhadi.
\newblock Objaverse: {A} universe of annotated 3d objects.
\newblock In \emph{{IEEE/CVF} Conference on Computer Vision and Pattern Recognition, {CVPR} 2023, Vancouver, BC, Canada, June 17-24, 2023}, pages 13142--13153. {IEEE}, 2023.
\newblock \doi{10.1109/CVPR52729.2023.01263}.
\newblock URL \url{https://doi.org/10.1109/CVPR52729.2023.01263}.

\bibitem[Denninger et~al.(2023)Denninger, Winkelbauer, Sundermeyer, Boerdijk, Knauer, Strobl, Humt, and Triebel]{denninger2023blenderproc}
Maximilian Denninger, Dominik Winkelbauer, Martin Sundermeyer, Wout Boerdijk, Markus Knauer, Klaus~H. Strobl, Matthias Humt, and Rudolph Triebel.
\newblock Blenderproc2: A procedural pipeline for photorealistic rendering.
\newblock \emph{Journal of Open Source Software}, 8\penalty0 (82):\penalty0 4901, 2023.
\newblock \doi{10.21105/joss.04901}.
\newblock URL \url{https://doi.org/10.21105/joss.04901}.

\bibitem[Dhariwal and Nichol(2021)]{dhariwal2021diffusionClassGuidance}
Prafulla Dhariwal and Alexander~Quinn Nichol.
\newblock Diffusion models beat {GAN}s on image synthesis.
\newblock In A.~Beygelzimer, Y.~Dauphin, P.~Liang, and J.~Wortman Vaughan, editors, \emph{Advances in Neural Information Processing Systems}, 2021.
\newblock URL \url{https://openreview.net/forum?id=AAWuCvzaVt}.

\bibitem[Dong et~al.(2022)Dong, Guo, Song, Chen, Geiger, and Hilliges]{dong2022pina}
Zijian Dong, Chen Guo, Jie Song, Xu~Chen, Andreas Geiger, and Otmar Hilliges.
\newblock Pina: Learning a personalized implicit neural avatar from a single rgb-d video sequence.
\newblock \emph{arXiv}, 2022.

\bibitem[Downs et~al.(2022)Downs, Francis, Koenig, Kinman, Hickman, Reymann, McHugh, and Vanhoucke]{downs2022gso}
Laura Downs, Anthony Francis, Nate Koenig, Brandon Kinman, Ryan Hickman, Krista Reymann, Thomas~Barlow McHugh, and Vincent Vanhoucke.
\newblock Google scanned objects: {A} high-quality dataset of 3d scanned household items.
\newblock In \emph{2022 International Conference on Robotics and Automation, {ICRA} 2022, Philadelphia, PA, USA, May 23-27, 2022}, pages 2553--2560. {IEEE}, 2022.
\newblock \doi{10.1109/ICRA46639.2022.9811809}.
\newblock URL \url{https://doi.org/10.1109/ICRA46639.2022.9811809}.

\bibitem[Feng et~al.(2022)Feng, Liu, Lai, ingyu Yang, and Li]{feng2022fof}
Qiao Feng, Yebin Liu, Yu-Kun Lai, ingyu Yang, and Kun Li.
\newblock Fof: Learning fourier occupancy field for monocular real-time human reconstruction.
\newblock In \emph{Advances in Neural Information Processing Systems 35: Annual Conference on Neural Information Processing Systems 2022, NeurIPS 2022, New Orleans, LA, USA, November 28 - December 9, 2022}, 2022.

\bibitem[Gao* et~al.(2024)Gao*, Holynski*, Henzler, Brussee, Martin-Brualla, Srinivasan, Barron, and Poole*]{gao2024cat3d}
Ruiqi Gao*, Aleksander Holynski*, Philipp Henzler, Arthur Brussee, Ricardo Martin-Brualla, Pratul~P. Srinivasan, Jonathan~T. Barron, and Ben Poole*.
\newblock Cat3d: Create anything in 3d with multi-view diffusion models.
\newblock \emph{arXiv}, 2024.

\bibitem[Guzov et~al.(2021)Guzov, Mir, Sattler, and Pons{-}Moll]{guzov2021hps}
Vladimir Guzov, Aymen Mir, Torsten Sattler, and Gerard Pons{-}Moll.
\newblock Human poseitioning system {(HPS):} 3d human pose estimation and self-localization in large scenes from body-mounted sensors.
\newblock \emph{CoRR}, abs/2103.17265, 2021.
\newblock URL \url{https://arxiv.org/abs/2103.17265}.

\bibitem[Han et~al.(2023)Han, Park, Yoon, Kang, Park, and Jeon]{han20232k2k}
Sang-Hun Han, Min-Gyu Park, Ju~Hong Yoon, Ju-Mi Kang, Young-Jae Park, and Hae-Gon Jeon.
\newblock High-fidelity 3d human digitization from single 2k resolution images.
\newblock In \emph{IEEE Conference on Computer Vision and Pattern Recognition (CVPR2023)}, June 2023.

\bibitem[Heusel et~al.(2017)Heusel, Ramsauer, Unterthiner, Nessler, and Hochreiter]{heusel2017fid}
Martin Heusel, Hubert Ramsauer, Thomas Unterthiner, Bernhard Nessler, and Sepp Hochreiter.
\newblock Gans trained by a two time-scale update rule converge to a local nash equilibrium.
\newblock In Isabelle Guyon, Ulrike von Luxburg, Samy Bengio, Hanna~M. Wallach, Rob Fergus, S.~V.~N. Vishwanathan, and Roman Garnett, editors, \emph{Advances in Neural Information Processing Systems 30: Annual Conference on Neural Information Processing Systems 2017, December 4-9, 2017, Long Beach, CA, {USA}}, pages 6626--6637, 2017.
\newblock URL \url{https://proceedings.neurips.cc/paper/2017/hash/8a1d694707eb0fefe65871369074926d-Abstract.html}.

\bibitem[Ho et~al.(2023)Ho, Song, and Hilliges]{ho2023sith}
Hsuan{-}I Ho, Jie Song, and Otmar Hilliges.
\newblock Sith: Single-view textured human reconstruction with image-conditioned diffusion.
\newblock \emph{CoRR}, abs/2311.15855, 2023.
\newblock \doi{10.48550/ARXIV.2311.15855}.
\newblock URL \url{https://doi.org/10.48550/arXiv.2311.15855}.

\bibitem[Ho and Salimans(2021)]{ho2021classifierfreeCFG}
Jonathan Ho and Tim Salimans.
\newblock Classifier-free diffusion guidance.
\newblock In \emph{NeurIPS 2021 Workshop on Deep Generative Models and Downstream Applications}, 2021.
\newblock URL \url{https://openreview.net/forum?id=qw8AKxfYbI}.

\bibitem[Ho et~al.(2020)Ho, Jain, and Abbeel]{Ho2020DDPM}
Jonathan Ho, Ajay Jain, and Pieter Abbeel.
\newblock Denoising diffusion probabilistic models.
\newblock In Hugo Larochelle, Marc'Aurelio Ranzato, Raia Hadsell, Maria{-}Florina Balcan, and Hsuan{-}Tien Lin, editors, \emph{Advances in Neural Information Processing Systems 33: Annual Conference on Neural Information Processing Systems 2020, NeurIPS 2020, December 6-12, 2020, virtual}, 2020.
\newblock URL \url{https://proceedings.neurips.cc/paper/2020/hash/4c5bcfec8584af0d967f1ab10179ca4b-Abstract.html}.

\bibitem[Hong et~al.(2023)Hong, Zhang, Gu, Bi, Zhou, Liu, Liu, Sunkavalli, Bui, and Tan]{hong2023lrm}
Yicong Hong, Kai Zhang, Jiuxiang Gu, Sai Bi, Yang Zhou, Difan Liu, Feng Liu, Kalyan Sunkavalli, Trung Bui, and Hao Tan.
\newblock {LRM:} large reconstruction model for single image to 3d.
\newblock \emph{CoRR}, abs/2311.04400, 2023.
\newblock \doi{10.48550/ARXIV.2311.04400}.
\newblock URL \url{https://doi.org/10.48550/arXiv.2311.04400}.

\bibitem[Hsuan-I et~al.(2023)Hsuan-I, Lixin, Jie, and Otmar]{ho2023customhuman}
Ho~Hsuan-I, Xue Lixin, Song Jie, and Hilliges Otmar.
\newblock Learning locally editable virtual humans.
\newblock In \emph{Proceedings of the IEEE Conference on Computer Vision and Pattern Recognition (CVPR)}, 2023.

\bibitem[Hu et~al.(2024)Hu, Zhou, Jampani, and Tulsiani]{huMVDFusionSingleview3D2024}
Hanzhe Hu, Zhizhuo Zhou, Varun Jampani, and Shubham Tulsiani.
\newblock {{MVD-Fusion}}: {{Single-view 3D}} via {{Depth-consistent Multi-view Generation}}, April 2024.

\bibitem[Huang et~al.(2024{\natexlab{a}})Huang, Yu, Chen, Geiger, and Gao]{Huang20242dgs}
Binbin Huang, Zehao Yu, Anpei Chen, Andreas Geiger, and Shenghua Gao.
\newblock 2d gaussian splatting for geometrically accurate radiance fields.
\newblock \emph{CoRR}, abs/2403.17888, 2024{\natexlab{a}}.
\newblock \doi{10.48550/ARXIV.2403.17888}.
\newblock URL \url{https://doi.org/10.48550/arXiv.2403.17888}.

\bibitem[Huang et~al.(2024{\natexlab{b}})Huang, Wen, Dong, Wang, Li, Chen, Cao, Liang, Qiao, Dai, and Sheng]{huangEpiDiffEnhancingMultiView2024}
Zehuan Huang, Hao Wen, Junting Dong, Yaohui Wang, Yangguang Li, Xinyuan Chen, Yan-Pei Cao, Ding Liang, Yu~Qiao, Bo~Dai, and Lu~Sheng.
\newblock {{EpiDiff}}: {{Enhancing Multi-View Synthesis}} via {{Localized Epipolar-Constrained Diffusion}}, April 2024{\natexlab{b}}.

\bibitem[Jiang et~al.(2022)Jiang, Chen, Song, and Hilliges]{jiang2022instantavatar}
Tianjian Jiang, Xu~Chen, Jie Song, and Otmar Hilliges.
\newblock Instantavatar: Learning avatars from monocular video in 60 seconds.
\newblock \emph{arXiv}, 2022.

\bibitem[Jinka et~al.(2023)Jinka, Srivastava, Pokhariya, Sharma, and Narayanan]{jinka2023iiit}
Sai~Sagar Jinka, Astitva Srivastava, Chandradeep Pokhariya, Avinash Sharma, and P.~J. Narayanan.
\newblock {SHARP:} shape-aware reconstruction of people in loose clothing.
\newblock \emph{Int. J. Comput. Vis.}, 131\penalty0 (4):\penalty0 918--937, 2023.
\newblock \doi{10.1007/S11263-022-01736-Z}.
\newblock URL \url{https://doi.org/10.1007/s11263-022-01736-z}.

\bibitem[Kabadayi et~al.(2024)Kabadayi, Zielonka, Bhatnagar, Pons-Moll, and Thies]{kabadayi24ganavatar}
Berna Kabadayi, Wojciech Zielonka, Bharat~Lal Bhatnagar, Gerard Pons-Moll, and Justus Thies.
\newblock Gan-avatar: Controllable personalized gan-based human head avatar.
\newblock In \emph{International Conference on 3D Vision (3DV)}, March 2024.

\bibitem[Kant et~al.(2024)Kant, Wu, Vasilkovsky, Qian, Ren, G{\"{u}}ler, Ghanem, Tulyakov, Gilitschenski, and Siarohin]{kant2024spad}
Yash Kant, Ziyi Wu, Michael Vasilkovsky, Guocheng Qian, Jian Ren, Riza~Alp G{\"{u}}ler, Bernard Ghanem, Sergey Tulyakov, Igor Gilitschenski, and Aliaksandr Siarohin.
\newblock {SPAD} : Spatially aware multiview diffusers.
\newblock \emph{CoRR}, abs/2402.05235, 2024.
\newblock \doi{10.48550/ARXIV.2402.05235}.
\newblock URL \url{https://doi.org/10.48550/arXiv.2402.05235}.

\bibitem[Karnewar et~al.(2023)Karnewar, Vedaldi, Novotny, and Mitra]{karnewar2023holodiffusion}
Animesh Karnewar, Andrea Vedaldi, David Novotny, and Niloy Mitra.
\newblock Holodiffusion: Training a {3D} diffusion model using {2D} images.
\newblock In \emph{Proceedings of the IEEE/CVF conference on computer vision and pattern recognition}, 2023.

\bibitem[Kerbl et~al.(2023)Kerbl, Kopanas, Leimk{\"{u}}hler, and Drettakis]{Kerbl20233dgs}
Bernhard Kerbl, Georgios Kopanas, Thomas Leimk{\"{u}}hler, and George Drettakis.
\newblock 3d gaussian splatting for real-time radiance field rendering.
\newblock \emph{{ACM} Trans. Graph.}, 42\penalty0 (4):\penalty0 139:1--139:14, 2023.
\newblock \doi{10.1145/3592433}.
\newblock URL \url{https://doi.org/10.1145/3592433}.

\bibitem[Kong et~al.(2024)Kong, Liu, Lyu, Taher, Qi, and Davison]{kong2024eschernet}
Xin Kong, Shikun Liu, Xiaoyang Lyu, Marwan Taher, Xiaojuan Qi, and Andrew~J. Davison.
\newblock Eschernet: {A} generative model for scalable view synthesis.
\newblock \emph{CoRR}, abs/2402.03908, 2024.
\newblock \doi{10.48550/ARXIV.2402.03908}.
\newblock URL \url{https://doi.org/10.48550/arXiv.2402.03908}.

\bibitem[Li et~al.(2024)Li, Chen, Larionov, Sarafianos, Matusik, and Stuyck]{li2023diffavatar}
Yifei Li, Hsiao-yu Chen, Egor Larionov, Nikolaos Sarafianos, Wojciech Matusik, and Tuur Stuyck.
\newblock {DiffAvatar}: Simulation-ready garment optimization with differentiable simulation.
\newblock In \emph{Proceedings of the IEEE/CVF Conference on Computer Vision and Pattern Recognition (CVPR)}, June 2024.

\bibitem[Liao et~al.(2023)Liao, Yi, Xiu, Tang, Huang, Thies, and Black]{liao2023tada}
Tingting Liao, Hongwei Yi, Yuliang Xiu, Jiaxiang Tang, Yangyi Huang, Justus Thies, and Michael~J. Black.
\newblock Tada! text to animatable digital avatars.
\newblock \emph{CoRR}, abs/2308.10899, 2023.
\newblock \doi{10.48550/ARXIV.2308.10899}.
\newblock URL \url{https://doi.org/10.48550/arXiv.2308.10899}.

\bibitem[Liu et~al.(2023{\natexlab{a}})Liu, Shi, Chen, Zhang, Xu, Wei, Chen, Zeng, Gu, and Su]{liu2023one2345++}
Minghua Liu, Ruoxi Shi, Linghao Chen, Zhuoyang Zhang, Chao Xu, Xinyue Wei, Hansheng Chen, Chong Zeng, Jiayuan Gu, and Hao Su.
\newblock One-2-3-45++: Fast single image to 3d objects with consistent multi-view generation and 3d diffusion.
\newblock \emph{CoRR}, abs/2311.07885, 2023{\natexlab{a}}.
\newblock \doi{10.48550/ARXIV.2311.07885}.
\newblock URL \url{https://doi.org/10.48550/arXiv.2311.07885}.

\bibitem[Liu et~al.(2023{\natexlab{b}})Liu, Xu, Jin, Chen, T, Xu, and Su]{liu2023one2345}
Minghua Liu, Chao Xu, Haian Jin, Linghao Chen, Mukund~Varma T, Zexiang Xu, and Hao Su.
\newblock One-2-3-45: Any single image to 3d mesh in 45 seconds without per-shape optimization.
\newblock In Alice Oh, Tristan Naumann, Amir Globerson, Kate Saenko, Moritz Hardt, and Sergey Levine, editors, \emph{Advances in Neural Information Processing Systems 36: Annual Conference on Neural Information Processing Systems 2023, NeurIPS 2023, New Orleans, LA, USA, December 10 - 16, 2023}, 2023{\natexlab{b}}.
\newblock URL \url{http://papers.nips.cc/paper\_files/paper/2023/hash/4683beb6bab325650db13afd05d1a14a-Abstract-Conference.html}.

\bibitem[Liu et~al.(2023{\natexlab{c}})Liu, Wu, Hoorick, Tokmakov, Zakharov, and Vondrick]{Liu2023zero123}
Ruoshi Liu, Rundi Wu, Basile~Van Hoorick, Pavel Tokmakov, Sergey Zakharov, and Carl Vondrick.
\newblock Zero-1-to-3: Zero-shot one image to 3d object.
\newblock In \emph{{IEEE/CVF} International Conference on Computer Vision, {ICCV} 2023, Paris, France, October 1-6, 2023}, pages 9264--9275. {IEEE}, 2023{\natexlab{c}}.
\newblock \doi{10.1109/ICCV51070.2023.00853}.
\newblock URL \url{https://doi.org/10.1109/ICCV51070.2023.00853}.

\bibitem[Liu et~al.(2023{\natexlab{d}})Liu, Lin, Zeng, Long, Liu, Komura, and Wang]{liu2023syncdreamer}
Yuan Liu, Cheng Lin, Zijiao Zeng, Xiaoxiao Long, Lingjie Liu, Taku Komura, and Wenping Wang.
\newblock Syncdreamer: Generating multiview-consistent images from a single-view image.
\newblock \emph{CoRR}, abs/2309.03453, 2023{\natexlab{d}}.
\newblock \doi{10.48550/ARXIV.2309.03453}.
\newblock URL \url{https://doi.org/10.48550/arXiv.2309.03453}.

\bibitem[Liu et~al.(2023{\natexlab{e}})Liu, Feng, Xiu, Liu, Paull, Black, and Schölkopf]{liu2023gshell}
Zhen Liu, Yao Feng, Yuliang Xiu, Weiyang Liu, Liam Paull, Michael~J. Black, and Bernhard Schölkopf.
\newblock Ghost on the shell: An expressive representation of general 3d shapes.
\newblock \emph{arXiv preprint arXiv:2310.15168}, 2023{\natexlab{e}}.

\bibitem[Long et~al.(2023)Long, Guo, Lin, Liu, Dou, Liu, Ma, Zhang, Habermann, Theobalt, and Wang]{long2023wonder3d}
Xiaoxiao Long, Yuan{-}Chen Guo, Cheng Lin, Yuan Liu, Zhiyang Dou, Lingjie Liu, Yuexin Ma, Song{-}Hai Zhang, Marc Habermann, Christian Theobalt, and Wenping Wang.
\newblock Wonder3d: Single image to 3d using cross-domain diffusion.
\newblock \emph{CoRR}, abs/2310.15008, 2023.
\newblock \doi{10.48550/ARXIV.2310.15008}.
\newblock URL \url{https://doi.org/10.48550/arXiv.2310.15008}.

\bibitem[Loper et~al.(2015)Loper, Mahmood, Romero, Pons{-}Moll, and Black]{loper2015smpl}
Matthew Loper, Naureen Mahmood, Javier Romero, Gerard Pons{-}Moll, and Michael~J. Black.
\newblock {SMPL:} a skinned multi-person linear model.
\newblock \emph{{ACM} Trans. Graph.}, 34\penalty0 (6):\penalty0 248:1--248:16, 2015.
\newblock \doi{10.1145/2816795.2818013}.
\newblock URL \url{https://doi.org/10.1145/2816795.2818013}.

\bibitem[Ma et~al.(2019)Ma, Tang, Pujades, Pons{-}Moll, Ranjan, and Black]{ma2019cape}
Qianli Ma, Siyu Tang, Sergi Pujades, Gerard Pons{-}Moll, Anurag Ranjan, and Michael~J. Black.
\newblock Dressing 3d humans using a conditional mesh-vae-gan.
\newblock \emph{CoRR}, abs/1907.13615, 2019.
\newblock URL \url{http://arxiv.org/abs/1907.13615}.

\bibitem[Melas-Kyriazi et~al.(2023)Melas-Kyriazi, Rupprecht, Laina, and Vedaldi]{melaskyriazi2023realfusion}
Luke Melas-Kyriazi, Christian Rupprecht, Iro Laina, and Andrea Vedaldi.
\newblock Realfusion: 360° reconstruction of any object from a single image.
\newblock In \emph{Arxiv}, 2023.

\bibitem[Mildenhall et~al.(2022)Mildenhall, Srinivasan, Tancik, Barron, Ramamoorthi, and Ng]{Mildenhall2020NeRF}
Ben Mildenhall, Pratul~P. Srinivasan, Matthew Tancik, Jonathan~T. Barron, Ravi Ramamoorthi, and Ren Ng.
\newblock Nerf: representing scenes as neural radiance fields for view synthesis.
\newblock \emph{Commun. {ACM}}, 65\penalty0 (1):\penalty0 99--106, 2022.
\newblock \doi{10.1145/3503250}.
\newblock URL \url{https://doi.org/10.1145/3503250}.

\bibitem[Pavlakos et~al.(2019)Pavlakos, Choutas, Ghorbani, Bolkart, Osman, Tzionas, and Black]{pavlakos2019smplx}
Georgios Pavlakos, Vasileios Choutas, Nima Ghorbani, Timo Bolkart, Ahmed A.~A. Osman, Dimitrios Tzionas, and Michael~J. Black.
\newblock Expressive body capture: 3d hands, face, and body from a single image.
\newblock \emph{CoRR}, abs/1904.05866, 2019.
\newblock URL \url{http://arxiv.org/abs/1904.05866}.

\bibitem[Petrov et~al.(2023)Petrov, Marin, Chibane, and Pons{-}Moll]{petriv2023objectpopup}
Ilya~A. Petrov, Riccardo Marin, Julian Chibane, and Gerard Pons{-}Moll.
\newblock Object pop-up: Can we infer 3d objects and their poses from human interactions alone?
\newblock In \emph{{IEEE/CVF} Conference on Computer Vision and Pattern Recognition, {CVPR} 2023, Vancouver, BC, Canada, June 17-24, 2023}, pages 4726--4736. {IEEE}, 2023.
\newblock \doi{10.1109/CVPR52729.2023.00458}.
\newblock URL \url{https://doi.org/10.1109/CVPR52729.2023.00458}.

\bibitem[Pons{-}Moll et~al.(2017)Pons{-}Moll, Pujades, Hu, and Black]{ponsmoll2017clothcap}
Gerard Pons{-}Moll, Sergi Pujades, Sonny Hu, and Michael~J. Black.
\newblock Clothcap: seamless 4d clothing capture and retargeting.
\newblock \emph{{ACM} Trans. Graph.}, 36\penalty0 (4):\penalty0 73:1--73:15, 2017.
\newblock \doi{10.1145/3072959.3073711}.
\newblock URL \url{https://doi.org/10.1145/3072959.3073711}.

\bibitem[Poole et~al.(2022)Poole, Jain, Barron, and Mildenhall]{poole2022dreamfusion}
Ben Poole, Ajay Jain, Jonathan~T. Barron, and Ben Mildenhall.
\newblock Dreamfusion: Text-to-3d using 2d diffusion.
\newblock \emph{arXiv}, 2022.

\bibitem[Qian et~al.(2023)Qian, Mai, Hamdi, Ren, Siarohin, Li, Lee, Skorokhodov, Wonka, Tulyakov, and Ghanem]{qian2023magic123}
Guocheng Qian, Jinjie Mai, Abdullah Hamdi, Jian Ren, Aliaksandr Siarohin, Bing Li, Hsin{-}Ying Lee, Ivan Skorokhodov, Peter Wonka, Sergey Tulyakov, and Bernard Ghanem.
\newblock Magic123: One image to high-quality 3d object generation using both 2d and 3d diffusion priors.
\newblock \emph{CoRR}, abs/2306.17843, 2023.
\newblock \doi{10.48550/ARXIV.2306.17843}.
\newblock URL \url{https://doi.org/10.48550/arXiv.2306.17843}.

\bibitem[Rombach et~al.(2022)Rombach, Blattmann, Lorenz, Esser, and Ommer]{Rombach2022StableDiffusion}
Robin Rombach, Andreas Blattmann, Dominik Lorenz, Patrick Esser, and Bj{\"{o}}rn Ommer.
\newblock High-resolution image synthesis with latent diffusion models.
\newblock In \emph{{IEEE/CVF} Conference on Computer Vision and Pattern Recognition, {CVPR} 2022, New Orleans, LA, USA, June 18-24, 2022}, pages 10674--10685. {IEEE}, 2022.
\newblock \doi{10.1109/CVPR52688.2022.01042}.
\newblock URL \url{https://doi.org/10.1109/CVPR52688.2022.01042}.

\bibitem[Saito et~al.(2019)Saito, Huang, Natsume, Morishima, Li, and Kanazawa]{saito2019pifu}
Shunsuke Saito, Zeng Huang, Ryota Natsume, Shigeo Morishima, Hao Li, and Angjoo Kanazawa.
\newblock Pifu: Pixel-aligned implicit function for high-resolution clothed human digitization.
\newblock In \emph{2019 {IEEE/CVF} International Conference on Computer Vision, {ICCV} 2019, Seoul, Korea (South), October 27 - November 2, 2019}, pages 2304--2314. {IEEE}, 2019.
\newblock \doi{10.1109/ICCV.2019.00239}.
\newblock URL \url{https://doi.org/10.1109/ICCV.2019.00239}.

\bibitem[Saito et~al.(2020)Saito, Simon, Saragih, and Joo]{saito2020pifuhd}
Shunsuke Saito, Tomas Simon, Jason~M. Saragih, and Hanbyul Joo.
\newblock Pifuhd: Multi-level pixel-aligned implicit function for high-resolution 3d human digitization.
\newblock In \emph{2020 {IEEE/CVF} Conference on Computer Vision and Pattern Recognition, {CVPR} 2020, Seattle, WA, USA, June 13-19, 2020}, pages 81--90. Computer Vision Foundation / {IEEE}, 2020.
\newblock \doi{10.1109/CVPR42600.2020.00016}.
\newblock URL \url{https://openaccess.thecvf.com/content\_CVPR\_2020/html/Saito\_PIFuHD\_Multi-Level\_Pixel-Aligned\_Implicit\_Function\_for\_High-Resolution\_3D\_Human\_Digitization\_CVPR\_2020\_paper.html}.

\bibitem[Schuhmann et~al.(2022)Schuhmann, Beaumont, Vencu, Gordon, Wightman, Cherti, Coombes, Katta, Mullis, Wortsman, Schramowski, Kundurthy, Crowson, Schmidt, Kaczmarczyk, and Jitsev]{Schumann2022Laion5B}
Christoph Schuhmann, Romain Beaumont, Richard Vencu, Cade Gordon, Ross Wightman, Mehdi Cherti, Theo Coombes, Aarush Katta, Clayton Mullis, Mitchell Wortsman, Patrick Schramowski, Srivatsa Kundurthy, Katherine Crowson, Ludwig Schmidt, Robert Kaczmarczyk, and Jenia Jitsev.
\newblock {LAION-5B:} an open large-scale dataset for training next generation image-text models.
\newblock In Sanmi Koyejo, S.~Mohamed, A.~Agarwal, Danielle Belgrave, K.~Cho, and A.~Oh, editors, \emph{Advances in Neural Information Processing Systems 35: Annual Conference on Neural Information Processing Systems 2022, NeurIPS 2022, New Orleans, LA, USA, November 28 - December 9, 2022}, 2022.
\newblock URL \url{http://papers.nips.cc/paper\_files/paper/2022/hash/a1859debfb3b59d094f3504d5ebb6c25-Abstract-Datasets\_and\_Benchmarks.html}.

\bibitem[Sengupta et~al.(2024)Sengupta, Alldieck, Kolotouros, Corona, Zanfir, and Sminchisescu]{sengupta_diffhuman_2024}
Akash Sengupta, Thiemo Alldieck, Nikos Kolotouros, Enric Corona, Andrei Zanfir, and Cristian Sminchisescu.
\newblock {DiffHuman}: {Probabilistic} {Photorealistic} {3D} {Reconstruction} of {Humans}, March 2024.
\newblock URL \url{http://arxiv.org/abs/2404.00485}.
\newblock arXiv:2404.00485 [cs].

\bibitem[Shi et~al.(2023{\natexlab{a}})Shi, Chen, Zhang, Liu, Xu, Wei, Chen, Zeng, and Su]{shi2023zero123++}
Ruoxi Shi, Hansheng Chen, Zhuoyang Zhang, Minghua Liu, Chao Xu, Xinyue Wei, Linghao Chen, Chong Zeng, and Hao Su.
\newblock Zero123++: a single image to consistent multi-view diffusion base model.
\newblock \emph{CoRR}, abs/2310.15110, 2023{\natexlab{a}}.
\newblock \doi{10.48550/ARXIV.2310.15110}.
\newblock URL \url{https://doi.org/10.48550/arXiv.2310.15110}.

\bibitem[Shi et~al.(2023{\natexlab{b}})Shi, Wang, Ye, Long, Li, and Yang]{shi2023mvdream}
Yichun Shi, Peng Wang, Jianglong Ye, Mai Long, Kejie Li, and Xiao Yang.
\newblock Mvdream: Multi-view diffusion for 3d generation.
\newblock \emph{CoRR}, abs/2308.16512, 2023{\natexlab{b}}.
\newblock \doi{10.48550/ARXIV.2308.16512}.
\newblock URL \url{https://doi.org/10.48550/arXiv.2308.16512}.

\bibitem[Simonyan and Zisserman(2015)]{Simonyan2015vgg}
Karen Simonyan and Andrew Zisserman.
\newblock Very deep convolutional networks for large-scale image recognition.
\newblock In Yoshua Bengio and Yann LeCun, editors, \emph{3rd International Conference on Learning Representations, {ICLR} 2015, San Diego, CA, USA, May 7-9, 2015, Conference Track Proceedings}, 2015.
\newblock URL \url{http://arxiv.org/abs/1409.1556}.

\bibitem[Song et~al.(2021{\natexlab{a}})Song, Meng, and Ermon]{song2021ddim}
Jiaming Song, Chenlin Meng, and Stefano Ermon.
\newblock Denoising diffusion implicit models.
\newblock In \emph{9th International Conference on Learning Representations, {ICLR} 2021, Virtual Event, Austria, May 3-7, 2021}. OpenReview.net, 2021{\natexlab{a}}.
\newblock URL \url{https://openreview.net/forum?id=St1giarCHLP}.

\bibitem[Song et~al.(2021{\natexlab{b}})Song, Sohl-Dickstein, Kingma, Kumar, Ermon, and Poole]{song_score-based_2021}
Yang Song, Jascha Sohl-Dickstein, Diederik~P. Kingma, Abhishek Kumar, Stefano Ermon, and Ben Poole.
\newblock Score-{Based} {Generative} {Modeling} through {Stochastic} {Differential} {Equations}, February 2021{\natexlab{b}}.
\newblock URL \url{http://arxiv.org/abs/2011.13456}.
\newblock arXiv:2011.13456 [cs, stat].

\bibitem[Su et~al.(2023)Su, Yu, Wang, and Liu]{su2022thuman3}
Zhaoqi Su, Tao Yu, Yangang Wang, and Yebin Liu.
\newblock Deepcloth: Neural garment representation for shape and style editing.
\newblock \emph{IEEE Transactions on Pattern Analysis and Machine Intelligence}, 45\penalty0 (2):\penalty0 1581--1593, 2023.
\newblock \doi{10.1109/TPAMI.2022.3168569}.

\bibitem[Suhail et~al.(2022)Suhail, Esteves, Sigal, and Makadia]{suhail2022epipolarattention}
Mohammed Suhail, Carlos Esteves, Leonid Sigal, and Ameesh Makadia.
\newblock Generalizable patch-based neural rendering.
\newblock In Shai Avidan, Gabriel~J. Brostow, Moustapha Ciss{\'{e}}, Giovanni~Maria Farinella, and Tal Hassner, editors, \emph{Computer Vision - {ECCV} 2022 - 17th European Conference, Tel Aviv, Israel, October 23-27, 2022, Proceedings, Part {XXXII}}, volume 13692 of \emph{Lecture Notes in Computer Science}, pages 156--174. Springer, 2022.
\newblock \doi{10.1007/978-3-031-19824-3\_10}.
\newblock URL \url{https://doi.org/10.1007/978-3-031-19824-3\_10}.

\bibitem[Tang et~al.(2024{\natexlab{a}})Tang, Chen, Chen, Wang, Zeng, and Liu]{Tang2024LGM}
Jiaxiang Tang, Zhaoxi Chen, Xiaokang Chen, Tengfei Wang, Gang Zeng, and Ziwei Liu.
\newblock {LGM:} large multi-view gaussian model for high-resolution 3d content creation.
\newblock \emph{CoRR}, abs/2402.05054, 2024{\natexlab{a}}.
\newblock \doi{10.48550/ARXIV.2402.05054}.
\newblock URL \url{https://doi.org/10.48550/arXiv.2402.05054}.

\bibitem[Tang et~al.(2024{\natexlab{b}})Tang, Chen, Wang, Tang, Zhang, Fan, Chandra, Furukawa, and Ranjan]{tang2023mvdiffusion++}
Shitao Tang, Jiacheng Chen, Dilin Wang, Chengzhou Tang, Fuyang Zhang, Yuchen Fan, Vikas Chandra, Yasutaka Furukawa, and Rakesh Ranjan.
\newblock Mvdiffusion++: {A} dense high-resolution multi-view diffusion model for single or sparse-view 3d object reconstruction.
\newblock \emph{CoRR}, abs/2402.12712, 2024{\natexlab{b}}.
\newblock \doi{10.48550/ARXIV.2402.12712}.
\newblock URL \url{https://doi.org/10.48550/arXiv.2402.12712}.

\bibitem[Tatarchenko et~al.(2019)Tatarchenko, Richter, Ranftl, Li, Koltun, and Brox]{Tatarchenko2019fscore}
Maxim Tatarchenko, Stephan~R. Richter, Ren{\'{e}} Ranftl, Zhuwen Li, Vladlen Koltun, and Thomas Brox.
\newblock What do single-view 3d reconstruction networks learn?
\newblock In \emph{{IEEE} Conference on Computer Vision and Pattern Recognition, {CVPR} 2019, Long Beach, CA, USA, June 16-20, 2019}, pages 3405--3414. Computer Vision Foundation / {IEEE}, 2019.
\newblock \doi{10.1109/CVPR.2019.00352}.
\newblock URL \url{http://openaccess.thecvf.com/content\_CVPR\_2019/html/Tatarchenko\_What\_Do\_Single-View\_3D\_Reconstruction\_Networks\_Learn\_CVPR\_2019\_paper.html}.

\bibitem[Tewari et~al.(2023)Tewari, Yin, Cazenavette, Rezchikov, Tenenbaum, Durand, Freeman, and Sitzmann]{Tewari2023DiffusionWithForward}
Ayush Tewari, Tianwei Yin, George Cazenavette, Semon Rezchikov, Josh Tenenbaum, Fr{\'{e}}do Durand, Bill Freeman, and Vincent Sitzmann.
\newblock Diffusion with forward models: Solving stochastic inverse problems without direct supervision.
\newblock In Alice Oh, Tristan Naumann, Amir Globerson, Kate Saenko, Moritz Hardt, and Sergey Levine, editors, \emph{Advances in Neural Information Processing Systems 36: Annual Conference on Neural Information Processing Systems 2023, NeurIPS 2023, New Orleans, LA, USA, December 10 - 16, 2023}, 2023.
\newblock URL \url{http://papers.nips.cc/paper\_files/paper/2023/hash/28e4ee96c94e31b2d040b4521d2b299e-Abstract-Conference.html}.

\bibitem[Thai et~al.(2020)Thai, Stojanov, Upadhya, and Rehg]{thai3dv2020SDFNet}
Anh Thai, Stefan Stojanov, Vijay Upadhya, and James~M. Rehg.
\newblock 3d reconstruction of novel object shapes from single images, 2020.

\bibitem[Tiwari et~al.(2020)Tiwari, Bhatnagar, Tung, and Pons{-}Moll]{tiwari2020sizer}
Garvita Tiwari, Bharat~Lal Bhatnagar, Tony Tung, and Gerard Pons{-}Moll.
\newblock {SIZER:} {A} dataset and model for parsing 3d clothing and learning size sensitive 3d clothing.
\newblock In Andrea Vedaldi, Horst Bischof, Thomas Brox, and Jan{-}Michael Frahm, editors, \emph{Computer Vision - {ECCV} 2020 - 16th European Conference, Glasgow, UK, August 23-28, 2020, Proceedings, Part {III}}, volume 12348 of \emph{Lecture Notes in Computer Science}, pages 1--18. Springer, 2020.
\newblock \doi{10.1007/978-3-030-58580-8\_1}.
\newblock URL \url{https://doi.org/10.1007/978-3-030-58580-8\_1}.

\bibitem[Tiwari et~al.(2021)Tiwari, Sarafianos, Tung, and Pons{-}Moll]{tiwari2021neuralgif}
Garvita Tiwari, Nikolaos Sarafianos, Tony Tung, and Gerard Pons{-}Moll.
\newblock Neural-gif: Neural generalized implicit functions for animating people in clothing.
\newblock In \emph{2021 {IEEE/CVF} International Conference on Computer Vision, {ICCV} 2021, Montreal, QC, Canada, October 10-17, 2021}, pages 11688--11698. {IEEE}, 2021.
\newblock \doi{10.1109/ICCV48922.2021.01150}.
\newblock URL \url{https://doi.org/10.1109/ICCV48922.2021.01150}.

\bibitem[Tochilkin et~al.(2024)Tochilkin, Pankratz, Liu, Huang, Letts, Li, Liang, Laforte, Jampani, and Cao]{tochilkin2024triposr}
Dmitry Tochilkin, David Pankratz, ZeXiang Liu, Zixuan Huang, Adam Letts, Yangguang Li, Ding Liang, Christian Laforte, Varun Jampani, and Yan{-}Pei Cao.
\newblock Triposr: Fast 3d object reconstruction from a single image.
\newblock \emph{CoRR}, abs/2403.02151, 2024.
\newblock \doi{10.48550/ARXIV.2403.02151}.
\newblock URL \url{https://doi.org/10.48550/arXiv.2403.02151}.

\bibitem[Vaswani et~al.(2017)Vaswani, Shazeer, Parmar, Uszkoreit, Jones, Gomez, Kaiser, and Polosukhin]{vaswani2017attention}
Ashish Vaswani, Noam Shazeer, Niki Parmar, Jakob Uszkoreit, Llion Jones, Aidan~N. Gomez, Lukasz Kaiser, and Illia Polosukhin.
\newblock Attention is all you need.
\newblock In Isabelle Guyon, Ulrike von Luxburg, Samy Bengio, Hanna~M. Wallach, Rob Fergus, S.~V.~N. Vishwanathan, and Roman Garnett, editors, \emph{Advances in Neural Information Processing Systems 30: Annual Conference on Neural Information Processing Systems 2017, December 4-9, 2017, Long Beach, CA, {USA}}, pages 5998--6008, 2017.
\newblock URL \url{https://proceedings.neurips.cc/paper/2017/hash/3f5ee243547dee91fbd053c1c4a845aa-Abstract.html}.

\bibitem[Voleti et~al.(2024)Voleti, Yao, Boss, Letts, Pankratz, Tochilkin, Laforte, Rombach, and Jampani]{voletiSV3DNovelMultiview2024}
Vikram Voleti, Chun-Han Yao, Mark Boss, Adam Letts, David Pankratz, Dmitry Tochilkin, Christian Laforte, Robin Rombach, and Varun Jampani.
\newblock {{SV3D}}: {{Novel Multi-view Synthesis}} and {{3D Generation}} from a {{Single Image}} using {{Latent Video Diffusion}}, March 2024.

\bibitem[Wang and Shi(2023)]{Wang2023ImageDream}
Peng Wang and Yichun Shi.
\newblock Imagedream: Image-prompt multi-view diffusion for 3d generation.
\newblock \emph{CoRR}, abs/2312.02201, 2023.
\newblock \doi{10.48550/ARXIV.2312.02201}.
\newblock URL \url{https://doi.org/10.48550/arXiv.2312.02201}.

\bibitem[Wang et~al.(2003)Wang, Simoncelli, and Bovik]{wang2003ssim}
Z.~Wang, E.P. Simoncelli, and A.C. Bovik.
\newblock Multiscale structural similarity for image quality assessment.
\newblock In \emph{The Thrity-Seventh Asilomar Conference on Signals, Systems \& Computers, 2003}, volume~2, pages 1398--1402 Vol.2, 2003.
\newblock \doi{10.1109/ACSSC.2003.1292216}.

\bibitem[Wang et~al.(2023)Wang, Lu, Wang, Bao, Li, Su, and Zhu]{wang2023prolificdreamer}
Zhengyi Wang, Cheng Lu, Yikai Wang, Fan Bao, Chongxuan Li, Hang Su, and Jun Zhu.
\newblock Prolificdreamer: High-fidelity and diverse text-to-3d generation with variational score distillation.
\newblock \emph{arXiv preprint arXiv:2305.16213}, 2023.

\bibitem[Weng et~al.(2022)Weng, Curless, Srinivasan, Barron, and Kemelmacher-Shlizerman]{weng_humannerf_2022_cvpr}
Chung-Yi Weng, Brian Curless, Pratul~P. Srinivasan, Jonathan~T. Barron, and Ira Kemelmacher-Shlizerman.
\newblock Human{N}e{RF}: Free-viewpoint rendering of moving people from monocular video.
\newblock In \emph{Proceedings of the IEEE/CVF Conference on Computer Vision and Pattern Recognition (CVPR)}, pages 16210--16220, June 2022.

\bibitem[Wu et~al.(2018)Wu, Zhang, Zhang, Zhang, Freeman, and Tenenbaum]{shapehd}
Jiajun Wu, Chengkai Zhang, Xiuming Zhang, Zhoutong Zhang, William~T Freeman, and Joshua~B Tenenbaum.
\newblock {Learning 3D Shape Priors for Shape Completion and Reconstruction}.
\newblock In \emph{European Conference on Computer Vision (ECCV)}, 2018.

\bibitem[Wu et~al.(2023)Wu, Zhang, Fu, Wang, Ren, Pan, Wu, Yang, Wang, Qian, Lin, and Liu]{wu2023omniobject3d}
Tong Wu, Jiarui Zhang, Xiao Fu, Yuxin Wang, Jiawei Ren, Liang Pan, Wayne Wu, Lei Yang, Jiaqi Wang, Chen Qian, Dahua Lin, and Ziwei Liu.
\newblock Omniobject3d: Large-vocabulary 3d object dataset for realistic perception, reconstruction and generation.
\newblock In \emph{{IEEE/CVF} Conference on Computer Vision and Pattern Recognition, {CVPR} 2023, Vancouver, BC, Canada, June 17-24, 2023}, pages 803--814. {IEEE}, 2023.
\newblock \doi{10.1109/CVPR52729.2023.00084}.
\newblock URL \url{https://doi.org/10.1109/CVPR52729.2023.00084}.

\bibitem[Xian et~al.(2022)Xian, Chibane, Bhatnagar, Schiele, Akata, and Pons-Moll]{Xian2022gin}
Yongqin Xian, Julian Chibane, Bharat~Lal Bhatnagar, Bernt Schiele, Zeynep Akata, and Gerard Pons-Moll.
\newblock Any-shot gin: Generalizing implicit networks for reconstructing novel classes.
\newblock In \emph{2022 International Conference on 3D Vision (3DV)}. IEEE, 2022.

\bibitem[Xie et~al.(2022)Xie, Bhatnagar, and Pons{-}Moll]{xie2022chore}
Xianghui Xie, Bharat~Lal Bhatnagar, and Gerard Pons{-}Moll.
\newblock {CHORE:} contact, human and object reconstruction from a single {RGB} image.
\newblock \emph{CoRR}, abs/2204.02445, 2022.
\newblock \doi{10.48550/ARXIV.2204.02445}.
\newblock URL \url{https://doi.org/10.48550/arXiv.2204.02445}.

\bibitem[Xie et~al.(2023)Xie, Bhatnagar, and Pons{-}Moll]{xie2023vistrack}
Xianghui Xie, Bharat~Lal Bhatnagar, and Gerard Pons{-}Moll.
\newblock Visibility aware human-object interaction tracking from single {RGB} camera.
\newblock In \emph{{IEEE/CVF} Conference on Computer Vision and Pattern Recognition, {CVPR} 2023, Vancouver, BC, Canada, June 17-24, 2023}, pages 4757--4768. {IEEE}, 2023.
\newblock \doi{10.1109/CVPR52729.2023.00461}.
\newblock URL \url{https://doi.org/10.1109/CVPR52729.2023.00461}.

\bibitem[Xie et~al.(2024)Xie, Bhatnagar, Lenssen, and Pons-Moll]{xie2023template_free}
Xianghui Xie, Bharat~Lal Bhatnagar, Jan~Eric Lenssen, and Gerard Pons-Moll.
\newblock Template free reconstruction of human-object interaction with procedural interaction generation.
\newblock In \emph{IEEE Conference on Computer Vision and Pattern Recognition (CVPR)}, June 2024.

\bibitem[Xiu et~al.(2022)Xiu, Yang, Tzionas, and Black]{xiu2023icon}
Yuliang Xiu, Jinlong Yang, Dimitrios Tzionas, and Michael~J. Black.
\newblock {ICON:} implicit clothed humans obtained from normals.
\newblock In \emph{{IEEE/CVF} Conference on Computer Vision and Pattern Recognition, {CVPR} 2022, New Orleans, LA, USA, June 18-24, 2022}, pages 13286--13296. {IEEE}, 2022.
\newblock \doi{10.1109/CVPR52688.2022.01294}.
\newblock URL \url{https://doi.org/10.1109/CVPR52688.2022.01294}.

\bibitem[Xiu et~al.(2023)Xiu, Yang, Cao, Tzionas, and Black]{xiu2023econ}
Yuliang Xiu, Jinlong Yang, Xu~Cao, Dimitrios Tzionas, and Michael~J. Black.
\newblock {ECON:} explicit clothed humans optimized via normal integration.
\newblock In \emph{{IEEE/CVF} Conference on Computer Vision and Pattern Recognition, {CVPR} 2023, Vancouver, BC, Canada, June 17-24, 2023}, pages 512--523. {IEEE}, 2023.
\newblock \doi{10.1109/CVPR52729.2023.00057}.
\newblock URL \url{https://doi.org/10.1109/CVPR52729.2023.00057}.

\bibitem[Xu et~al.(2024{\natexlab{a}})Xu, Cheng, Gao, Wang, Gao, and Shan]{xu2024instantmesh}
Jiale Xu, Weihao Cheng, Yiming Gao, Xintao Wang, Shenghua Gao, and Ying Shan.
\newblock Instantmesh: Efficient 3d mesh generation from a single image with sparse-view large reconstruction models.
\newblock \emph{arXiv preprint arXiv:2404.07191}, 2024{\natexlab{a}}.

\bibitem[Xu et~al.(2023)Xu, Tan, Luan, Bi, Wang, Li, Shi, Sunkavalli, Wetzstein, Xu, and Zhang]{Xu2023DMV3D}
Yinghao Xu, Hao Tan, Fujun Luan, Sai Bi, Peng Wang, Jiahao Li, Zifan Shi, Kalyan Sunkavalli, Gordon Wetzstein, Zexiang Xu, and Kai Zhang.
\newblock {DMV3D:} denoising multi-view diffusion using 3d large reconstruction model.
\newblock \emph{CoRR}, abs/2311.09217, 2023.
\newblock \doi{10.48550/ARXIV.2311.09217}.
\newblock URL \url{https://doi.org/10.48550/arXiv.2311.09217}.

\bibitem[Xu et~al.(2024{\natexlab{b}})Xu, Shi, Yifan, Peng, Yang, Shen, and Gordon]{xu2024grm}
Yinghao Xu, Zifan Shi, Wang Yifan, Sida Peng, Ceyuan Yang, Yujun Shen, and Wetzstein Gordon.
\newblock Grm: Large gaussian reconstruction model for efficient 3d reconstruction and generation.
\newblock \emph{arxiv: 2403.14621}, 2024{\natexlab{b}}.

\bibitem[Xue et~al.(2022)Xue, Li, Leutenegger, and Stueckler]{xue2022e-nr}
Yuxuan Xue, Haolong Li, Stefan Leutenegger, and Joerg Stueckler.
\newblock Event-based non-rigid reconstruction from contours.
\newblock In \emph{33rd British Machine Vision Conference 2022, {BMVC} 2022, London, UK, November 21-24, 2022}, page~78. {BMVA} Press, 2022.
\newblock URL \url{https://bmvc2022.mpi-inf.mpg.de/78/}.

\bibitem[Xue et~al.(2023)Xue, Bhatnagar, Marin, Sarafianos, Xu, Pons{-}Moll, and Tung]{xue2023nsf}
Yuxuan Xue, Bharat~Lal Bhatnagar, Riccardo Marin, Nikolaos Sarafianos, Yuanlu Xu, Gerard Pons{-}Moll, and Tony Tung.
\newblock {NSF:} neural surface fields for human modeling from monocular depth.
\newblock In \emph{{IEEE/CVF} International Conference on Computer Vision, {ICCV} 2023, Paris, France, October 1-6, 2023}, pages 15004--15014. {IEEE}, 2023.
\newblock \doi{10.1109/ICCV51070.2023.01382}.
\newblock URL \url{https://doi.org/10.1109/ICCV51070.2023.01382}.

\bibitem[Xue et~al.(2024)Xue, Li, Leutenegger, and Stückler]{xue2024e-nr-ijcv}
Yuxuan Xue, Haolong Li, Stefan Leutenegger, and Jörg Stückler.
\newblock Event-based non-rigid reconstruction of low-rank parametrized deformations from contours.
\newblock In \emph{International Journal of Computer Vision (IJCV)}. Springer Science and Business Media LLC, February 2024.
\newblock \doi{10.1007/s11263-024-02011-z}.
\newblock URL \url{http://dx.doi.org/10.1007/s11263-024-02011-z}.

\bibitem[Yang et~al.(2023)Yang, Luo, Xiu, Wang, Xu, and Fan]{yang_d-if_2023}
Xueting Yang, Yihao Luo, Yuliang Xiu, Wei Wang, Hao Xu, and Zhaoxin Fan.
\newblock D-{IF}: {Uncertainty}-aware {Human} {Digitization} via {Implicit} {Distribution} {Field}.
\newblock In \emph{2023 {IEEE}/{CVF} {International} {Conference} on {Computer} {Vision} ({ICCV})}, pages 9088--9098, Paris, France, October 2023. IEEE.
\newblock ISBN 9798350307184.
\newblock \doi{10.1109/ICCV51070.2023.00837}.
\newblock URL \url{https://ieeexplore.ieee.org/document/10377954/}.

\bibitem[Youwang et~al.(2024)Youwang, Oh, and Pons-Moll]{kim2024paintit}
Kim Youwang, Tae-Hyun Oh, and Gerard Pons-Moll.
\newblock Paint-it: Text-to-texture synthesis via deep convolutional texture map optimization and physically-based rendering.
\newblock In \emph{IEEE Conference on Computer Vision and Pattern Recognition (CVPR)}, 2024.

\bibitem[Yu et~al.(2021{\natexlab{a}})Yu, Ye, Tancik, and Kanazawa]{Yu2021pixelnerf}
Alex Yu, Vickie Ye, Matthew Tancik, and Angjoo Kanazawa.
\newblock pixelnerf: Neural radiance fields from one or few images.
\newblock In \emph{{IEEE} Conference on Computer Vision and Pattern Recognition, {CVPR} 2021, virtual, June 19-25, 2021}, pages 4578--4587. Computer Vision Foundation / {IEEE}, 2021{\natexlab{a}}.
\newblock \doi{10.1109/CVPR46437.2021.00455}.
\newblock URL \url{https://openaccess.thecvf.com/content/CVPR2021/html/Yu\_pixelNeRF\_Neural\_Radiance\_Fields\_From\_One\_or\_Few\_Images\_CVPR\_2021\_paper.html}.

\bibitem[Yu et~al.(2021{\natexlab{b}})Yu, Zheng, Guo, Liu, Dai, and Liu]{tao2021thuman2}
Tao Yu, Zerong Zheng, Kaiwen Guo, Pengpeng Liu, Qionghai Dai, and Yebin Liu.
\newblock Function4d: Real-time human volumetric capture from very sparse consumer rgbd sensors.
\newblock In \emph{IEEE Conference on Computer Vision and Pattern Recognition (CVPR2021)}, June 2021{\natexlab{b}}.

\bibitem[Yu et~al.(2023)Yu, Xu, Zhang, Liu, Ye, Wu, Yan, Zhu, Xiong, Liang, Chen, Cui, and Han]{yu2023mvimagenet}
Xianggang Yu, Mutian Xu, Yidan Zhang, Haolin Liu, Chongjie Ye, Yushuang Wu, Zizheng Yan, Chenming Zhu, Zhangyang Xiong, Tianyou Liang, Guanying Chen, Shuguang Cui, and Xiaoguang Han.
\newblock Mvimgnet: {A} large-scale dataset of multi-view images.
\newblock In \emph{{IEEE/CVF} Conference on Computer Vision and Pattern Recognition, {CVPR} 2023, Vancouver, BC, Canada, June 17-24, 2023}, pages 9150--9161. {IEEE}, 2023.
\newblock \doi{10.1109/CVPR52729.2023.00883}.
\newblock URL \url{https://doi.org/10.1109/CVPR52729.2023.00883}.

\bibitem[Yu et~al.(2024)Yu, Sattler, and Geiger]{Yu2024gof}
Zehao Yu, Torsten Sattler, and Andreas Geiger.
\newblock Gaussian opacity fields: Efficient high-quality compact surface reconstruction in unbounded scenes.
\newblock \emph{arXiv:2404.10772}, 2024.

\bibitem[Zablotskaia et~al.(2019)Zablotskaia, Siarohin, Zhao, and Sigal]{zablotskaia2019ubcfashion}
Polina Zablotskaia, Aliaksandr Siarohin, Bo~Zhao, and Leonid Sigal.
\newblock Dwnet: Dense warp-based network for pose-guided human video generation.
\newblock In \emph{30th British Machine Vision Conference 2019, {BMVC} 2019, Cardiff, UK, September 9-12, 2019}, page~51. {BMVA} Press, 2019.
\newblock URL \url{https://bmvc2019.org/wp-content/uploads/papers/1039-paper.pdf}.

\bibitem[Zeng et~al.(2017)Zeng, Song, Nie{\ss}ner, Fisher, Xiao, and Funkhouser]{zeng2016tsdf}
Andy Zeng, Shuran Song, Matthias Nie{\ss}ner, Matthew Fisher, Jianxiong Xiao, and Thomas Funkhouser.
\newblock 3dmatch: Learning local geometric descriptors from rgb-d reconstructions.
\newblock In \emph{CVPR}, 2017.

\bibitem[Zhang et~al.(2017)Zhang, Pujades, Black, and Pons{-}Moll]{zhang2017buff}
Chao Zhang, Sergi Pujades, Michael~J. Black, and Gerard Pons{-}Moll.
\newblock Detailed, accurate, human shape estimation from clothed 3d scan sequences.
\newblock In \emph{2017 {IEEE} Conference on Computer Vision and Pattern Recognition, {CVPR} 2017, Honolulu, HI, USA, July 21-26, 2017}, pages 5484--5493. {IEEE} Computer Society, 2017.
\newblock \doi{10.1109/CVPR.2017.582}.
\newblock URL \url{https://doi.org/10.1109/CVPR.2017.582}.

\bibitem[Zhang et~al.(2018{\natexlab{a}})Zhang, Isola, Efros, Shechtman, and Wang]{zhang2018lpips}
Richard Zhang, Phillip Isola, Alexei~A. Efros, Eli Shechtman, and Oliver Wang.
\newblock The unreasonable effectiveness of deep features as a perceptual metric.
\newblock In \emph{2018 {IEEE} Conference on Computer Vision and Pattern Recognition, {CVPR} 2018, Salt Lake City, UT, USA, June 18-22, 2018}, pages 586--595. Computer Vision Foundation / {IEEE} Computer Society, 2018{\natexlab{a}}.
\newblock \doi{10.1109/CVPR.2018.00068}.
\newblock URL \url{http://openaccess.thecvf.com/content\_cvpr\_2018/html/Zhang\_The\_Unreasonable\_Effectiveness\_CVPR\_2018\_paper.html}.

\bibitem[Zhang et~al.(2022)Zhang, Bhatnagar, Starke, Guzov, and Pons{-}Moll]{zhang2022couch}
Xiaohan Zhang, Bharat~Lal Bhatnagar, Sebastian Starke, Vladimir Guzov, and Gerard Pons{-}Moll.
\newblock {COUCH:} towards controllable human-chair interactions.
\newblock In Shai Avidan, Gabriel~J. Brostow, Moustapha Ciss{\'{e}}, Giovanni~Maria Farinella, and Tal Hassner, editors, \emph{Computer Vision - {ECCV} 2022 - 17th European Conference, Tel Aviv, Israel, October 23-27, 2022, Proceedings, Part {V}}, volume 13665 of \emph{Lecture Notes in Computer Science}, pages 518--535. Springer, 2022.
\newblock \doi{10.1007/978-3-031-20065-6\_30}.
\newblock URL \url{https://doi.org/10.1007/978-3-031-20065-6\_30}.

\bibitem[Zhang et~al.(2024)Zhang, Bhatnagar, Starke, Petrov, Guzov, Dhamo, P{\'{e}}rez{-}Pellitero, and Pons{-}Moll]{zhang2024force}
Xiaohan Zhang, Bharat~Lal Bhatnagar, Sebastian Starke, Ilya Petrov, Vladimir Guzov, Helisa Dhamo, Eduardo P{\'{e}}rez{-}Pellitero, and Gerard Pons{-}Moll.
\newblock {FORCE:} dataset and method for intuitive physics guided human-object interaction.
\newblock \emph{CoRR}, abs/2403.11237, 2024.
\newblock \doi{10.48550/ARXIV.2403.11237}.
\newblock URL \url{https://doi.org/10.48550/arXiv.2403.11237}.

\bibitem[Zhang et~al.(2018{\natexlab{b}})Zhang, Zhang, Zhang, Tenenbaum, Freeman, and Wu]{genre}
Xiuming Zhang, Zhoutong Zhang, Chengkai Zhang, Joshua~B Tenenbaum, William~T Freeman, and Jiajun Wu.
\newblock {Learning to Reconstruct Shapes From Unseen Classes}.
\newblock In \emph{Advances in Neural Information Processing Systems (NeurIPS)}, 2018{\natexlab{b}}.

\bibitem[Zhang et~al.(2023)Zhang, Yang, and Yang]{zhang2023sifu}
Zechuan Zhang, Zongxin Yang, and Yi~Yang.
\newblock {SIFU:} side-view conditioned implicit function for real-world usable clothed human reconstruction.
\newblock \emph{CoRR}, abs/2312.06704, 2023.
\newblock \doi{10.48550/ARXIV.2312.06704}.
\newblock URL \url{https://doi.org/10.48550/arXiv.2312.06704}.

\bibitem[Zhou and Tulsiani(2023)]{zhou2023sparsefusion}
Zhizhuo Zhou and Shubham Tulsiani.
\newblock Sparsefusion: Distilling view-conditioned diffusion for 3d reconstruction.
\newblock In \emph{CVPR}, 2023.

\bibitem[Zou et~al.(2023)Zou, Yu, Guo, Li, Liang, Cao, and Zhang]{zou2023triplanegaussian}
Zi{-}Xin Zou, Zhipeng Yu, Yuan{-}Chen Guo, Yangguang Li, Ding Liang, Yan{-}Pei Cao, and Song{-}Hai Zhang.
\newblock Triplane meets gaussian splatting: Fast and generalizable single-view 3d reconstruction with transformers.
\newblock \emph{CoRR}, abs/2312.09147, 2023.
\newblock \doi{10.48550/ARXIV.2312.09147}.
\newblock URL \url{https://doi.org/10.48550/arXiv.2312.09147}.

\end{thebibliography}



\clearpage
\newpage

\appendix
\onecolumn

\addcontentsline{toc}{section}{Appendix} 
\renewcommand \thepart{} 
\renewcommand \partname{}
\part{\Large{\centerline{Appendix}}}
\parttoc

\newpage
\section{Implementation Details}
\label{secsupp:formulation}
\subsection{Training Details}
\label{suppsec:training_details}
As described in~\cref{subsec:exp-setup}, we use an effective batch size of 256.
Each batch involved sampling 4 orthogonal images with zero elevation angle as target views $\mathbf{x}^{\text{tgt}}_0$, and 12 additional images as novel views $\mathbf{x}^{\text{novel}}_0$ to supervise the 3D generative model~\cref{eq:loss_all}. The hyperparameters for training~\cref{eq:loss_all} were set as follows: $\lambda_1=1.0$, $\lambda_2=1.0$, and $\lambda_3=100.0$.

During training, we employed the standard DDPM scheduler~\cite{Ho2020DDPM} to construct noisy target images $\mathbf{x}_{t}^{\text{tgt}}$. The maximum diffusion step $T$ is set to $1000$. At inference time, we use DDIM scheduler~\cite{song2021ddim} to perform faster reverse sampling. The reverse steps is set to 50 in following experiments. The text prompt $y$ used in our multi-view diffusion model(\cref{eq:one-step-mvd}) is set to "Photorealistic 3D human" for both training and inference across all subjects.

\subsection{Joint Framework}
\textbf{Implementation.} Our 2D multi-view diffusion model $\boldsymbol{\epsilon}_{\theta}$ is a latent diffusion model~\cite{Wang2023ImageDream}. Thus, we use the frozen VAE in~\cite{Rombach2022StableDiffusion} to obtain input $\mathbf{x}_t^{\text{tgt}}$ in image space for the 3D generative model $g_{\phi}$ and encode refined $\tilde{\mathbf{x}}_{t}^{\text{tgt}}$ back to latent space for $\boldsymbol{\epsilon}_{\theta}$. We extract triangle mesh from predicted Gaussian splats using Gaussian Opacity field~\cite{Yu2024gof} and TSDF~\cite{zeng2016tsdf}. Please refer to \cref{suppsec:mesh_via_tsdf} for more details. 

\subsection{Generative 3D-GS Reconstruction Model}
\label{sec_supp:3d_state_diffusion}

In this section, we provide details about our 3D generative model $g_{\phi}$ in ~\cref{eq:diffusion_wo_x0} and ~\cref{eq:diffusion_with_x0_clean} as well as the $\renderer(\circ)$. Following ~\cite{anciukevicius2023renderdiffusion, karnewar2023holodiffusion,Tewari2023DiffusionWithForward}, we learn the 3D generative model by adding and removing noise on the rendered 2D images from a 3D representation. A pseudo algorithm of the training and sampling process of our 3D generative model can be found in \cref{algm:train_app_3d}.

Since we integrate both function into the reverse sampling process iteratively (eq.~\ref{eq:ddpm_reverse}), we expect them be efficient and fast to execute. 
Tewari et al.~\cite{Tewari2023DiffusionWithForward} base their model on pixelNeRF~\cite{Yu2021pixelnerf}, which is a generalizable NeRF~\cite{Mildenhall2020NeRF} conditioned on a context view image.
We adopt 3D Gaussian Splats~\cite{Kerbl20233dgs} as our 3D state representation $\gsplat$ due to its efficiency and simplicity. 
Our $\renderer(\circ)$ is the differentiable rasteraizer accelerated and implemented in CUDA, which achieves around $2700$ times faster rendering than volume-rendering-based $\renderer(\circ)$ in ~\cite{Mildenhall2020NeRF, Tewari2023DiffusionWithForward, Yu2021pixelnerf}.

For sampling the 3D State $\mathbf{S}$ from $\mathbf{x}_t^{\text{tgt}} $, $\tilde{\mathbf{x}}_{0}^{\text{tgt}}$, $\mathbf{x}^{\text{c}}$, and $t$ (eq.~\ref{eq:diffusion_with_x0_clean}), we adopt the time-conditioned UNet-Transformer architecture~\cite{Rombach2022StableDiffusion} due to the efficiency of convolutional layers and the scalability of transformers. For enabling the awareness of camera poses in the encoding process, we concatenate the Pl{\"u}cker Camera Ray Embedding $\{\mathbf{o}_i \times \mathbf{d}_i, \mathbf{d}_i \}$~\cite{Tang2024LGM, Xu2023DMV3D} with the image $\mathbf{x}_t^{\text{tgt}}$ and $\tilde{\mathbf{x}}_{0}^{\text{tgt}}$. To enhance the control ability of context view in the 3D generation process, we additionally concatenate the clear context view $\rvx^{\text{c}}$ with target images $\mathbf{x}_t^{\text{tgt}} $ following~\cite{Wang2023ImageDream}. This operation enables 3D dense self-attention process between the input multi-view target images and the clear context view image, provides pixel-level local conditional signal. Since the camera pose of context view $\mathbf{x}^{c}$ is unknown, we use the $0$-vector as its embedding. 

 \begin{figure}[!htp]
    \centering 
     \caption{Visualization of intermediate sampling steps from a Gaussian Noise ($t=1000$) to the last denoising step ($t=0$). From top to bottom: current state $\mathbf{x}^{\text{tgt}}_{t}$, estimated clear view by 2D diffusion models $\tilde{\mathbf{x}}^{\text{tgt}}_{0}$, and corrected clear view by generated 3D Gaussian Splatting $\hat{\mathbf{x}}^{\text{tgt}}_{0}$. Our 2D diffusion model $\boldsymbol{\epsilon}_{\phi}(\circ)$ already provides strong multi-view prior at an early stage with large $t$. Our 3D reconstruction model $\mathbf{g}_{\phi}(\circ)$ can correct the inconsistency in $\tilde{\mathbf{x}}^{\text{tgt}}_{0}$ illustrated in red circle. }
    \includegraphics[width=1.5\textwidth, angle=270]{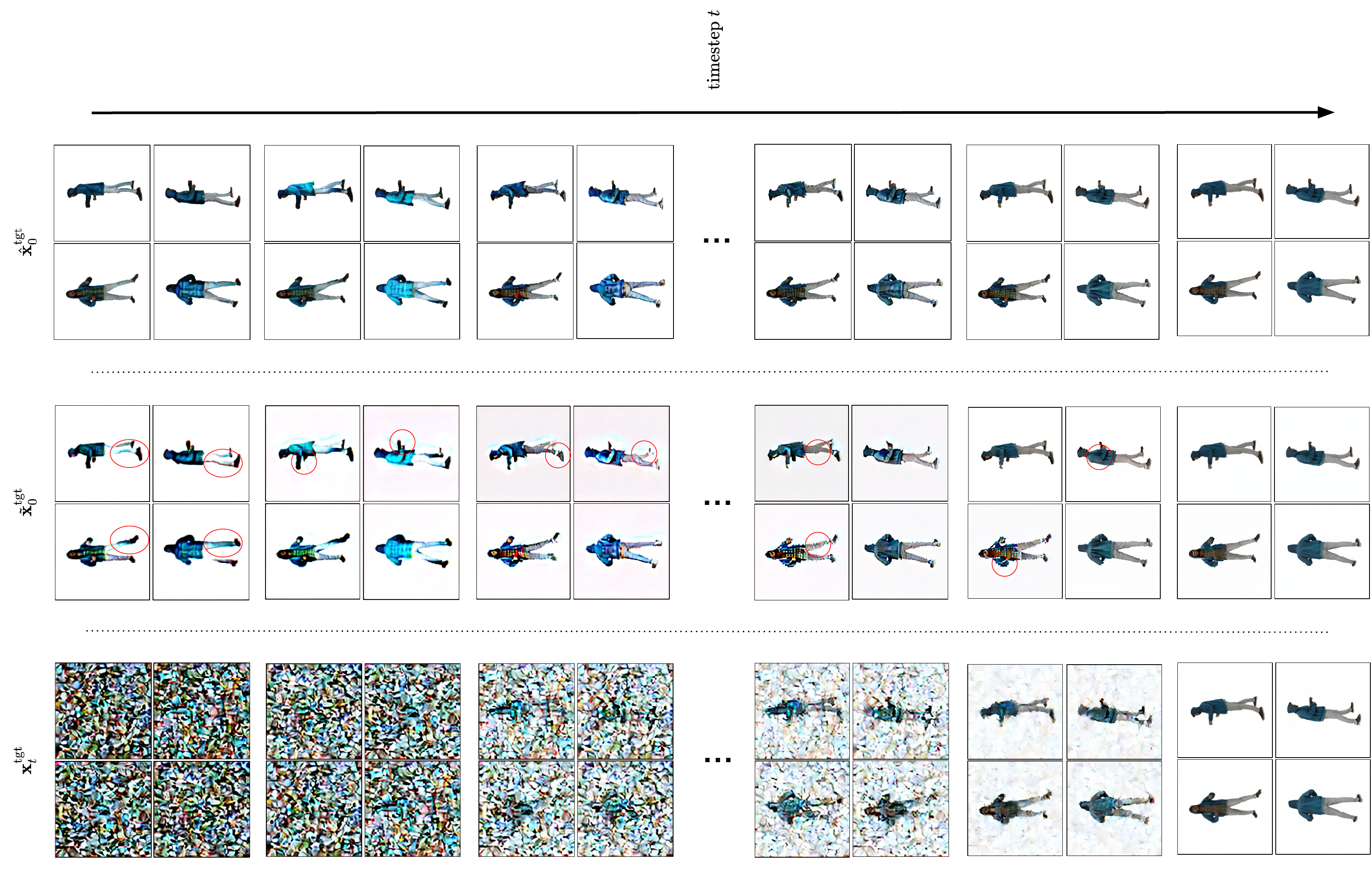}
    \label{fig:ddim_intermediate_visualization}
\end{figure}

\algrenewcommand\algorithmicrequire{\textbf{Input:}} 
\algrenewcommand\algorithmicensure{\textbf{Output:}}
\algrenewcommand\algorithmicindent{0.5em}%
\begin{figure}[H]
\begin{minipage}[t]{0.49\textwidth}
\begin{algorithm}[H]
  \caption{Learn 3D distribution} \label{alg:pure_training_3d}
  \small
  \begin{algorithmic}[1]
  \Require Dataset of posed multi-view images $\rvx^{\text{tgt}}_{0}$, $\pi^{\text{tgt}}$, $\rvx^{\text{novel}}_{0}$, $\pi^{\text{novel}}$, a context image $\rvx^{\text{c}}$
  \Ensure Optimized 3D State diffusion network $g_{\phi}$
    \Repeat
      \State $\{\rvx^{\text{tgt}}_{0}, \rvx^{\text{novel}}_{0}, \rvx^{\text{c}}, y\} \sim q(\{\rvx^{\text{tgt}}_{0}, \rvx^{\text{novel}}_{0}, \rvx^{\text{c}}, y\} )$
      \State $t \sim \mathrm{Uniform}(\{1, \dotsc, T\})$; $\boldsymbol{\epsilon} \sim \mathcal{N}(\mathbf{0},\mathbf{I})$
      \State $\mathbf{x}^{\text{tgt}}_{t} = \sqrt{\bar\alpha_t} \rvx_{0}^{\text{tgt}} + \sqrt{1-\bar{\alpha}_t}\boldsymbol{\epsilon}$
      \State $\hat{\gsplat} = g_{\phi}\left(\mathbf{x}^{\text{c}}, \mathbf{x}_t^{\text{tgt}}, t\right) $
      \State $\{ \hat{\mathbf{x}}_{0}^{\text{tgt}}, \hat{\mathbf{x}}_{0}^{\text{novel}}\} = \renderer\left(\hat{\gsplat}, \{\pi^{\text{tgt}}, \pi^{\text{novel}}\} \right)$
      \State Compute loss $\mathcal{L}_{gs}$ (~\cref{eq:loss_3D_diffusion}) 
      \State Gradient step to update $g_\phi$
      
    \Until{converged}
  \end{algorithmic}
  \label{algm:train_app_3d}
\end{algorithm}
\end{minipage}
\hfill
\begin{minipage}[t]{0.49\textwidth}
\begin{algorithm}[H]
  \caption{Sample from 3D distribution} \label{alg:pure_sampling_3d}
  \small
  \begin{algorithmic}[1]
  \Require A context image $\rvx^c$; Converged 3D diffusion model $g_{\phi}$
  \Ensure A 3D Gaussian Avatar $\gsplat$ of the 2D image $\rvx^c$
  
    \vspace{.015in}
    \State $\rvx_T^{\text{tgt}} \sim \mathcal{N}(\mathbf{0}, \mathbf{I})$
    \For{$t=T, \dotsc, 1$}
      \State  $\hat{\gsplat} = g_{\phi}\left(\mathbf{x}^{\text{c}}, \mathbf{x}_t^{\text{tgt}}, t\right) $ 
      \State $\hat{\mathbf{x}}_{0}^{\text{tgt}} = \renderer\left(\hat{\gsplat}, \pi^{\text{tgt}}\right)$
      \State $\mu_{t-1}(\rvx_t^\text{tgt}, \hat{\rvx}_0^\text{tgt}) = \frac{\sqrt{\alpha_{t}}\left(1-\bar{\alpha}_\text{t-1}\right)}{1-\bar{\alpha}_{t}} \rvx^{\text{tgt}}_{t} + \frac{\sqrt{\bar{\alpha}_\text{t-1}} \beta_{t}}{1-\bar{\alpha}_{t}} \hat{\rvx}_{0}^{\text{tgt}}$
      \State $\rvx^{\text{tgt}}_{t-1} \sim \mathcal{N}\left(\mathbf{x}^{\text{tgt}}_{t-1}; \tilde{\bm{\mu}}_{t}\left(\rvx^{\text{tgt}}_t, \hat\rvx^{\text{tgt}}_{0} \right), \tilde{\beta}_{t-1}\mathbf{I}) \right)$
    \EndFor
    \State \textbf{return} $\gsplat =  g_{\phi}\left(\mathbf{x}_{0}^{\text{tgt}}, \mathbf{x}^{\text{c}}, t=0\right) $
  \end{algorithmic}
\end{algorithm}
\label{algm:pure_3ddiffusion}
\end{minipage}
\vspace{-1em}
\end{figure}

\subsection{Textured Mesh Extraction}
\label{suppsec:mesh_via_tsdf}
Gaussian Opacity Fields~\cite{Yu2024gof} enables extraction of triangle meshes from an existing 3D Gaussian Splatting. However, because the location of 3D-Gaussian Splats is not necessary to be on the real surface, we observe that the extracted meshes as well as the rendered depth maps are noisy. Since our method generate realistic RGB images, we use PiFU-HD~\cite{saito2020pifuhd} to estimate the normals and use Bilateral Normal Integration (BiNI)~\cite{bini2022cao} to refine the noisy rendered depth with the estimated normal. As we only want the estimated normal to denoise the rendered depth map instead of modifying geometry, we set up the hyperparameter in BiNI with $\lambda=1\times10^4$. Such a large number ensures that the normal map is not used to modify the geometry but just regularize the depth map.

Assuming we have a generated 3D-Gaussian Splats $\gsplat$ from $g_{\phi}(\circ)$ and $n$ camera views ${\pi^1, \pi^2, ... \pi^n}$, we obtain $n$ paris of posed RGB-D images by \emph{Gaussian Splatting}, \emph{Normal Estimation}, and \emph{Bilateral Normal Integration}. Finally, we perform volumetric TSDF fusion~\cite{zeng2016tsdf} to obtain high quality textured mesh from $n$ pairs RGB-D images. Given generated 3D-GS $\gsplat$, we set up 36 views to obtain the refined RGB-D image pairs. The rendering view of each camera i can be calculated as:
\begin{align}
    \text{elevation}_i &= -\frac{1}{4} \pi + \frac{1}{4}\pi  * \frac{i}{36}, \\
    \text{azimuth}_i &= 0 + 3\pi  * \frac{i}{36}.
\end{align}

\newpage
\section{Comparison}
\subsection{Qualitative Comparison}
\label{secsupp:comparison_img23d}

\begin{figure*}[!htp]
\centering
\begin{tabular}{@{}c@{\,}}
\hspace{-0.65in} \includegraphics[width=1.16\textwidth]{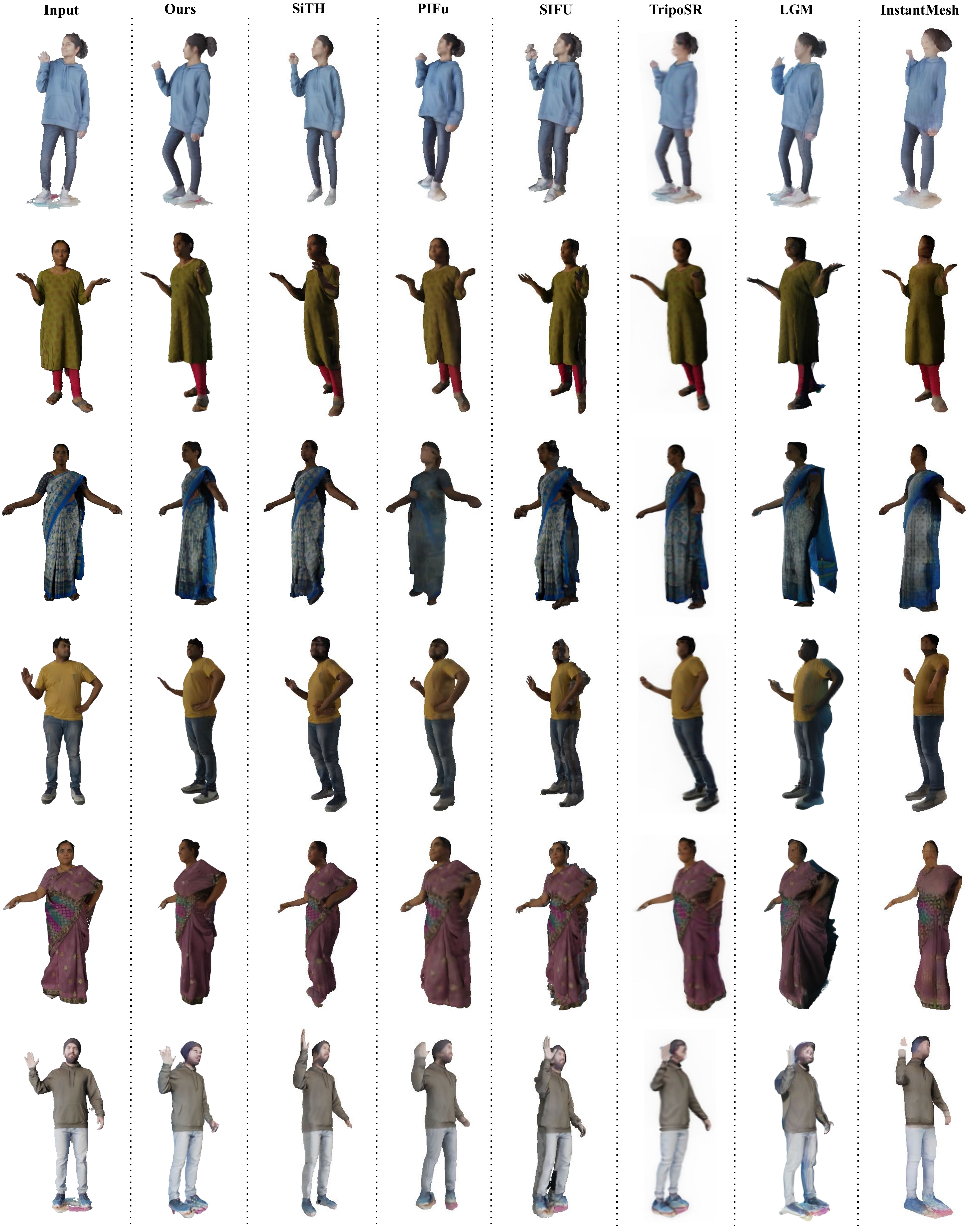}\\
\end{tabular}
\caption{Qualitative comparison on Sizer~\cite{tiwari2020sizer} and IIIT~\cite{jinka2023iiit}. }
    \label{fig:comparison_more}
\end{figure*}

\newpage

\begin{figure*}[!htp]
\centering
\begin{tabular}{@{}c@{\,}}
\hspace{-0.65in} \includegraphics[width=1.16\textwidth]{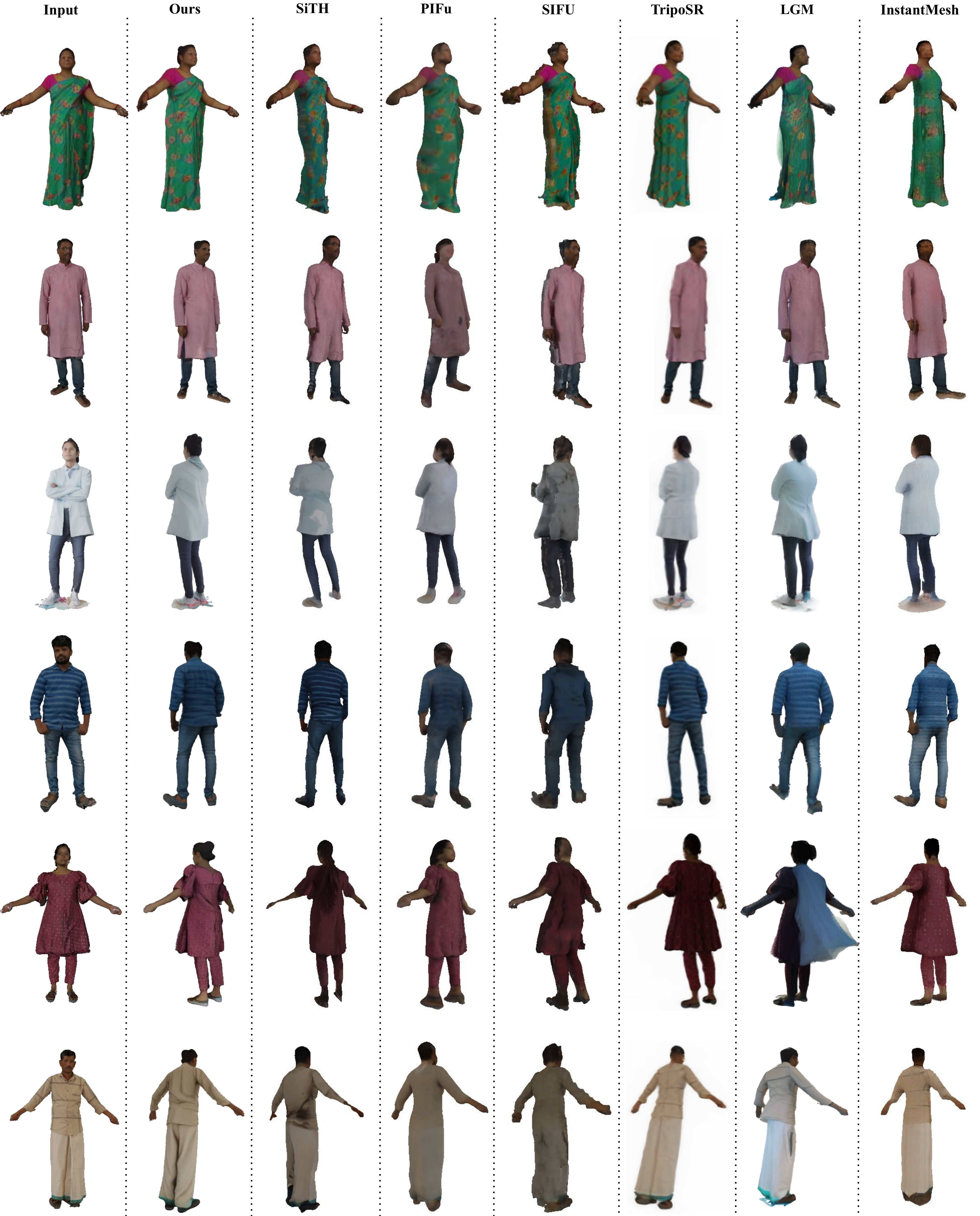}\\
\end{tabular}
\caption{Qualitative comparison on Sizer~\cite{tiwari2020sizer} and IIIT~\cite{jinka2023iiit}.}
    \label{fig:comparison_more2}
\end{figure*}

\newpage
\section{More Qualitative Results}
\label{suppsec:moreresults}
In this section, we show more qualitative results on in-the-wild data, UBC fashion dataset~\cite{zablotskaia2019ubcfashion}, GSO dataset~\cite{downs2022gso}, and human-object interaction data~\cite{xie2023template_free}.

\subsection{In-the-wild Data}
\begin{figure}[!htp]
    \centering
    \includegraphics[width=0.98\textwidth]{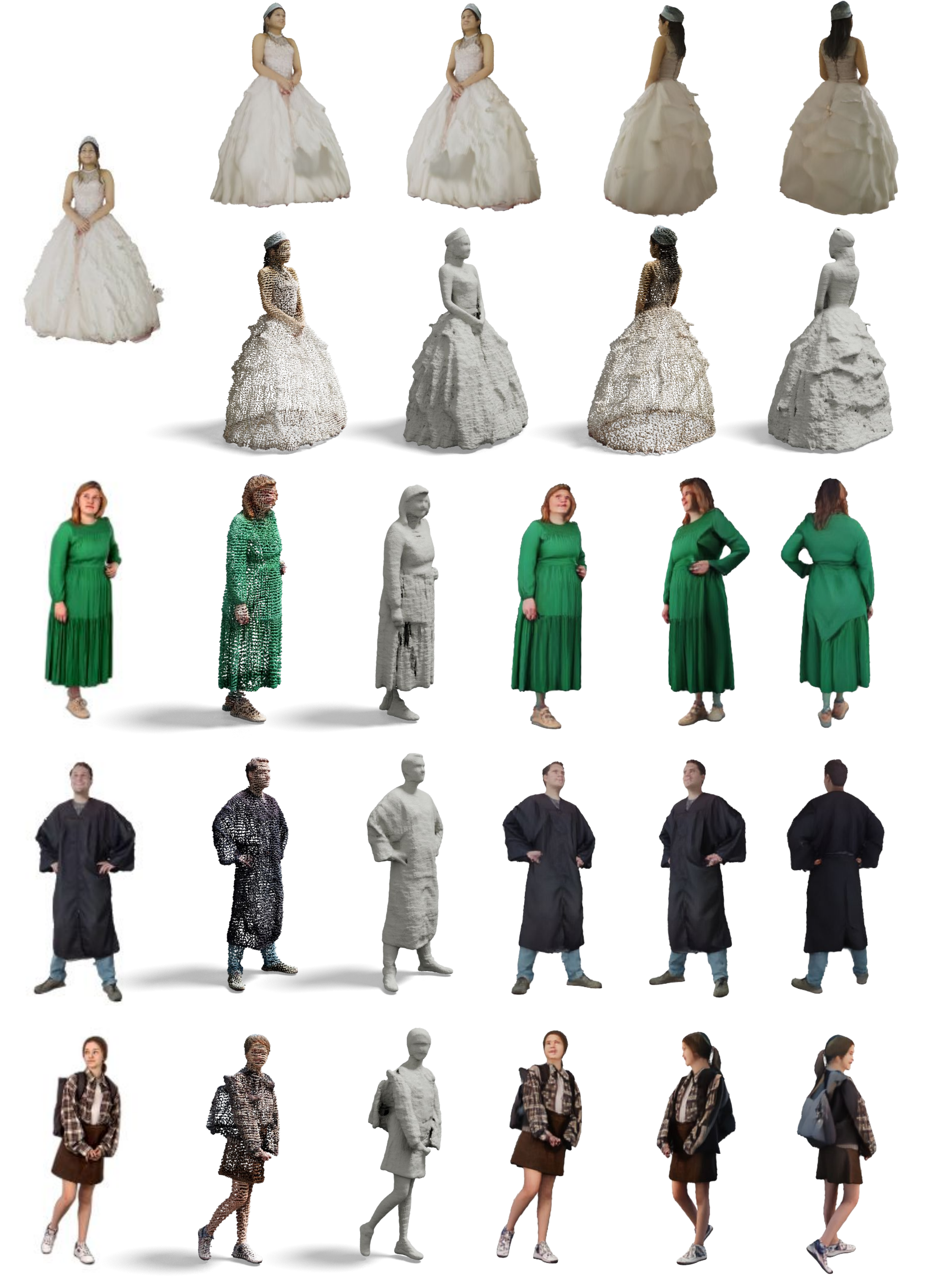}
    \caption{Qualitative results on unseen data during training. Input image is in left column. Our method successfully reconstructs different degree of loose clothing.}
    \label{fig:results_more}
\end{figure}

\begin{figure}[!htp]
    \centering
    \includegraphics[width=1\textwidth]{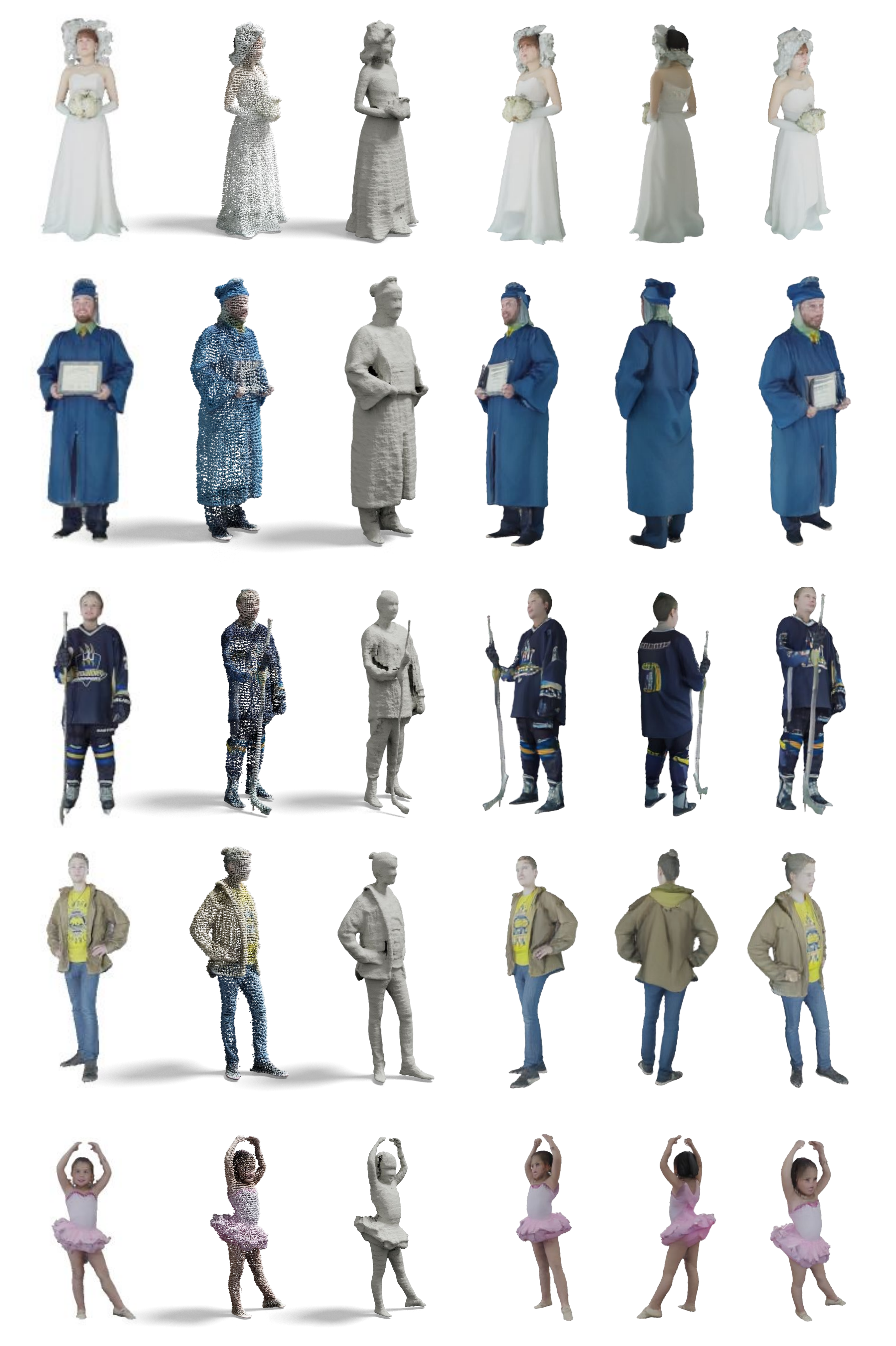}
    \caption{Qualitative results on more unseen data during training. Input image is in left column. Our method successfully reconstructs different types of clothing, including casual, sport, suits, custom, etc., in both appearance and geometry.}
    \label{fig:results_more2}
\end{figure}

\begin{figure}[!htp]
    \centering
    \includegraphics[width=1\textwidth]{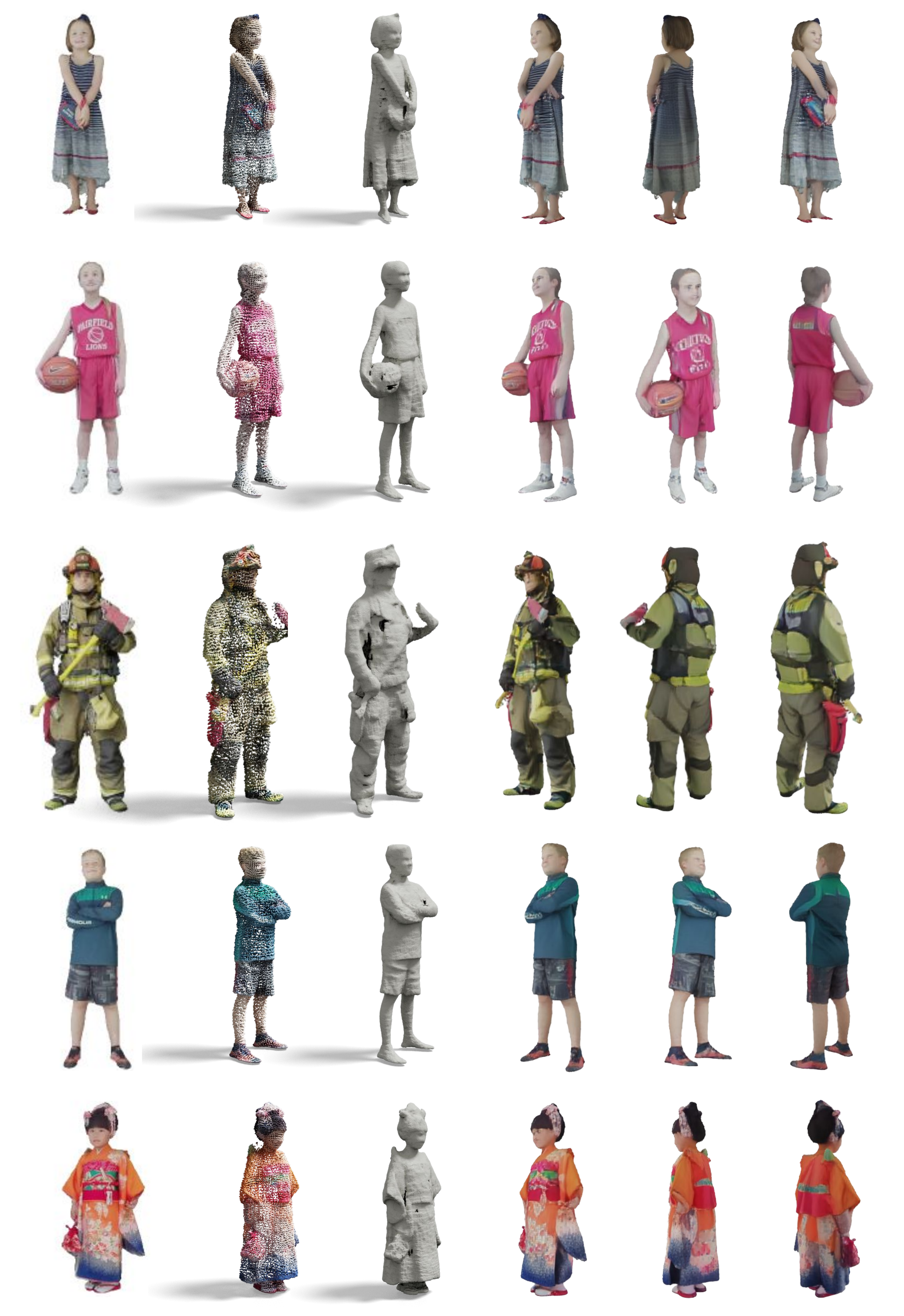}
    \caption{Qualitative results on more unseen data during training. Input image is in left column. Our method successfully reconstructs clothing and interacting objects (racket and bag here) in both appearance and geometry.}
    \label{fig:results_more3}
\end{figure}

\begin{figure}[!htp]
    \centering
    \includegraphics[width=1\textwidth]{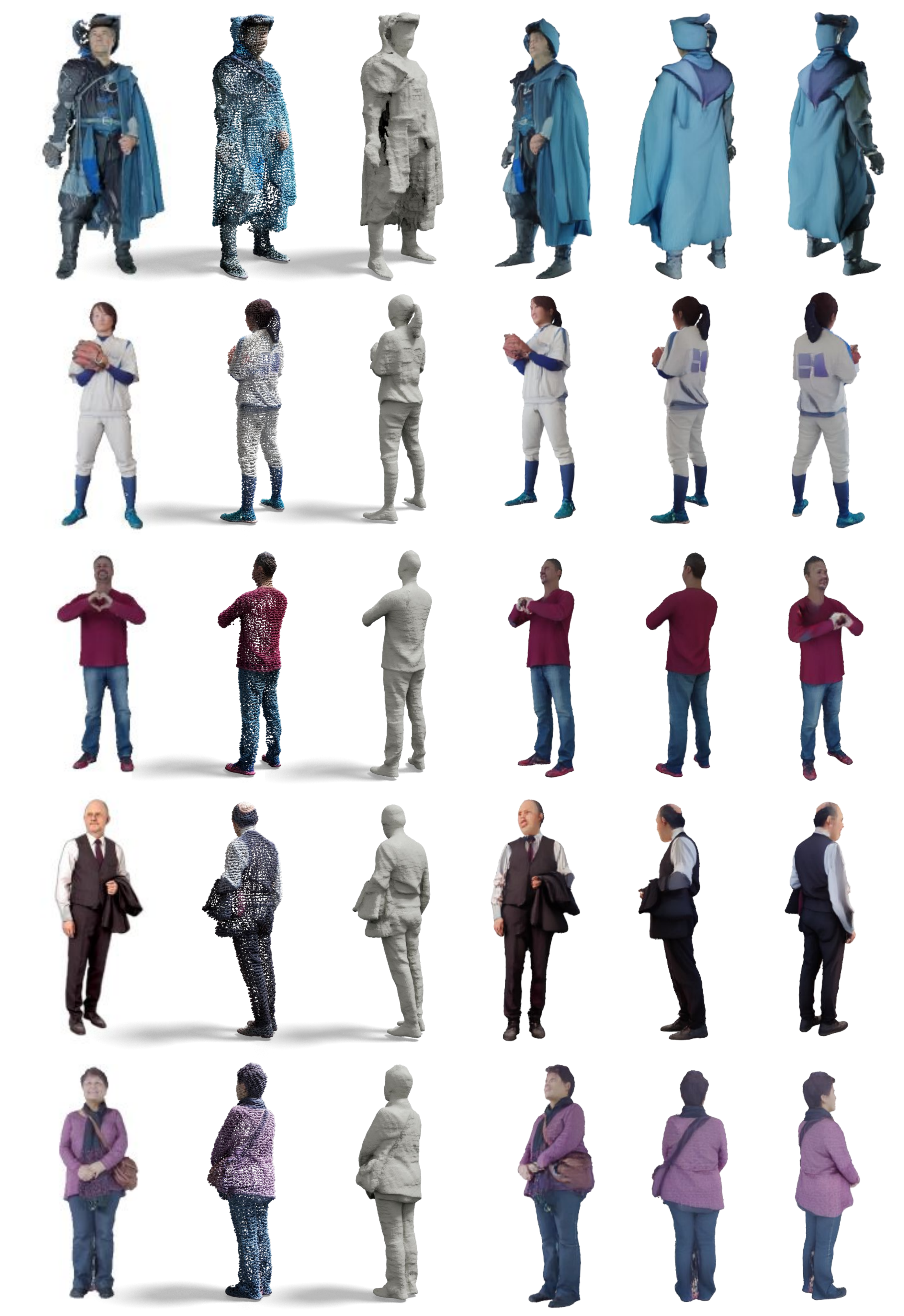}
    \caption{Qualitative results on more unseen data during training. Input image is in left column. Our method successfully reconstructs rarely seen suits and objects, in both appearance and geometry.}
    \label{fig:results_more4}
\end{figure}

\subsection{UBC Fashion Dataset}
In this section, we show qualitative result of our model on UBC fashion~\cite{zablotskaia2019ubcfashion} dataset. The input images are the first frame extracted from each video in the dataset.

\begin{figure}[!htp]
    \centering
    \vspace{10pt}
    \includegraphics[width=\textwidth]{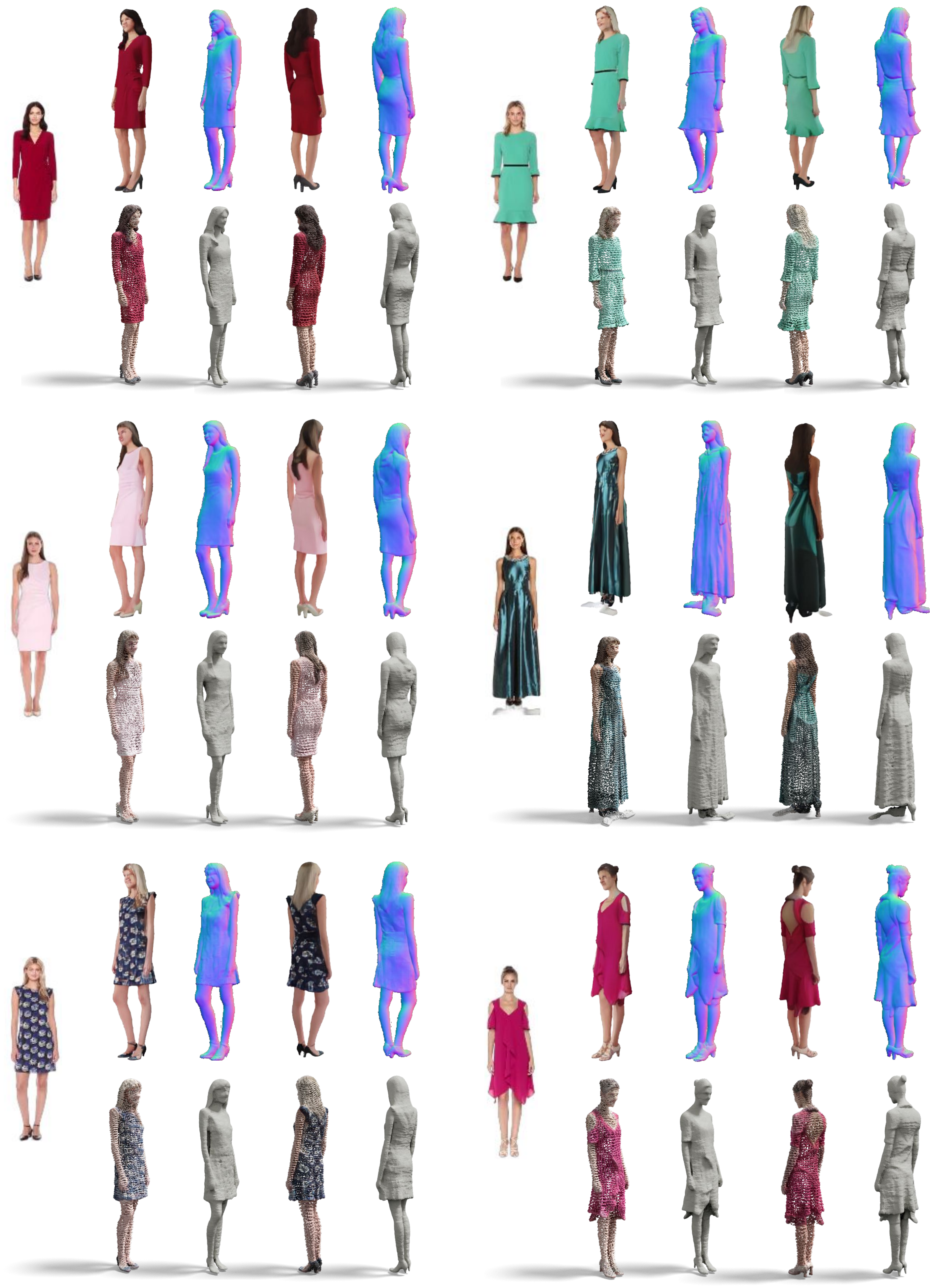}
    \caption{Qualitative results on UBC fashion~\cite{zablotskaia2019ubcfashion} dataset. Results demonstrate that our model generalizes well to real world images in both geometry and appearance.}
    \label{fig:results_ubc_fashion}
\end{figure}

\newpage
\subsection{Google Scan Objects (GSO)}

\begin{figure}[!htp]
    \centering
    \includegraphics[width=0.93\textwidth]{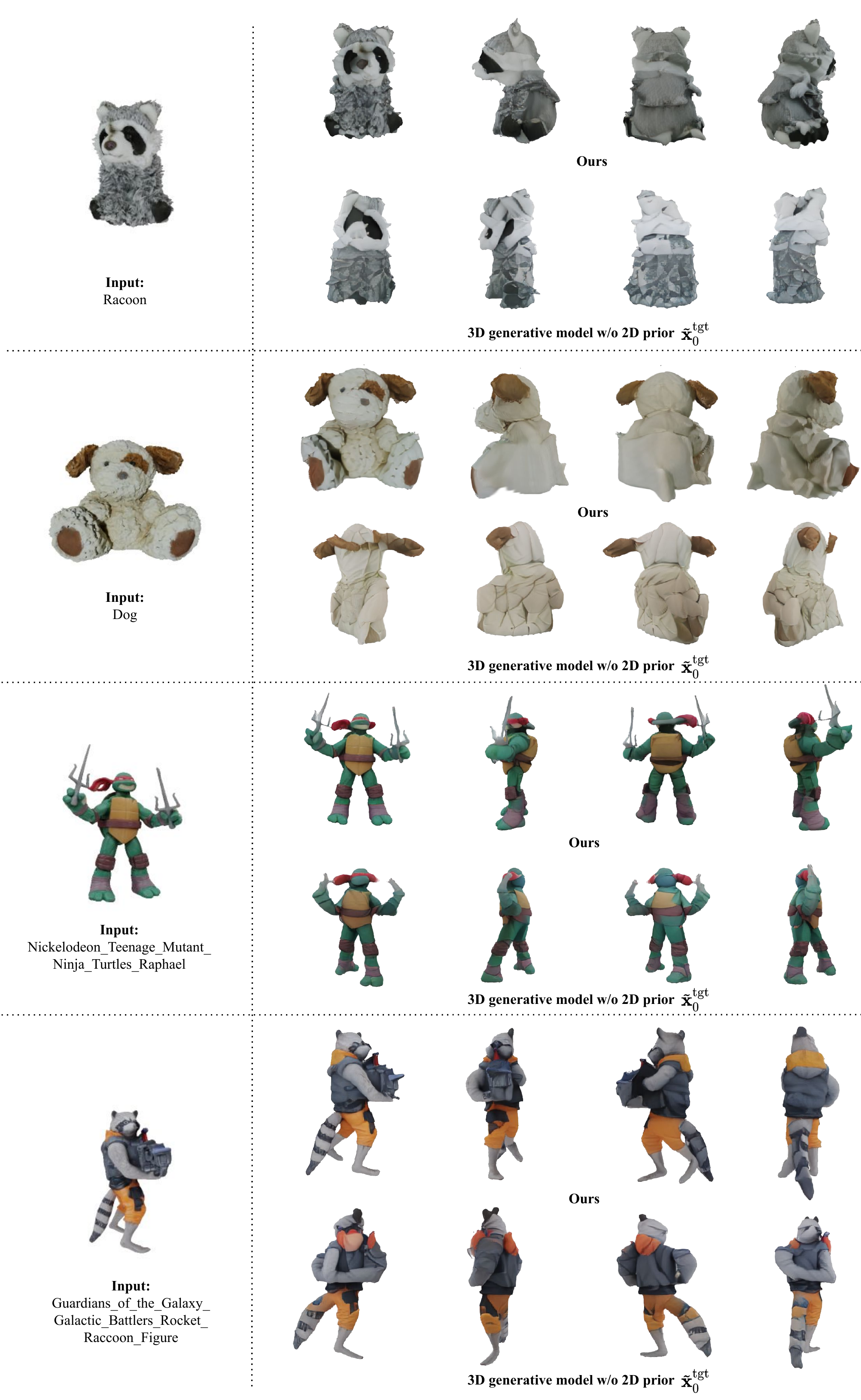}
    \caption{Ablation study: benefit of 2D multi-view prior $\tilde{\mathbf{x}}_{0}^{\text{tgt}}$ in 3D generation. The 2D prior from 2D diffusion model is essntial for generalization on general objects dataset GSO~\cite{downs2022gso}.}
    \label{fig:ablate_multiview_cond_obj_supp}
\end{figure}

\newpage
\subsection{Human Object Interaction}

\begin{figure}[!htp]
    \centering
    \vspace{10pt}
    \includegraphics[width=\textwidth]{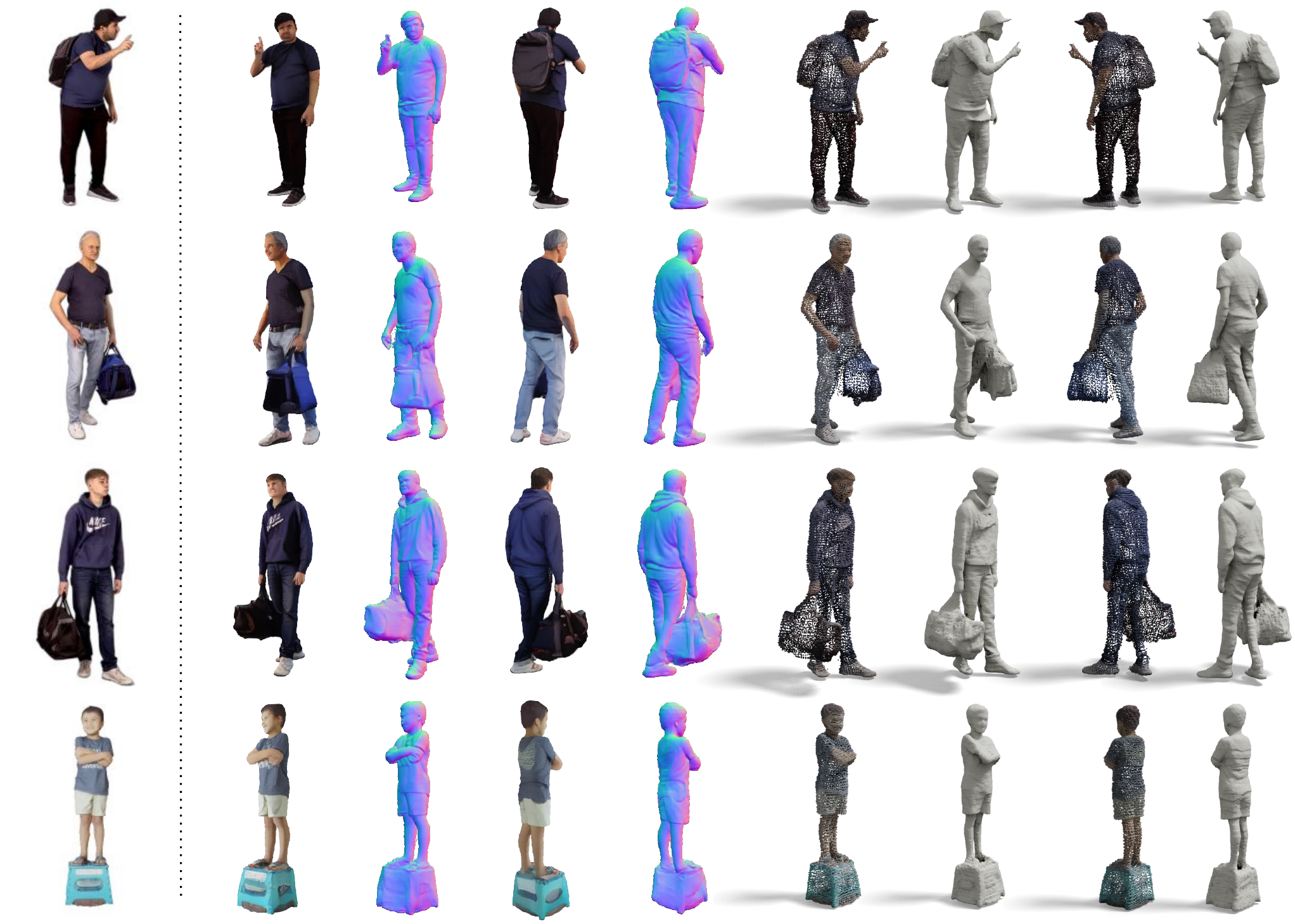}
    \caption{Qualitative results of Human-Object Interaction reconstruction on online stock images. Results show that our model is able to generalize to casual human-object-interactions. }
    \label{fig:results_hoi_stockimages}
\end{figure}

\begin{figure}[!htp]
    \centering
    \vspace{2cm}
    \includegraphics[width=\textwidth]{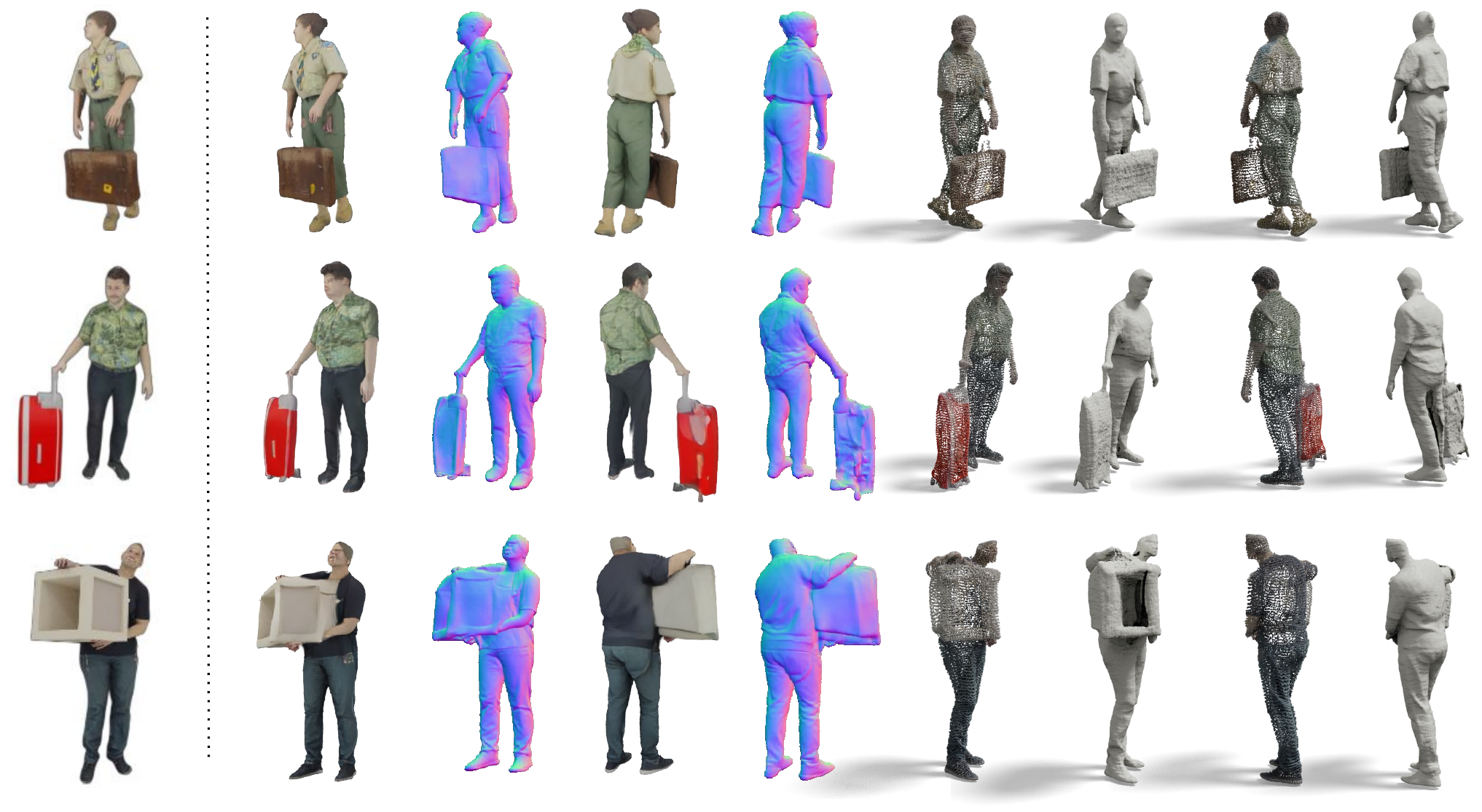}
    \caption{Qualitative results of Human-Object Interaction reconstruction on ProciGen~\cite{xie2023template_free} dataset. Results demonstrate that our model can reconstruct some simple Human-Object-Interaction images with large objects.}
    \label{fig:results_procigen}
\end{figure}

\subsection{Generative Power in Reconstruction}
\label{suppsec:generative_power}
\begin{figure}[!htp]
    \centering
    \vspace{0.5cm}
    \includegraphics[width=0.8\textwidth]{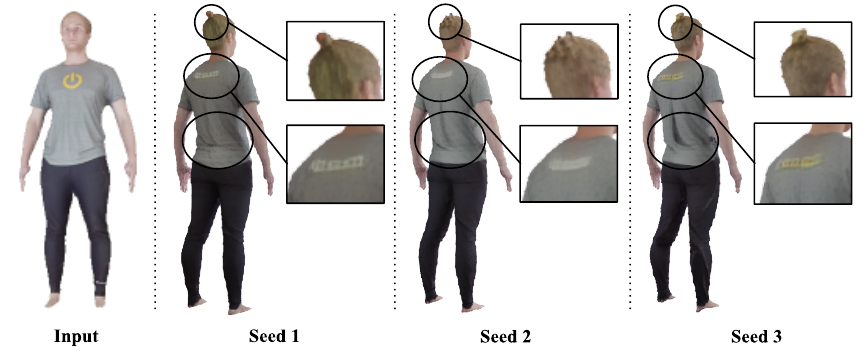}
    \caption{Our model learns 3D distribution. By different sampling from the learned distribution, we obtain diverse yet plausible 3D representations. The generative power is a key to generate clear self-occluded region, which is impossible in non-generative reconstruction approaches~\cite{saito2019pifu, saito2020pifuhd, tochilkin2024triposr, zhang2023sifu}.}
    \label{fig:GenerativePower}
\end{figure}

Our model learns a conditional distribution of the 3D representation given 2D context image. Thus, by sampling from the distribution with different seed, we obtain diverse yet plausible 3D representation. As illustrated in~\cref{fig:GenerativePower}, the appearance of the occluded region (back side of subject) is different with different sampling in hair styple, texture, and cloth wrinkles.

The generative power of our approach is the key to generate clear self-occluded regions, which is impossible by non-generative reconstruction methods~\cite{saito2019pifu, saito2020pifuhd, tochilkin2024triposr, zhang2023sifu}. As shown in~\cref{fig:compare_baselines} and~\cref{fig:comparison_more}, non-generative approaches tend to generate blurry self-occluded results because they cannot sample from distribution but only regress to a mean value of the training datasets.

\newpage
\section{Dataset Overview}
To ensure robust performance and generalization, we train our model on a combined dataset comprising 3520 scans from publicly available datasets~\cite{han20232k2k, ho2023customhuman, su2022thuman3, tao2021thuman2} and 2320 scans from commercial 3D human datasets~\cite{axyz, treedy, twindom, renderpeople}. These datasets encompass a diverse range of body shapes, genders, ages, clothing, accessories, and interacting objects. For further details and examples of our training datasets, please refer to section~\ref{suppsec:traninigdataset}. 
All 3D scans are rendered into RGB-A images using BlenderProc~\cite{denninger2023blenderproc} along a spiral path as described in~\cref{eq:spiral_rendering_path}.
\\
For evaluation, we note that the commonly-used CAPE dataset~\cite{ma2019cape, ponsmoll2017clothcap, zhang2017buff} in previous works~\cite{ho2023sith, xiu2023icon,xiu2023econ, zhang2023sifu} often contains artifacts in scans, such as holes, and not all 3D scans are fully publicly available.
To effectively and fairly evaluate performance, we propose using Sizer~\cite{tiwari2020sizer} and IIIT-Human~\cite{jinka2023iiit} datasets, from which we randomly sample 150 scans each for evaluation. While Sizer~\cite{tiwari2020sizer} provides scans with normal human appearance similar to our training datasets, IIIT-Human~\cite{jinka2023iiit} can be considered as out-of-distribution (o.o.d.) evaluation dataset due to its inclusion of unseen clothing types, such as traditional Indian suits. For more additional examples and analysis, please see \cref{fig:compare_baselines} and ~\cref{suppsec:evaluationdataset}.

\subsection{Training Dataset}
\label{suppsec:traninigdataset}

\paragraph{Datasets}

To prevent the overfitting of our large neural network, namely the 2D multi-view diffusion models $\boldsymbol{\epsilon}_{\theta}(\circ)$ and 3D generative models $\mathbf{g}_{\phi}(\circ)$, we train on as much data as we can. Unlike general objects community which has massive dataset such as Objaverse (800K) and Objaverse-XL (10M)~\cite{Deitke2023Objaverse})\, OmniObject3D (6K)~\cite{wu2023omniobject3d}, MVImageNet (87K)~\cite{yu2023mvimagenet}, we don't have a single 3D human dataset available at such a scale.
To collect data as much as possible, we collect both following public datasets and commercial human scans.

We collect several publicly available datasets inluding 2k2k (2K)~\cite{han20232k2k}, CustomHuman (640)~\cite{ho2023customhuman}, Thuman2.0 (520)~\cite{tao2021thuman2}, and Thuman3.0 (360)~\cite{su2022thuman3}). Among them, CustomHuman, Thuman2.0, and Thuman3.0 have more repeating subjects with different poses, which have less diverse subject appearance compared to 2k2k. It is worth mentioning that 2k2k~\cite{han20232k2k} is a high quality dataset which contains human with diverse clothing (such as skirt) and accessories (such as cap, hat, scarf).

We also utilize in total 2320 high quality commercial scans from AXYZ~\cite{axyz}, Treedy~\cite{treedy}, Twindom~\cite{twindom}, and RenderPeople~\cite{renderpeople}. All of these scans are with casual clothing and without interaction with objects.

\paragraph{Rendering}
For each scan, we render 100 views following a spiral trajectory with each view $i$:
\begin{align}
    \text{elevation}_i &= -\frac{1}{4} \pi + \frac{7}{8}\pi  * \frac{i}{100}, 
    \\
    \text{azimuth}_i &= 0 + 5\pi  * \frac{i}{100}.
    \label{eq:spiral_rendering_path}
\end{align}

Additionally, we render 32 views uniformly around z-axis with each view $j$:
\begin{align}
    \text{elevation}_j &= 0, \\
    \text{azimuth}_j &= 0 + \pi * \frac{j}{32}.
\end{align}

To protect the privacy of subjects in the training dataset, we only use the frontal view (with $\text{azimuth}_j \in [-\frac{\pi}{2}, \frac{\pi}{2}]$) as the input context view during training. Thus, we expect the model will not learn faces of subjects when it takes the back view as input.

\subsection{Evaluation Dataset}

\label{suppsec:evaluationdataset}
For quantitative evaluation, we use Sizer~\cite{tiwari2020sizer} and IIIT 3D human dataset~\cite{jinka2023iiit}. In this section, we start with introducing the two used evaluation datasets, and explain why we omit the commonly used CAPE dataset~\cite{ma2019cape} in our experiments. Finally, we provide a summary of the evaluation datasets.

\begin{figure}[!htp]
  \centering
  \includegraphics[width=\textwidth]{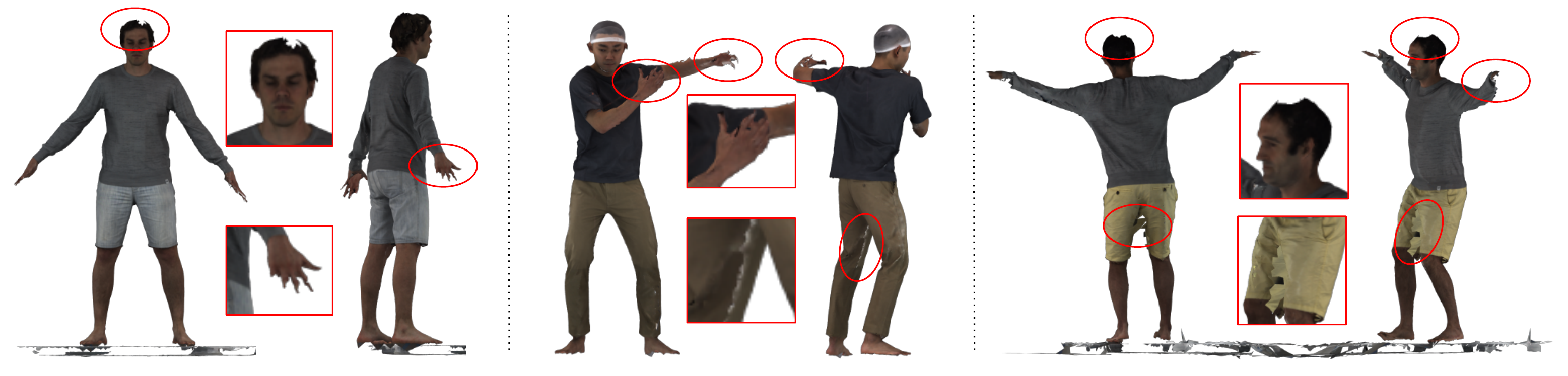}
  \caption{Example artifacts in CAPE~\cite{ma2019cape} dataset. Images shown here are rendered by ICON~\cite{xiu2023icon} due to the inaccessibility of original scans.}
  \label{fig:cape_example}
\end{figure}

\paragraph{CAPE}
Cape~\cite{ma2019cape} is a popular evaluation dataset which is widely used by previous methods~\cite{ho2023sith, xiu2023icon, xiu2023econ, zhang2023sifu}. We observe that CAPE has limitation in appearance variety, geometrical artifacts as well as the publicly unavailability. CAPE only contains simple clothing such as T-shirts and jeans, but no garments of loose clothing. As illustrated in Fig.~\ref{fig:cape_example}, CAPE contains several artifacts, such as holes on the head, missing hands, and wrong mesh geometry. We manually removed the noise and artifacts in scan to serve as our evaluation dataset. Moreover, CAPE doesn't have most original scans publicly available, but only the SMPL+D fitting. Due to this reason, we cannot render the CAPE scan to RGB images at desired camera view to evaluate the appearance performance such as PSNR, SSIM, and LPIPS. For above mentioned reasons, we also propose to use Sizer~\cite{tiwari2020sizer} and IIIT~\cite{jinka2023iiit} which are publicly available.

\paragraph{Sizer}
Sizer~\cite{tiwari2020sizer} is a high quality 3D human scan dataset which contains 100 different subjects wearing casual clothing items in various sizes. We randomly sample 150 scans from Sizer~\cite{tiwari2020sizer} as one of our evaluation dataset. 

\begin{figure}[!htp]
  \centering
  \includegraphics[width=\textwidth]{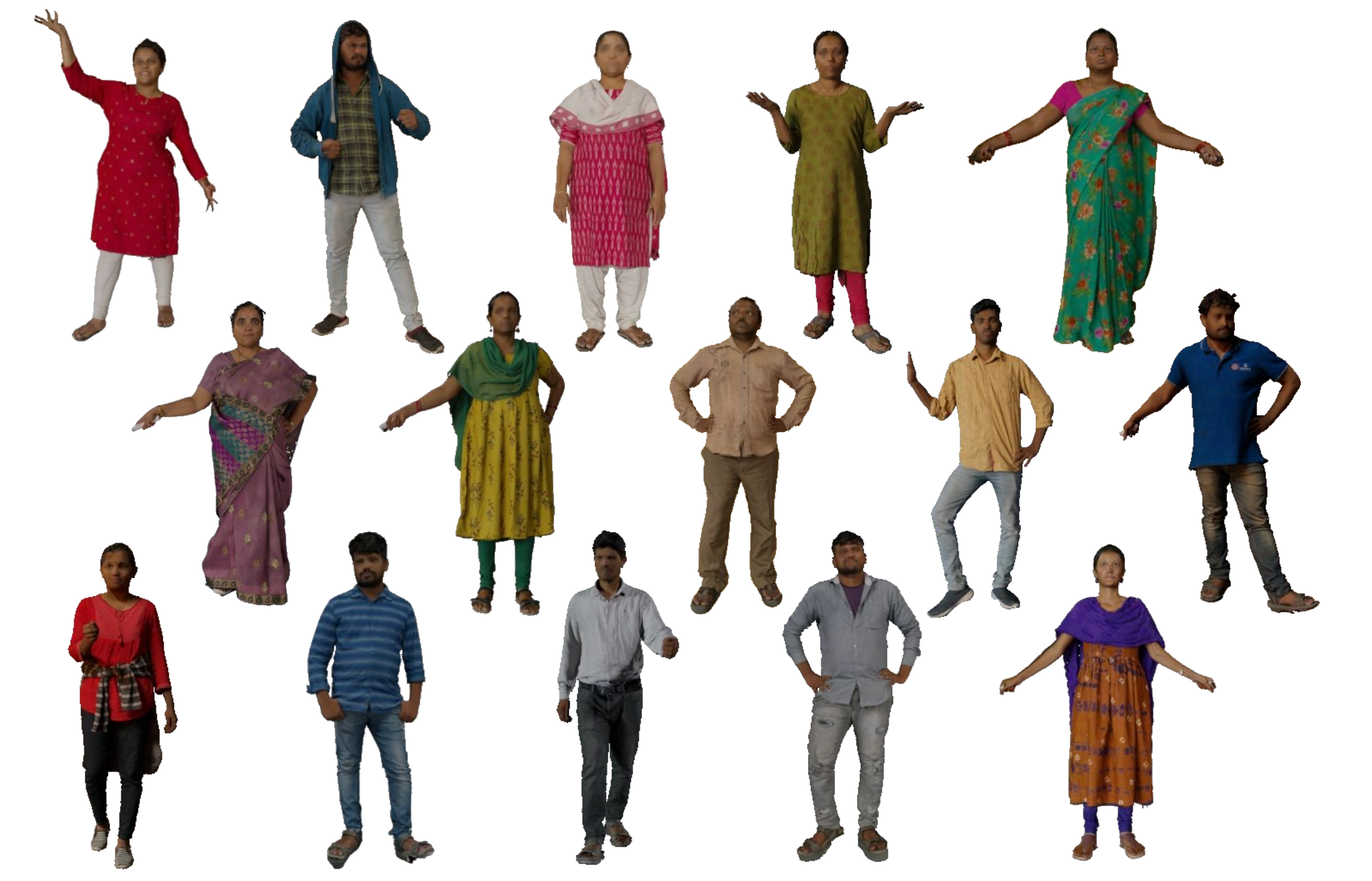}
  \caption{Example scans in IIIT~\cite{jinka2023iiit} dataset. IIIT contains subjects with diverse color pattern and loose garments, which rarely appear in training datasets.}
  \label{fig:iiit_examples}
\end{figure}

\paragraph{IIIT}
IIIT 3D Humans~\cite{jinka2023iiit} is a high quality dataset from IIIT Hyderabad in India. Different from the casual clothing setup in Sizer~\cite{tiwari2020sizer}, IIIT dataset mainly focuses on subjects wearing traditional India custom suits, including ethnicity, diverse color pattern and extremely loose clothing (Fig.~\ref{fig:iiit_examples}). It brings the huge variety of the subject appearance which can be considered as o.o.d. evaluation set to our model and baselines. We randomly sample 150 scans from IIIT dataset~\cite{jinka2023iiit} for evaluation. 

\paragraph{Summary}
We observe that the high quality datasets Sizer~\cite{tiwari2020sizer} and IIIT 3D Human~\cite{jinka2023iiit} are unexplored for the community of 3D avatars reconstruction. In fact, Sizer~\cite{tiwari2020sizer} contains casual clothing which is suitable to evaluate performance, and IIIT~\cite{jinka2023iiit} contains challenging texture and loose clothing which is suitable to evaluate robustness. All high quality scans in~\cite{jinka2023iiit, tiwari2020sizer} have no severe artifacts and are fully publicly available, which are the benefits unprovided in CAPE dataset~\cite{ma2019cape}. By evaluating on these datasets~\cite{jinka2023iiit, tiwari2020sizer} and release our randomly sampled subjects which are used in our experiments, we hope the 3D avatars community can discover and benefit from them.

\begin{figure}[!htp]
  \centering
  \includegraphics[width=\textwidth]{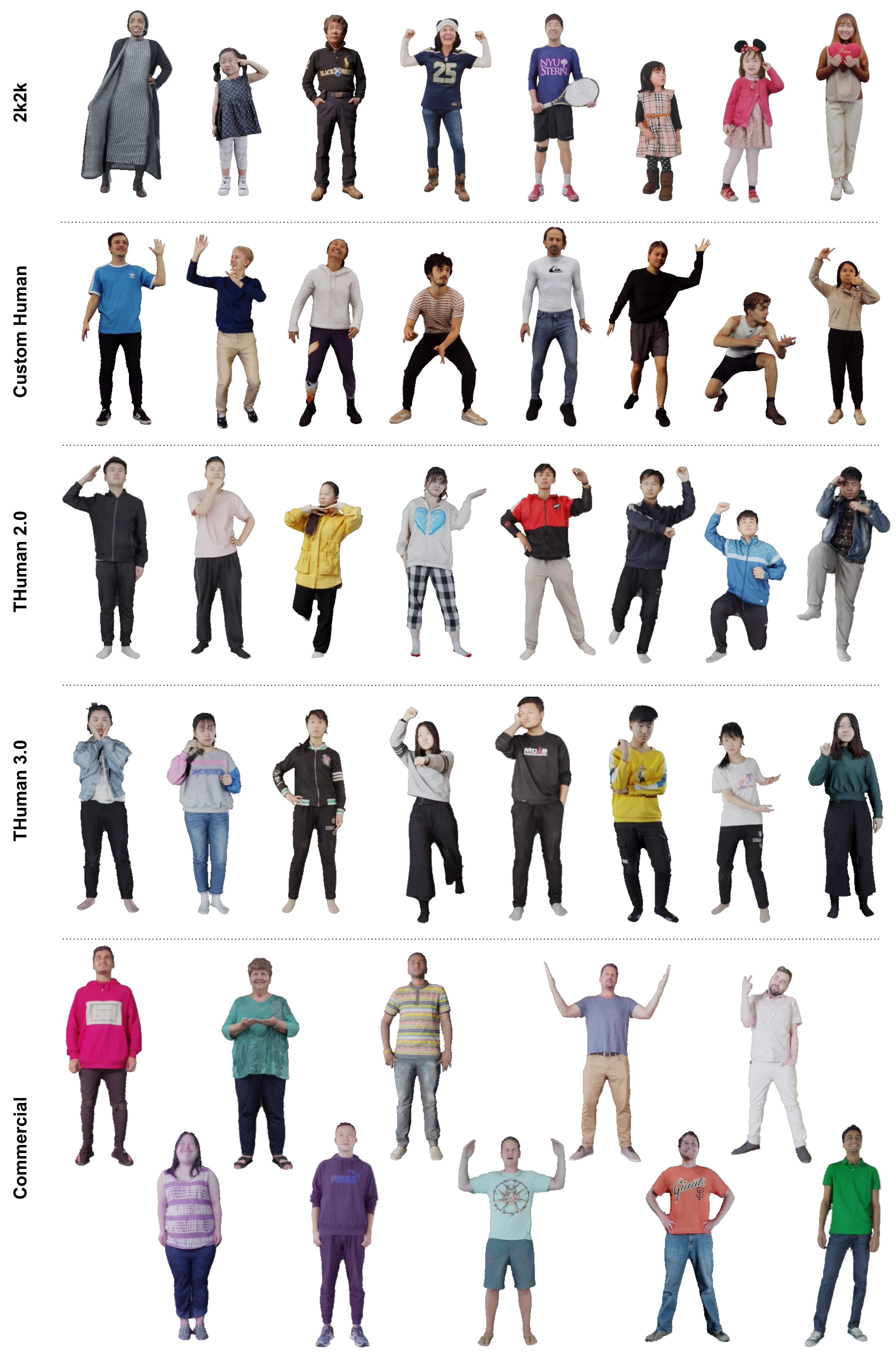}
  \caption{Example scans in training datasets~\cite{axyz, renderpeople, treedy, twindom, han20232k2k, ho2023customhuman, su2022thuman3, tao2021thuman2}.}
  \label{fig:training_example}
\end{figure}

\begin{figure}[!htp]
  \centering
  \includegraphics[width=\textwidth]{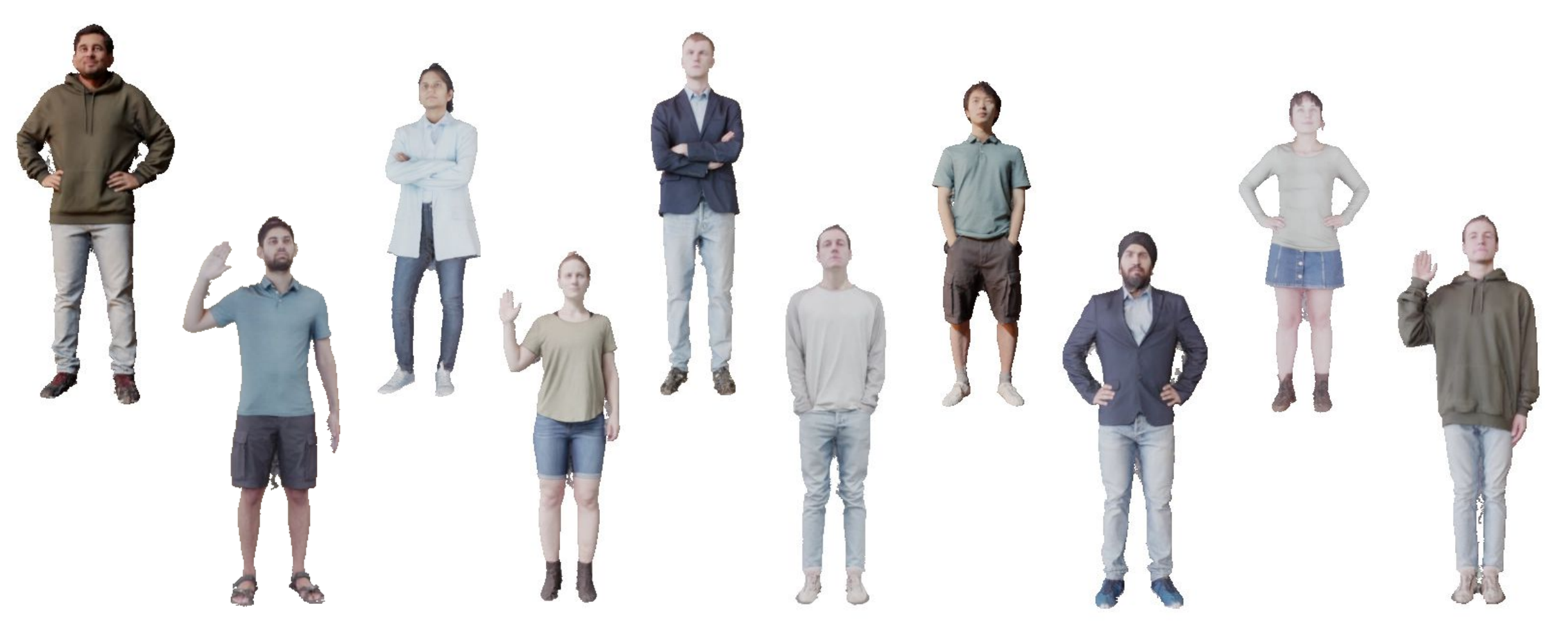}
  \caption{Example scans in Sizer~\cite{tiwari2020sizer} dataset. Sizer contains human in casual clothing.}
  \label{fig:sizer_examples}
\end{figure}

\newpage
\section{Failure Cases}
\label{sec:limitations}
\begin{figure}[!htp]
  \centering
  \includegraphics[width=0.7\textwidth]{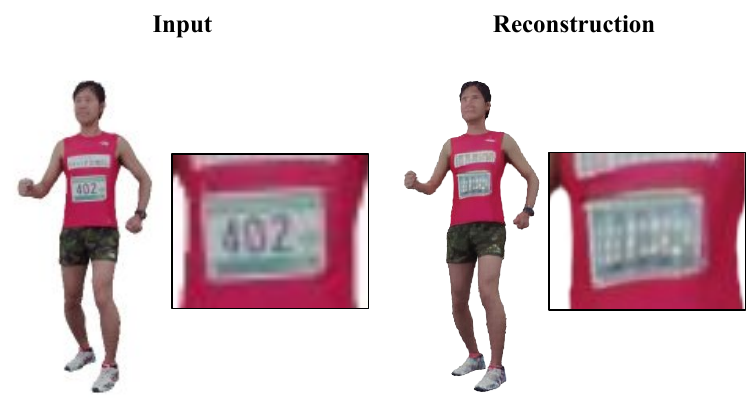}
  \caption{Failure Case: our model cannot reconstruct the numbers on the cloth.}
  \label{fig:failure_case1}
\end{figure}

Limited by low resolution ($256\times256$) of our multi-view diffusion model~\cite{Wang2023ImageDream}, our model can often fail in reconstructing fine details such as text on the cloth as illustrated in ~\cref{fig:failure_case1}. One potential solution is to switch to a recent powerful high-resolution multi-view diffusion models~\cite{gao2024cat3d, tang2023mvdiffusion++}.

\begin{figure}[!htp]
  \centering
  \includegraphics[width=\textwidth]{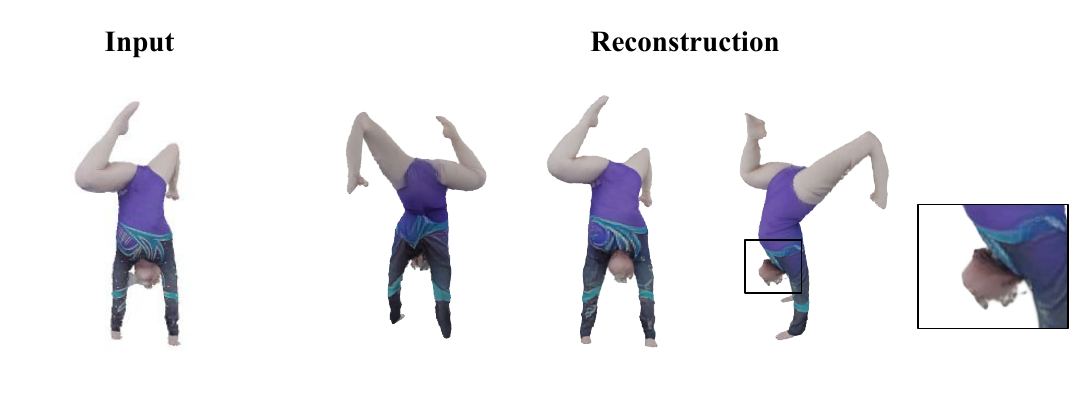}
  \caption{Failure Case: our model fails in infer appearance of human with challenging pose.}
  \label{fig:failure_case2}
\end{figure}

\vspace{0.5in}
In addition, we observe that our model can fail when reconstructing human extremely challenging poses. As shown in ~\cref{fig:failure_case2}, our model cannot infer head geometry and appearance accurately due to the challenging pose in input image.

\section{Broader Impacts}
\label{sec:a_broader}
Our work shows generality across different ethnicities and humans, providing a useful tool for a fair representation of different cultures.
Having a robust method to synthesize realistic 3D geometry from a single RGB image may be used in surveillance and inappropriate content generation. 


\newpage
\newpage
\section*{NeurIPS Paper Checklist}

\begin{enumerate}

\item {\bf Claims}
    \item[] Question: Do the main claims made in the abstract and introduction accurately reflect the paper's contributions and scope?
    \item[] Answer: \answerYes{} 
    \item[] Justification: Contributions and scope are included in the abstract and the introduction.

\item {\bf Limitations}
    \item[] Question: Does the paper discuss the limitations of the work performed by the authors?
    \item[] Answer: \answerYes{} 
    \item[] Justification: Please refer to Appendix \ref{sec:limitations}

\item {\bf Theory Assumptions and Proofs}
    \item[] Question: For each theoretical result, does the paper provide the full set of assumptions and a complete (and correct) proof?
    \item[] Answer: \answerNA{} 
    \item[] Justification: The paper does not contain theoretical results.

    \item {\bf Experimental Result Reproducibility}
    \item[] Question: Does the paper fully disclose all the information needed to reproduce the main experimental results of the paper to the extent that it affects the main claims and/or conclusions of the paper (regardless of whether the code and data are provided or not)?
    \item[] Answer: \answerYes{} 
    \item[] Justification: Please refer to our Appendix for Implementation details.

\item {\bf Open access to data and code}
    \item[] Question: Does the paper provide open access to the data and code, with sufficient instructions to faithfully reproduce the main experimental results, as described in supplemental material?
    \item[] Answer: \answerYes{} 
    \item[] Justification: All code and scripts are publicly available \href{https://yuxuan-xue.com/human-3diffusion}{here}.

\item {\bf Experimental Setting/Details}
    \item[] Question: Does the paper specify all the training and test details (e.g., data splits, hyperparameters, how they were chosen, type of optimizer, etc.) necessary to understand the results?
    \item[] Answer: \answerYes{} 
    \item[] Justification: We include details about datasets and implementation in ~\cref{subsec:exp-setup}, ~\cref{suppsec:training_details}, ~\cref{suppsec:traninigdataset}, ~\cref{suppsec:evaluationdataset}.

\item {\bf Experiment Statistical Significance}
    \item[] Question: Does the paper report error bars suitably and correctly defined or other appropriate information about the statistical significance of the experiments?
    \item[] Answer: \answerNo{} 
    \item[] Justification: We follow standard evaluation protocols reporting average errors on an extensive test set.

\item {\bf Experiments Compute Resources}
    \item[] Question: For each experiment, does the paper provide sufficient information on the computer resources (type of compute workers, memory, time of execution) needed to reproduce the experiments?
    \item[] Answer: \answerYes{} 
    \item[] Justification: The paper includes the computational resources used in our experiments in Section \ref{sec:experiments}.
    
\item {\bf Code Of Ethics}
    \item[] Question: Does the research conducted in the paper conform, in every respect, with the NeurIPS Code of Ethics \url{https://neurips.cc/public/EthicsGuidelines}?
    \item[] Answer: \answerYes{} 
    \item[] Justification: We read and adhere to the NeurIPS Code of Ethics.

\item {\bf Broader Impacts}
    \item[] Question: Does the paper discuss both potential positive societal impacts and negative societal impacts of the work performed?
    \item[] Answer: \answerYes{} 
    \item[] Justification: The paper include the Broader Impacts statement in Appendix \ref{sec:a_broader}.
    
\item {\bf Safeguards}
    \item[] Question: Does the paper describe safeguards that have been put in place for responsible release of data or models that have a high risk for misuse (e.g., pretrained language models, image generators, or scraped datasets)?
    \item[] Answer: \answerNA{} 
    \item[] Justification: We do not see such high risk posed by our paper.

\item {\bf Licenses for existing assets}
    \item[] Question: Are the creators or original owners of assets (e.g., code, data, models), used in the paper, properly credited and are the license and terms of use explicitly mentioned and properly respected?
    \item[] Answer: \answerYes{} 
    \item[] Justification: We cite and explicitly refer to all the legitimate sources of code, data, and models.

\item {\bf New Assets}
    \item[] Question: Are new assets introduced in the paper well documented and is the documentation provided alongside the assets?
    \item[] Answer: \answerNA{} 
    \item[] Justification: We do not include new assets, but we will include appropriate documentation upon code release after acceptance.

\item {\bf Crowdsourcing and Research with Human Subjects}
    \item[] Question: For crowdsourcing experiments and research with human subjects, does the paper include the full text of instructions given to participants and screenshots, if applicable, as well as details about compensation (if any)? 
    \item[] Answer: \answerNA{} 
    \item[] Justification: Although the method involves 3D human models, we rely on datasets collected before this work; we refer to them for their specifics.

\item {\bf Institutional Review Board (IRB) Approvals or Equivalent for Research with Human Subjects}
    \item[] Question: Does the paper describe potential risks incurred by study participants, whether such risks were disclosed to the subjects, and whether Institutional Review Board (IRB) approvals (or an equivalent approval/review based on the requirements of your country or institution) were obtained?
    \item[] Answer: \answerNA{} 
    \item[] Justification: NA{} 
    \item[] Justification: Although the method involves 3D human models, we rely on datasets collected before this work; we refer to them for their IRB approvals.

\end{enumerate}

\end{document}